\documentclass[10pt]{elsarticle}

\usepackage{bm}
\usepackage{mathtools}
\usepackage{graphicx}
\usepackage{multirow}
\usepackage{comment}
\newcommand{\norm}[1]{\left\lVert #1 \right\rVert}
\usepackage{algorithm}
\usepackage{algpseudocode}
\usepackage{algpascal}
 \usepackage{textcomp}
 \usepackage{amsthm}

\theoremstyle{remark}
\newtheorem{rem}{Remark}
\newtheorem{prop}{Proposition}

 \usepackage{tikz}

 \usepackage{smartdiagram}
 \usepackage[utf8]{inputenc}
 \usetikzlibrary{arrows,positioning}
\usepackage{makecell} 
\setcellgapes{5pt}

\usepackage{amsmath,amsfonts,amssymb}

\DeclareMathOperator*{\argmin}{arg\,min}

\usepackage{tikz}

\usepackage{fancybox}
\usepackage{colortbl}
\definecolor{shadecolor}{RGB}{255 ,45 ,0}

\newcommand{\circo}{~\raisebox{1pt}{\tikz \draw[line width=0.6pt] circle(1.1pt);}~}

\usepackage{hyperref}

\journal{}

\begin{document}

\begin{frontmatter}

\title{Fuzzy clustering of ordinal time series based on two novel distances with economic applications}


\author[mymainaddress]{\'Angel L\'opez-Oriona\corref{mycorrespondingauthor} (ORCID 0000-0003-1456-7342)}
\ead{oriona38@hotmail.com, a.oriona@udc.es}


 \author[weissaddress]{Christian H.\ Wei\ss{} (ORCID 0000-0001-8739-6631)}
 \ead{weissc@hsu-hh.de}

 \author[mymainaddress]{Jos\'e A. Vilar (ORCID 0000-0001-5494-171X)}
 \cortext[mycorrespondingauthor]{Corresponding author}
 \ead{jose.vilarf@udc.es}

\address[mymainaddress]{Research Group MODES, Research Center for Information and Communication Technologies (CITIC), University of A Coru\~na, A Coru\~na, Spain.}

  
\address[weissaddress]{Department of Mathematics and Statistics, Helmut Schmidt University, Hamburg, Germany.}

\begin{abstract}
Time series clustering is a central machine learning task with applications in many fields. While the majority of the methods focus on real-valued time series, very few works consider series with discrete response. In this paper, the problem of clustering ordinal time series is addressed. To this aim, two novel distances between ordinal time series are introduced and used to construct fuzzy clustering procedures. Both metrics are functions of the estimated cumulative probabilities, thus automatically taking advantage of the ordering inherent to the series' range. The resulting clustering algorithms are computationally efficient and able to group series generated from similar stochastic processes, reaching accurate results even though the series come from a wide variety of models. Since the dynamic of the series may vary over the time, we adopt a fuzzy approach, thus enabling the procedures to locate each series into several clusters with different membership degrees. An extensive simulation study shows that the proposed methods outperform several alternative procedures. Weighted versions of the clustering algorithms are also presented and their advantages with respect to the original methods are discussed. Two specific applications involving economic time series illustrate the usefulness of the proposed approaches.
\end{abstract}


\end{frontmatter}

\section{Introduction}\label{sectionintroduction}

Time series clustering concerns the problem of splitting a set of unlabelled time series into homogeneous groups in such a way that similar series are placed together in the same group and dissimilar series are located in different groups. Indeed, the clustering task is driven by the desired similarity notion, which can be established in different ways by dealing with time series. Frequently, the purpose is to identify groups with similar generating models, which allows to characterize a few dynamic patterns without requiring to analyze and model each single time series. The latter, besides being computationally demanding, rarely is the objective when dealing with a huge number of series. Complexity inherent to clustering objects evolving over time (fixing a suitable dissimilarity principle, dealing with  series of unequal length, high computational complexity,\ldots) together with the vast range of applications where time series clustering plays a fundamental role account for the growing interest on this challenging topic. Comprehensive overviews including current advances, future prospects, interesting
references, and specific application areas are provided by \cite{liao2005clustering, aghabozorgi2015time, maharaj2019time}.


The majority of clustering methods focus on real-valued time series. For instance, some techniques are based on discriminating between different geometric profiles in the time series data set by employing the dynamic time warping (DTW) distance or some related dissimilarities \cite{izakian2015fuzzy, luczak2016hierarchical, d2021trimmed}. A different approach consists of assuming that each time series has been generated from a specific class of models and then executing a clustering algorithm based on estimated models \cite{frohwirth2008model, corduas2008time, d2013autoregressive, d2016garch}. Some other works propose to replace each series in the collection by a vector of model-free features describing its behaviour in a suitable way. Then, the computed vectors are used as input to a standard clustering method \cite{d2009autocorrelation, maharaj2011fuzzy, d2012wavelets, lafuente2016clustering, lopez2021quantile, lopez2022quantile1, lopez2022quantile2}. Alternative techniques are based on reducing the dimensionality of the original time series as a preliminary step \cite{singhal2005clustering, egri2017cross, pealat2021improved}. Then, a specific clustering procedure is applied to the set of reduced objects. The suitability of each class of algorithms usually depends on the nature of the time series and the final goal of the user, with no approach dominating the remaining ones in every possible context. 

According to the cluster assignment criterion, two different paradigms are considered depending on whether a “hard” or “soft” partition is constructed. 
Traditional clustering leads to hard solutions, where each data object is located in exactly one cluster. Overlapping groups are not allowed, which could become too inflexible in many real life applications where the cluster boundaries are not clearly determined or some objects are equidistant from various clusters. Soft cluster techniques provide a more versatile tool by allowing membership of data objects to clusters. In a soft partition, every object is associated with a vector of membership degrees indicating the amount of confidence in the assignment to each of the respective clusters. A well known approach to perform soft clustering is via the fuzzy clustering methods \cite{bezdek2013pattern, miyamoto2008algorithms}, based on minimizing a cost function involving distances to centroids and the so-called \textit{fuzzifier} controlling the allowed overlapping level. Adoption of fuzzy approach is usually advantageous when dealing with time series data sets because regime shifts are frequent in practice.


A considerably smaller number of works have dealt with the clustering of time series having another range than a real-valued one. For instance, \cite{etienne2014model} and \cite{cerqueti2022ingarch} introduced clustering algorithms for count time series based on Poisson mixtures and integer-valued generalized autoregressive conditional heteroscedasticity (INGARCH) models, respectively. Several approaches to cluster categorical time series (CTS) were also proposed. \cite{pamminger2010model} considered two model-based procedures relying on time-homogeneous first-order Markov chains, which are applied to a panel of Austrian wage mobility data. A dissimilarity assessing both closeness of raw categorical values and proximity between dynamic behaviours was proposed by \cite{garcia2015framework}. The metric is used to perform clustering aimed at identifying different web-user profiles according to their navigation behaviour. \cite{jahanshahi2022ntreeclus} constructed a robust tree-based sequence encoder for clustering CTS, which is applied to some real-world data sets containing biological sequences. Two novel distances between CTS were introduced by \cite{lopez2023hard} and employed to perform hard and soft clustering. An overview of model-based clustering of categorical sequences is provided in \cite{melnykov2016clickclust}, and several methods based on finite mixtures of Markov models are implemented in the R package \textbf{ClickClust} \cite{melnykov2014package}.

To the best of our knowledge, all the proposed clustering methods for CTS are designed for the general case of series taking \emph{nominal} values, i.e., if no underlying ordering exists in the categorical range. Clearly, these methods are still valid to cluster \textit{ordinal} time series (OTS). However, when applying these procedures to OTS data sets, one completely ignores the latent ordering, which could be of great help  to identify the underlying partition. For instance, consider a data set including three clusters, $\mathcal{C}_1, \mathcal{C}_2$, and $\mathcal{C}_3$, characterised by time series taking ``low'', ``moderate'', and ``large'' values, respectively. In such a case, it is reasonable to consider the groups $\mathcal{C}_1$ and $\mathcal{C}_3$ to be furthest away, and some degree of ordinal information should be provided to the clustering algorithm. Additionally, OTS data sets appear rather naturally in several application domains, including economics (e.g., wage mobility data of different individuals \cite{pamminger2010model} or credit ratings of different countries \cite{weiss2019distance}), environmental sciences (e.g., amount of cloud coverage in different regions \cite{weiss2020regime}), or medicine (e.g., clinical scores of different subjects \cite{koss2022hierarchical}), among others. 
The previous considerations clearly highlight the need for clustering algorithms specifically designed to deal with OTS. Moreover, given the complex nature of time series databases, the adoption of the fuzzy approach would allow to gain versatility in the resulting partition to capture changes in the dynamic behaviours of the series over time.


The main goal of this paper is to introduce fuzzy clustering algorithms for OTS being capable of: (i) grouping together ordinal sequences generated from similar stochastic processes, (ii) achieving accurate results with series coming from a broad variety of ordinal models, and (iii) performing the clustering task in an efficient manner. To this aim, we first introduce two dissimilarity measures between OTS. Since our objective is to group series with similar underlying structures, both metrics are based on extracted features providing information about marginal properties and serial dependence patterns. Specifically, the first dissimilarity considers proper estimates of the cumulative probabilities, while the second one combines structure-based statistical features characterizing a given OTS (dispersion, skewness, serial dependence \ldots{}), which, in turn, are defined in terms of the estimated cumulative probabilities. Thus, the distances take advantage of the latent ordering existing in the series' range. Both metrics are used as input to the standard fuzzy $C$-medoids algorithm, which allows for the assignment of gradual memberships of the OTS to clusters. 

Assessment of the clustering approaches is carried out by means of a comprehensive simulation study including different ordinal processes commonly used in the literature. The performance of some alternative dissimilarities designed to deal with real-valued or nominal time series are also examined for comparison. Two evaluation schemes are considered. The first one aims at analysing the ability of the procedures to assign high (low) membership values if a given series pertains (not pertains) to a specific cluster defined in advance. The second scheme also assesses the ability of the approaches to handle OTS showing an ambiguous behaviour i.e., series whose dynamic structure is not associated with a specific group. More sophisticated versions of both clustering procedures are also constructed by giving different weights to the marginal and serial components of the proposed metrics. In this way, the influence of each component in the computation of the clustering solution can be automatically determined during the optimisation process. Lastly, two specific applications involving economic time series are presented to show the usefulness of the proposed clustering techniques. 

The rest of the paper is organised as follows. Several quantities for describing an ordinal process and two distances between OTS based on proper estimates of these features are introduced in Section \ref{sectiondistances}, where some simple examples are also shown to illustrate the suitability of the metrics. In Section \ref{sectionfuzzyalgorithms}, fuzzy clustering algorithms based on the proposed dissimilarities are constructed. The methods are evaluated in Section \ref{sectionsimulationstudy} by means of a broad simulation study where several alternative procedures are analysed as well. Section \ref{sectionapplications} presents two applications of the proposed techniques to data sets containing economic time series. Some concluding remarks are summarised in Section \ref{sectionconclusions}. The Appendix provides the proof of two results presented in the manuscript.

\section{Two distance measures between ordinal time series}\label{sectiondistances}

In this section, after providing some background on ordinal stochastic processes, two novel distances between ordinal time series are introduced. The use of estimated cumulative probabilities to construct both metrics is also discussed and properly motivated.


\subsection{Some background on ordinal processes}\label{subsectionbackground}

Let $\{X_t\}_{t \in \mathbb{Z}}$, $\mathbb{Z}=\{\ldots,-1,0,1,\ldots\}$, be a strictly stationary stochastic process having the ordered categorical range $\mathcal{S}=\{s_0, \ldots, s_n\}$, with $s_0<s_1<\ldots<s_n$. The process $\{X_t\}_{t \in \mathbb{Z}}$ is often referred to as an \textit{ordinal process}, while the categories in $\mathcal{S}$ are frequently called the \textit{states}. Let $\{C_t\}_{t \in \mathbb{Z}}$ be the count process with range $\{0, \ldots, n\}$ generating  the ordinal process $\{X_t\}_{t \in \mathbb{Z}}$, i.e., $X_t=s_{C_t}$. It is well known that the distributional properties of $\{C_t\}_{t \in \mathbb{Z}}$ (e.g., stationarity) are properly inherited by $\{X_t\}_{t \in \mathbb{Z}}$ \cite{weiss2018introduction}. In particular, the marginal probabilities can be expressed as
\begin{equation}\label{ncp1}
p_i=P(X_t=s_i)=P(C_t=i), \quad i=0,\ldots,n,
\end{equation}
while the lagged joint probabilities (for a lag $l \in \mathbb{Z}$) are given by 
\begin{equation}\label{ncp2}
p_{ij}(l)=P(X_t=s_j, X_{t-l}=s_i)=P(C_t=j, C_{t-l}=i), \quad \, i,j=0,\ldots,n.
\end{equation}

Note that both the marginal and the joint probabilities are still well defined in the general case of a stationary stochastic process with nominal range. 
By contrast, in an ordinal process, one can also consider the corresponding cumulative probabilities defined, for $i,j=0,\ldots,n-1$ and $l \in \mathbb{Z}$, as 
\begin{equation}\label{cp}
	\begin{split}
		f_i=&P(X_t \le s_i)=P(C_t \le i), \\
		f_{ij}(l)=&P(X_t \le s_j, X_{t-l} \le s_i)=P(C_t \le j, C_{t-l}\le i), 
	\end{split}
\end{equation}
for the marginal and the joint case, respectively. 
 
In practice, the values of $p_i$, $p_{ij}(l)$, $f_i$, and $f_{ij}(l)$ must be estimated from a $T$-length realization of the ordinal process, $X_T=\{x_1, \ldots, x_T\}$, usually referred to as \textit{ordinal time series} (OTS). Natural estimates of these probabilities are given by
\begin{align}
\widehat{p}_i=\frac{1}{T}\sum_{k=1}^{T}I(x_k=s_i), & \quad \widehat{p}_{ij}(l)=\frac{1}{T-l}\sum_{k=1}^{T-l}I(x_k=s_i)I(x_{k+l}=s_j),  \label{estimatesprobabilities} \\ 
\widehat{f}_i=\frac{1}{T}\sum_{k=1}^{T}I(x_k \le s_i), & \quad \widehat{f}_{ij}(l)=\frac{1}{T-l}\sum_{k=1}^{T-l}I(x_k \le s_i)I(x_{k+l} \le s_j), \label{estimatescprobabilities}
\end{align}
where $I(\cdot)$ denotes the indicator function.

Probabilities $p_i$, $p_{ij}(l)$, $f_i$ and $f_{ij}(l)$ summarize the marginal and joint distributional properties of the process $\{X_t\}_{t \in \mathbb{Z}}$. An alternative way to characterize the process consists of constructing a fixed-length vector formed by statistical features measuring structural properties (centrality, dispersion, skewness, serial dependence\ldots). Following \cite{weiss2019distance}, a range of this type of features can be quantified by using expected values of suitable distances between ordinal categories. Thus, specific expected values of the so-called \textit{block} distance, $d_{\text{o},1}(s_i, s_j)=|i-j|$, lead to the set 
of structural features provided in Table~\ref{tablefeatures}. For example, 
$\text{loc}_{d_{\text{o},1}}$ is the expected value of the block distance between the marginal variable $X_t$ and the state $s_0$, $\text{disp}_{d_{\text{o},1}}$ is the expected value of the block distance between two copies of the marginal variable, \ldots (see \cite{weiss2019distance} for details). While the first four measures in Table~\ref{tablefeatures} summarise the marginal behaviour of the process, the ordinal Cohen's $\kappa$, $\kappa_{d_{\text{o},1}}(l)$, evaluates the degree of serial dependence at a given lag $l \in \mathbb{Z}$. Thus, we have available a unified distance-based approach to obtain relevant features providing a comprehensive picture of the process. Note that the block distance between two given categories simply counts the number of categories between them, but it makes use of the latent ordering thus providing a natural way of assessing dissimilarity between ordinal categories. Furthermore, $d_{\text{o},1}$ does not depend on the labeling selected for the categories, which ensures that the features based on the expected values of this distance are invariant to scale transformations. These nice properties justify the use of this distance-based approach.  




\begin{table}[ht]
\centering
	\begin{tabular}{lcl} \hline
		Feature & & Definition \\ \hline 
		Location &  &    $\text{loc}_{d_{\text{o},1}}=\sum_{i=0}^{n-1}(i+1)(f_{i+1}-f_i)$      \\
		Dispersion & &    $\text{disp}_{d_{\text{o},1}}=2\sum_{i=0}^{n-1}f_i(1-f_i)$    \\
		Asymmetry &  & $\text{asym}_{d_{\text{o},1}}=\sum_{i=0}^{n-1}(1-f_i-f_{n-i-1})^2$        \\
		Skewness &  &  $\text{skew}_{d_{\text{o},1}}=2\sum_{i=0}^{n-1}f_i-1$        \\
		Ordinal Cohen's $\kappa$ at lag $l \in \mathbb{Z}$&   & $\kappa_{d_{\text{o},1}}(l)=\frac{\sum_{i=0}^{n-1}(f_{ii}(l)-f_i^2)}{\sum_{i=0}^{n-1}f_i(1-f_i)}$         \\ \hline           
	\end{tabular}
	\caption{Some features of an ordinal process based on expected values of the block distance.}
	\label{tablefeatures}
\end{table}


When dealing with a realization $X_T$, estimates $\widehat{\text{loc}}_{d_{\text{o},1}}$, $\widehat{\text{disp}}_{d_{\text{o},1}}$, $\widehat{\text{asym}}_{d_{\text{o},1}}$, $\widehat{\text{skew}}_{d_{\text{o},1}}$, and $\widehat{\kappa}_{d_{\text{o},1}}(l)$ for the respective features in Table~\ref{tablefeatures} can be obtained by considering their sample counterparts, i.e. using $\widehat{f}_{i}$ and $\widehat{f}_{ij}(l)$ in \eqref{estimatescprobabilities}. A detailed analysis of the asymptotic properties of these estimators is provided in \cite{weiss2019distance}. 


\subsection{Two novel dissimilarities between ordinal time series}\label{subsection2dissimilarities}


Suppose we have two stationary ordinal processes $\{X_t^{(1)}\}_{t \in \mathbb{Z}}$ and $\{X_t^{(2)}\}_{t \in \mathbb{Z}}$ having the same range~$\mathcal{S}$. A simple dissimilarity criterion between both processes can be established by measuring the discrepancy between their corresponding representations in terms of cumulative probabilities. In this way, for a given collection of $L$ lags, $\mathcal{L}=\{l_1, \ldots, l_L\}$, we define a distance $d_1$ as
\begin{equation}\label{d1}
	d_1\big(X_t^{(1)}, X_t^{(2)}\big)=d_{1, M}\big(X_t^{(1)}, X_t^{(2)}\big)+d_{1, B}\big(X_t^{(1)}, X_t^{(2)}\big),
\end{equation} 
with 
\begin{equation}\label{d1components}
	\begin{split}
		d_{1, M}\big(X_t^{(1)}, X_t^{(2)}\big)=& \sum_{i=0}^{n-1}\Big(f_i^{(1)}-f_i^{(2)}\Big)^2,\\
		d_{1, B}\big(X_t^{(1)}, X_t^{(2)}\big)=& \sum_{k=1}^{L}\sum_{i=0}^{n-1}\sum_{j=0}^{n-1}\Big(f^{(1)}_{ij}(l_k)-f^{(2)}_{ij}(l_k)\Big)^2,
	\end{split}
\end{equation}
where the superscripts $(1)$ and $(2)$ indicate that the corresponding probabilities refer to the processes  $\{X_t^{(1)}\}_{t \in \mathbb{Z}}$ and $\{X_t^{(2)}\}_{t \in \mathbb{Z}}$, respectively. The terms $d_{1, M}$ and $d_{1, B}$ assess dissimilarity between marginal and lagged bivariate probabilities, respectively. The latter term involves the set $\mathcal{L}$, which must be fixed in advance according to the lags at which one wishes to evaluate the serial dependence. It is worth remarking that, by considering the cumulative probabilities in the definition of $d_1$, we obtain an appropriate dissimilarity measure taking into account the ordering existing in both processes (see Section~\ref{subsectionmotivating}). 

An alternative dissimilarity measure considering features based on the block distance $d_{\text{o},1}$ is defined as
\begin{equation}\label{d2}
	d_2\big(X_t^{(1)}, X_t^{(2)}\big)=d_{2, M}\big(X_t^{(1)}, X_t^{(2)}\big)+d_{2, B}\big(X_t^{(1)}, X_t^{(2)}\big),
\end{equation}
with 
\begin{equation}\label{d2components}
	\begin{split}
		d_{2, M}\big(X_t^{(1)}, X_t^{(2)}\big)=& \, \Big\|\frac{1}{n} \Big( \text{loc}_{d_{\text{o},1}}^{(1)}, 2\text{disp}_{d_{\text{o},1}}^{(1)}, \text{asym}_{d_{\text{o},1}}^{(1)}, \text{skew}_{d_{\text{o},1}}^{(1)} \Big) \\
  & - \, \frac{1}{n} \Big( \text{loc}_{d_{\text{o},1}}^{(2)}, 2\text{disp}_{d_{\text{o},1}}^{(2)}, \text{asym}_{d_{\text{o},1}}^{(2)}, \text{skew}_{d_{\text{o},1}}^{(2}\Big)\Big\|^2,\\
		d_{2, B}\big(X_t^{(1)}, X_t^{(2)}\big)=&\sum_{k=1}^{L}\Big(\kappa_{d_{o, 1}}^{(1)}(l_k)-\kappa_{d_{o, 1}}^{(2)}(l_k)\Big)^2.
	\end{split}
\end{equation}

In the same way as $d_1$, dissimilarity $d_2$ is formed by the terms $d_{2, M}$ and $d_{2, B}$ comparing the marginal and serial behaviours of both processes, respectively. Specifically, the marginal component contains the normalised versions of the quantities in Table \ref{tablefeatures} (see \cite{weiss2019distance}). This way, each one of the features is expected to exhibit approximately the same influence in the computation of $d_{2, M}$. 

In practice, $d_1$ and $d_2$ will be approximated on the basis of realizations $X_{T_1}^{(1)}$ and $X_{T_2}^{(2)}$ of both ordinal processes with respective lengths $T_1$ and $T_2$ by means of 
\begin{equation}
	\widehat{d}_p\big(X_{T_1}^{(1)}, X_{T_2}^{(2)}\big)=\widehat{d}_{p, M}\big(X_{T_1}^{(1)}, X_{T_2}^{(2)}\big)+\widehat{d}_{p, B}\big(X_{T_1}^{(1)}, X_{T_2}^{(2)}\big), \,\,\, \mbox{ for } \, p=1,2,
\end{equation}
where $\widehat{d}_{p, M}$ and $\widehat{d}_{p, B}$ are proper estimates of ${d}_{p, M}$ and ${d}_{p, B}$ computed by using the sample values $\widehat{f}_i^{(h)}$, $\widehat{f}^{(h)}_{ij}(l_k)$, $h=1,2$, given in \eqref{estimatescprobabilities}.


Some remarks concerning the proposed dissimilarities are provided below.

\vspace*{0.1cm}

    \begin{rem}
    \label{remJointMetrics}
\textit{Independent consideration of metrics $d_1$ and $d_2$}. Metrics $d_1$ and $d_2$ could be jointly considered to define a combined dissimilarity $d_1+d_2$. Although this distance could be seen as more informative than both $d_1$ and $d_2$, this is not usually the case in practice. In fact, some numerical experiments have revealed that, in most cases, a clustering algorithm based on one of the individual distances, $\widehat{d}_1$ or $\widehat{d}_2$, outperforms a method based on the combined distance $\widehat{d}_1+\widehat{d}_2$ in terms of clustering accuracy. This is due to the fact that, by using all features to describe a given OTS, redundant information is being provided (note that the features in Table~\ref{tablefeatures} are defined in terms of cumulative probabilities). It is worth noting that the use of redundant features is known to be counterproductive in clustering and classification contexts. 
    \end{rem}

\vspace*{0.1cm}

\begin{rem}
    \label{remFeatureMetric}
    \textit{Advantages of feature-based distances}. Both $\widehat{d}_1$ and $\widehat{d}_2$ belong to the class of feature-based distances, since they are aimed at comparing extracted features. The discriminatory capability of this kind of distances depends on selecting the most suitable features for a given context. Whether a proper set of features is used, then this class of distances present very nice properties such as dimensionality reduction, low computational complexity, robustness to the generating model, and versatility to compare series with different lengths. It is worth remarking that these properties are not satisfied by other dissimilarities between time series. For instance, metrics based on raw data usually involve high computational cost and require series having the same length, while model-based metrics are expected to be strongly sensitive to model misspecification.
    \end{rem}

\vspace*{0.1cm}

\begin{rem}
    \label{remDist2}
    \textit{On the distance $d_2$}. Distance $d_2$ and its estimate rely on features based on expectations of the block distance between ordinal categories, $d_{\text{o},1}$. Other feature-based distances can be introduced following an analogous approach, but starting from alternative distances defined on $\mathcal{S} \times\mathcal{S}$ (see Section~2 in \cite{weiss2019distance}). However, these alternative vias led to distances showing a worse performance than $d_2$ in the numerical experiments carried out throughout this work for clustering purposes, i.e. $d_2$ exhibited the highest capability to discriminate between different OTS (see Sections~\ref{sectionfuzzyalgorithms} and \ref{sectionsimulationstudy}). For this reason, $d_2$ was selected. 
    
    \end{rem}

 \subsection{Motivating the use of cumulative probabilities}\label{subsectionmotivating}

This section illustrates the advantages of using cumulative probabilities to differentiate between ordinal processes. For the sake of simplicity, we first consider a toy example involving synthetic data and put the focus on the marginal case.
Let us consider three stationary processes with ordinal range $\mathcal{S}=\{s_0, s_1, s_2, s_3\}$, denoted by $X_t^{(1)}$, $X_t^{(2)}$, and $X_t^{(3)}$, with marginal probabilities given by the vectors   $\boldsymbol{p}_i=\Big(P\big(X_t^{(i)}=s_0\big), \ldots, P\big(X_t^{(i)}=s_3\big)\Big)$, $i=1, 2, 3$, respectively, such that
\begin{equation}  \label{fmpexample}
	\boldsymbol p_1=(0.4, 0.1, 0.1, 0.4), \, \, \, \boldsymbol p_2=(0.1, 0.4, 0.1, 0.4), \, \, \, \boldsymbol p_3=(0.1, 0.1, 0.4, 0.4).
\end{equation} 

The distance between two processes can be measured as the squared Euclidean distance between their corresponding marginal probability vectors, that is by defining $d^*\big(X_t^{(i)}, X_t^{(j)}\big)=\|\boldsymbol p_i-\boldsymbol p_j\|^2$. Based on this metric, we have
\begin{equation}\label{3mp}
	d^*\big(X_t^{(1)}, X_t^{(2)}\big)=d^*\big(X_t^{(1)}, X_t^{(3)}\big)=d^*\big(X_t^{(2)}, X_t^{(3)}\big)=0.18,
 \end{equation}
thus concluding that the three processes are equidistant. However, the underlying ordering in the set $\mathcal{S}$ suggests that process $X_t^{(1)}$ should be closer to $X_t^{(2)}$ than to $X_t^{(3)}$, since category $s_1$ is closer to $s_0$ than category $s_2$. Therefore, distance $d^*$ ignores the latent ordering and one could conclude that it is not appropriate to compare two ordinal processes.

Now, consider $d_{1, M}$ in \eqref{d1components} defined as the squared Euclidean distance between the vectors of cumulative probabilities $\boldsymbol{f}_i=\Big(P\big(X_t^{(i)}\le s_0\big), \ldots, P\big(X_t^{(i)}\leq s_2\big)\Big)$, for $i=1, 2, 3$. From \eqref{fmpexample} follows that
\begin{equation}
	\boldsymbol f_1=(0.4, 0.5, 0.6), \, \, \, \boldsymbol f_2=(0.1, 0.5, 0.6), \, \, \, \boldsymbol f_3=(0.1, 0.2, 0.6),
\end{equation} 
and the pairwise distances based on $d_{1, M}$ take the values 
\begin{equation}\label{3cp}
	d_{1, M}\big(X_t^{(1)}, X_t^{(2)}\big) = d_{1, M}\big(X_t^{(2)}, X_t^{(3)}\big) = 0.09, \, \, \, d_{1, M}\big(X_t^{(1)}, X_t^{(3)}\big)=0.18.
\end{equation}

According to distance $d_{1, M}$, the pair $(X_t^{(1)}, X_t^{(2)})$ is closer than the pair $(X_t^{(1)}, X_t^{(3)})$. Moreover, process $X_t^{(2)}$ is located at the same distance from $X_t^{(1)}$ and $X_t^{(3)}$. This is reasonable since the marginal distribution of both $X_t^{(1)}$ and $X_t^{(3)}$ can be obtained from the distribution of $X_t^{(2)}$ by transferring the same amount of probability either one step backward ($s_0$) or upward ($s_2$) from category $s_1$, respectively. In essence, cumulative probabilities allow us to better differentiate between ordinal distributions because they implicitly take into account the underlying ordering of the states. Specifically, the amount of dissimilarity is lower when the differences between marginal distributions happen at closer categories. Therefore, metric $d_{1, M}$ assigns distance values consistent with the inherent order of the range $\mathcal{S}$. 

The above example highlights the importance of considering cumulative probabilities to properly measure dissimilarity between ordinal processes. In fact, computations in \eqref{3mp} show that the use of probability mass functions can lead to misleading results when an ordinal range is considered. These arguments can be justified by means of the following proposition, which expresses the metric $d_{1, M}$ in terms of discrepancies between the probability mass functions. 
    \begin{prop}
    \label{propd1M}
    Let  $\{X_t\}_{t \in \mathbb{Z}}$ and $\{Y_t\}_{t \in \mathbb{Z}}$ be two stationary ordinal processes with range $\mathcal{S}=\{s_0, s_1, \ldots, s_n\}$ and vectors of marginal probabilities $(p_0, p_1, \ldots, p_n)$ and $(q_0, q_1, \ldots, q_n)$, respectively. Then, the distance $d_{1, M}$ between them can be written as 
\begin{equation}
	d_{1, M}\big(X_t, Y_t\big)=\sum_{i=0}^{n-1}(n-i)(p_i-q_i)^2+2\sum_{j=0}^{n-2}\sum_{k=j+1}^{n-1}(n-k)(p_j-q_j)(p_k-q_k). 
\end{equation}
\end{prop}

The proof of Proposition~\ref{propd1M} is shown in the Appendix. 

Proposition~\ref{propd1M} expresses $d_{1, M}$ as the sum of two terms. The first term involves the squared differences $(p_i-q_i)^2$ appearing in the definition of the metric $d^*$, while the second one includes the cross products $(p_j-q_j)(p_k-q_k)$. In both cases, specific weights are given to the corresponding differences. The weights are higher when marginal probabilities at lower categories are considered, i.e. discrepancies in earlier states have a larger influence in the computation of $d_{1, M}$.
Note that this property sheds light on the differences encountered between the distance computations obtained in \eqref{3mp} and \eqref{3cp}.

Previous considerations illustrate the advantage of using cumulative probabilities when measuring dissimilarity between the marginal distributions of two ordinal processes. An analogous argument could be provided when assessing dissimilarity between lagged joint distributions. In other words, the metric $d_{1, B}$ is more appropriate to evaluate dissimilarity in the ordinal setting than an analogous distance based on the probabilities $p_{ij}(l)$ in \eqref{ncp2}. We omit the theoretical considerations for the bivariate case for the sake of simplicity.  

Next, we show an interesting example involving real-world data. Let us consider the data set described in Section 8 of \cite{weiss2019distance}, which contains credit ratings according to Standard \& Poor's (S\&P) for the 27 countries of the European Union (EU) plus the United Kingdom (UK). Each country is described by means of a monthly time series with values ranging from ``D'' (worst rating) to ``AAA'' (best rating). Specifically, the whole range consists of the $n+1=23$ states $s_0, \ldots, s_{22}$, given by ``D'', ``SD'', ``R'', ``CC'', ``CCC-'', ``CCC'', ``CCC+'', ``B--'', ``B'', ``B+'', ``BB--'', ``BB'', ``BB+'', ``BBB--'', ``BBB'', ``BBB+'', ``A--'', ``A'', ``A+'', ``AA--'', ``AA'', ``AA+'' and ``AAA'', respectively. The sample period spans from January 2000 to December 2017, thus resulting serial realizations of length $T=216$. 

\begin{figure}[ht]
	\centering
	\includegraphics[width=0.7\textwidth]{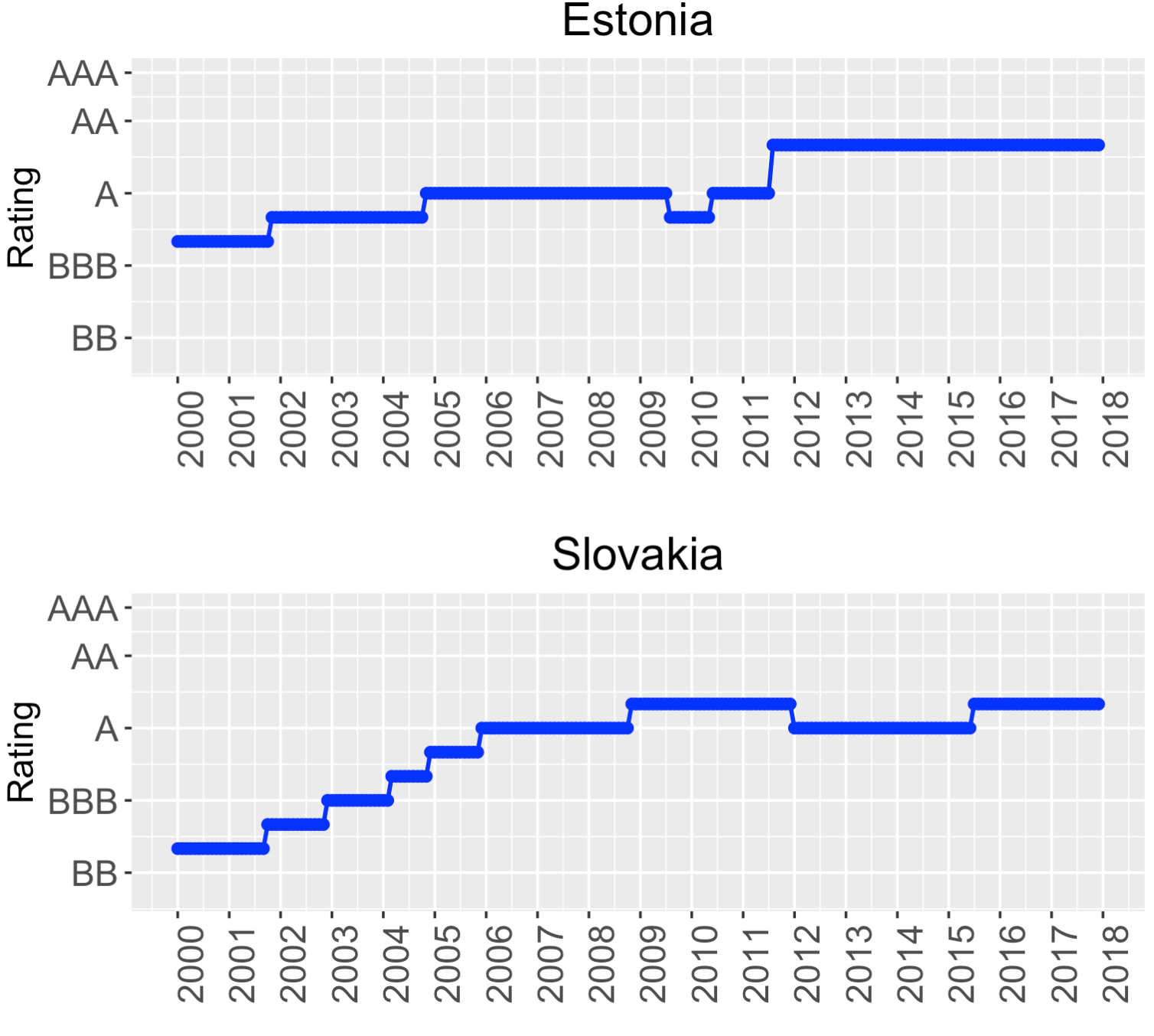}
	\caption{Monthly series of S\&P credit ratings for Estonia (top panel) and Slovakia (bottom panel).}
	\label{estoniaslovakia}
\end{figure}

Figure~\ref{estoniaslovakia} shows the time series associated with Estonia (top panel) and Slovakia (bottom panel). For a clear visualization, the $y$-axis was limited to ratings above ``B+''. It is clear from Figure \ref{estoniaslovakia} that both countries exhibit a stepwise upward pattern during the whole period, which indicates that their creditworthiness has shown a gradual improvement since the year 2000. However, the ascending trend involves different states for each one of the countries. For instance, Slovakia shows a broader range than Estonia in terms of monthly credit ratings, which leads to a higher number of different (1-step) transitions between the states. Therefore, to properly measure the distance between the serial dependence structures of both series,
one should take into consideration how far a given category is from the rest.
Note that, in this example, the distance based on the joint probabilities in \eqref{ncp2} could lead to meaningless conclusions, since the direction in which a particular transition occurs is ignored and therefore each pair of states would be treated as equidistant. 
This problem can be circumvented by employing the cumulative bivariate probabilities in \eqref{cp}, which would allow to detect the corresponding upward movements and identify a similar underlying pattern in both time series. It is worth highlighting that some of the series of the remaining countries show similar behaviours than the ones displayed in Figure \ref{estoniaslovakia}.

\section{Fuzzy clustering algorithms for ordinal time series}\label{sectionfuzzyalgorithms}

This section is devoted to introduce fuzzy clustering algorithms for ordinal series based on the proposed distances $\widehat{d}_1$ and $\widehat{d}_2$. First, a standard fuzzy $C$-medoids method relying on both metrics is presented. Next, an extension of this model is constructed by giving weights to the marginal and serial components of $\widehat{d}_1$ and $\widehat{d}_2$. The iterative solutions of the weighted models are derived. 

\subsection{A fuzzy $C$-medoids model based on the proposed dissimilarities}

Consider a set of $s$ ordinal time series, $\mathbb{S}=\{X_{T_1}^{(1)}, \ldots, X_{T_s}^{(s)}\}$, where the $i$th series has length $T_i$. We wish to perform fuzzy clustering on the elements of $\mathbb{S}$ in such a way that the series generated from similar stochastic processes are grouped together. To this aim, we propose to use fuzzy $C$-medoids clustering models based on the distances $\widehat{d}_1$ and $\widehat{d}_2$ introduced in Section~\ref{subsection2dissimilarities}. Thus, the objective is to find the subset of $\mathbb{S}$ of size $C$, $\widetilde{\mathbb{S}}=\{\widetilde{X}_t^{(1)}, \ldots, \widetilde{X}_t^{(C)}\}$, whose elements are usually referred to as medoids, and the $s \times C$ matrix of fuzzy coefficients, $\boldsymbol U=(u_{ic})$, with $i=1, \ldots, s$ and $c=1, \ldots, C$, solving the minimization problem 
\begin{equation}\label{fcm}
	\min_{\widetilde{\mathbb{S}}, \bm U}\sum_{i=1}^{s}\sum_{c=1}^{C}u_{ic}^m\widehat{d}_p(i, c),  \, \, \,  \text{ with respect to} \, \, \sum_{c=1}^{C}u_{ic}=1, \, u_{ic} \ge 0,
\end{equation}
where $\widehat{d}_p(i,c)=\widehat{d}_p\big(X_{T_i}^{(i)}, \widetilde{X}_t^{(c)}\big)$, $p=1,2$, $u_{ic} \in [0,1]$ represents the membership degree of the $i$th CTS in the $c$th cluster, and $m > 1$ is a real number, usually referred to as the fuzziness parameter, regulating the fuzziness of the partition. For $m=1$, the crisp version of the algorithm is obtained, so the solution takes the form $u_{ic}=1$ if the $i$th series pertains to cluster $c$ and $u_{ic}=0$ otherwise. As the value of $m$ increases, the boundaries between clusters get softer and the resulting partition is fuzzier. 

The constrained optimisation problem in \eqref{fcm} can be solved by means of the Lagrangian multipliers method, which leads to an iterative algorithm that alternately optimizes the membership degrees and the medoids. Specifically (see \cite{hoppner1999fuzzy}), the iterative solutions for the membership degrees are given by  
\begin{equation}\label{updatemem}
	u_{ic}=\Bigg[\sum_{c'=1}^{C}\Bigg(\frac{\widehat{d}_p(i, c)}{\widehat{d}_p(i, c')}\Bigg)^{\frac{1}{m-1}}\Bigg]^{-1},
\end{equation}
for $p=1,2$, $i=1,\ldots,s$, and $c=1,\ldots, C$. 

Once the membership degrees are obtained through \eqref{updatemem}, the $C$ series minimising the objective function in \eqref{fcm} are selected as the new medoids. Specifically, for each $c \in \{1,\ldots,C\}$, it is obtained the index $j_c$ satisfying
\begin{equation}\label{jc}
	j_c = \argmin_{1 \leq j \leq s} \sum_{i=1}^{s} u_{ic}^m \widehat{d}_p\big(X_{T_i}^{(i)}, X_{T_j}^{(j)}\big), \, \, \, p=1,2.
\end{equation}

 This two-step procedure is repeated until there is no change in the medoids or a maximum number of iterations is reached. An outline of the corresponding clustering algorithm is given in Algorithm~\ref{algorithm1}.

\begin{algorithm}[H]
	\textcolor{black}{\caption{\textcolor{black}{Fuzzy $C$-medoids algorithm based on the proposed distances. \label{algorithm1}}}}
	\begin{algorithmic}[1]
		\State \textcolor{black}{Fix $C$, $m$, \textit{max.iter} and $p \in \{1, 2\}$} 
		\State \textcolor{black}{Set $iter \, =0$}
		\State \textcolor{black}{Pick the initial medoids $\widetilde{\mathbb{S}}=\{\widetilde{X}_t^{(1)}, \ldots, \widetilde{X}_t^{(C)}\}$}
		\Repeat
		\State \textcolor{black}{Set $\widetilde{\mathbb{S}}_{\text{OLD}}=\widetilde{\mathbb{S}}$}   
		\Comment{\textcolor{black}{Store the current medoids}}
		\State \textcolor{black}{Compute $u_{ic}$, $i=1,\ldots,s$, $c=1,\ldots,C$, using (\ref{updatemem})}
		\State \textcolor{black}{For each $c \in \{1,\ldots,C\}$, determine the index $j_c \in \{1,\ldots,s\}$ using \eqref{jc}}
		\State \textbf{return} \textcolor{black}{$\widetilde{X}_t^{(c)}=X_t^{(j_c)}$, for $c=1,\ldots,C$}  
		\Comment{\textcolor{black}{Update the medoids}}
		\State \textcolor{black}{$iter \, \gets iter \, + 1$}
		\Until{ \mbox{ \textcolor{black}{$\widetilde{\mathbb{S}}_{\text{OLD}}=\widetilde{\mathbb{S}} \mbox{ or } iter \, = \, max.iter$}} } 
		\State \textbf{return} \textcolor{black}{The final fuzzy partition and the corresponding set of medoids}
	\end{algorithmic}
\end{algorithm}


\begin{rem}
\label{remAdvantages}
    \textit{Advantages of the fuzzy $C$-medoids model}. The fuzzy $C$-medoids procedure outlined in Algorithm~\ref{algorithm1} allows us to identify a set of representative OTS belonging to the original collection, the medoids, whose overall distance to all other series in the set is minimal when the membership degrees with respect to a specific cluster are considered as weights (see the computation of $j_c$ in Algorithm~\ref{algorithm1}). As observed by \cite{kaufman2009finding}, it is often desirable that the prototypes synthesising the structural information of each cluster belong to the original data set, instead of obtaining ``virtual'' prototypes, as in the case of fuzzy $C$-means-based approaches \cite{dunn1973fuzzy, bezdek2013pattern}. For instance, the original set of series could be replaced by the set of medoids for exploratory purposes, thus substantially reducing the computational complexity of subsequent data mining tasks. The fuzzy $C$-medoids algorithm also exhibits classical advantages related to the fuzzy paradigm, including ability to produce richer clustering solutions than hard methods, identifying the vague nature of the prototypes, and the possibility of dealing with time series sharing different dynamic patterns, among others. 
    \end{rem}

The behaviour of the fuzzy $C$-medoids algorithm based on the metrics $\widehat{d}_1$ and $\widehat{d}_2$ is analysed in Section \ref{subsectionedr} through an extensive simulation study. 

\subsection{A weighted fuzzy $C$-medoids model based on the proposed dissimilarities}\label{subsectionw}


Both $\widehat{d}_1$ and $\widehat{d}_2$ are formed by two terms measuring respectively the amount of discrepancy between the marginal and bivariate features of the corresponding OTS. By construction, each term receives the same weight (one) in the objective function \eqref{fcm}. However, it is reasonable to think that one of these components may have a higher influence than the other one to identify the true clustering structure. This would be the case if, for example, the prototypes present different marginal distributions but all the series exhibit a similar serial dependence structure. By contrast, the lagged joint distributions might play a more important role when time series display significant serial dependence at different lags. According to these considerations, an extension of the fuzzy $C$-medoids model outlined in Algorithm~\ref{algorithm1} is proposed by modifying the objective function in \eqref{fcm} in order to permit different weights for each one of the components of $\widehat{d}_1$ and $\widehat{d}_2$. For $p=1,2$, the weighted model is formalized by means of the minimization problem 
\begin{equation}\label{wfcm}
	\begin{dcases}
		\min_{\widetilde{\mathbb{S}}, \bm U, \beta}\sum_{i=1}^{s}\sum_{c=1}^{C}u_{ic}^m\Big[\beta^2\widehat{d}_{p, M}(i, c)+(1-\beta)^2\widehat{d}_{p, B}\big(i, c\big)\Big] \\ 
        \text{with respect to} \\
		\sum_{c=1}^{C}u_{ic}=1, \, \, u_{ic} \geq 0, \, \mbox{ for } i=1,\ldots,s, \, c=1,\ldots,C, \, \, \text{and} \, \, \beta \in [0, 1],
	\end{dcases}
\end{equation}
where $\widehat{d}_{p, M}(i, c)=\widehat{d}_{p, M}\big(X_{T_i}^{(i)}, \widetilde{X}_t^{(c)}\big)$ and $\widehat{d}_{p, B}(i, c)=\widehat{d}_{p, B}\big(X_{T_i}^{(i)}, \widetilde{X}_t^{(c)}\big)$.

The minimization problem \eqref{wfcm} involves the additional parameter $\beta$, referred to as weight, regulating the influence of each distance component in the computation of the clustering solution. Note that this approach implies that $\beta$ has to be objectively estimated via the optimization algorithm, instead of being fixed a priori by the user. It is worth highlighting that the weighted approach for fuzzy clustering of time series has been considered in several works (see e.g. \cite{coppi2010fuzzy, d2016garch, lopez2022spatial}).


The following proposition provides the iterative solutions of problem \eqref{wfcm} regarding the membership degrees and the weight $\beta$.  
    \begin{prop}
        \label{propIterSol}
    For $p=1,2$, $i = 1,\ldots,s$ and $c = 1,\ldots,C$, the optimal iterative solutions of the minimization problem \eqref{wfcm} are given by
\begin{equation}  \label{updatememw}
	u_{ic} = \Bigg[\sum_{c'=1}^{C}  \Bigg( \frac{\beta^2\widehat{d}_{p,M}(i, c)+(1-\beta)^2\widehat{d}_{p, B}(i,c)}{\beta^2\widehat{d}_{p,M}(i, c')+(1-\beta)^2\widehat{d}_{p, B}(i,c')}\Bigg)^{\frac{1}{m-1}}\Bigg]^{-1}
\end{equation}
\noindent and 
\begin{equation}\label{updatebeta}
	\beta=
	\frac{\sum_{i=1}^{s}\sum_{c=1}^{C}u_{ic}^m\widehat{d}_{p, B}(i, c)}
	{\sum_{i=1}^{s}\sum_{c=1}^{C}u_{ic}^m\Big[\widehat{d}_{p, M}(i, c)+\widehat{d}_{p, B}(i, c)\Big]}. 
\end{equation}
    \end{prop}
\vspace{0.3 cm}

The proof of Proposition~\ref{propIterSol} is presented in the Appendix. 

Proposition~\ref{propIterSol} provides a way of updating the membership matrix and the weight $\beta$. For fixed $\bm U$ and $\beta$, the medoid for the cluster $c$, denoted by $j_c$, $c=1,\ldots,C$, is obtained as solution of the minimization problem
\begin{equation}\label{jcw}
	j_c=\argmin_{1\leq j \leq s}\sum_{i=1}^{s}u_{ic}^m\Big[\beta^2\widehat{d}_{p, M}(X_{T_i}^{(i)}, X_{T_j}^{(j)}\big)+(1-\beta)^2\widehat{d}_{p, B}\big(X_{T_i}^{(i)}, X_{T_j}^{(j)}\big)\Big].
\end{equation}

The three-step procedure given by \eqref{updatememw}, \eqref{updatebeta}, and \eqref{jcw} is repeated until there is no change in the medoids anymore, or a maximum number of iterations is reached. An outline of the corresponding clustering algorithm is given in Algorithm \ref{algorithm2}.

\begin{algorithm}[H]
	\textcolor{black}{\caption{\textcolor{black}{The weighted fuzzy $C$-medoids algorithm based on the proposed distances. \label{algorithm2}}}}
	\begin{algorithmic}[1]

		\State \textcolor{black}{Fix $C$, $m$, \textit{max.iter} and $p \in \{1, 2\}$} 
		\State \textcolor{black}{Set $iter \, =0$}
		\State \textcolor{black}{Pick the initial medoids $\widetilde{\mathbb{S}}=\{\widetilde{X}_t^{(1)}, \ldots, \widetilde{X}_t^{(C)}\}$} and $\beta \in [0, 1]$
		\Repeat
		\State \textcolor{black}{Set $\widetilde{\mathbb{S}}_{\text{OLD}}=\widetilde{\mathbb{S}}$}   
		\Comment{\textcolor{black}{Store the current medoids}}
		\State \textcolor{black}{Compute $u_{ic}$, $i=1,\ldots,s$, $c=1,\ldots,C$, using (\ref{updatememw})}
		\State \textcolor{black}{Compute $\beta$ using (\ref{updatebeta})}
		\State \textcolor{black}{For each $c \in \{1,\ldots,C\}$, determine the index $j_c \in \{1,\ldots,s\}$ using \eqref{jcw}}
		\State \textbf{return} \textcolor{black}{$\widetilde{X}_t^{(c)}=X_t^{(j_c)}$, for $c=1,\ldots,C$}  
		\Comment{\textcolor{black}{Update the medoids}}
		\State \textcolor{black}{$iter \, \gets iter \, + 1$}
		\Until{ \mbox{ \textcolor{black}{$\widetilde{\mathbb{S}}_{\text{OLD}}=\widetilde{\mathbb{S}} \mbox{ or } iter \, = \, max.iter$}} } 
		\State \textbf{return} \textcolor{black}{The final partition, corresponding set of medoids, and value for $\beta$}
	\end{algorithmic}
\end{algorithm}


\begin{rem}
    \label{remWeight}
    \textit{Meaning of the weight $\beta$}. The parameter $\beta$ in Algorithm \ref{algorithm2} has an interesting statistical meaning. In particular, it attempts to mirror the heterogeneity of the total intra-cluster deviation with respect to both component distances. Specifically, the value of $\beta$ increases as long as the total intra-cluster deviation concerning the marginal component decreases (in comparison with the serial component). An analogous reasoning holds for the weight $1-\beta$. Thus, the optimisation procedure tends to give more emphasis to the component distance capable of increasing the within-cluster similarity.
    \end{rem}

The performance of the weighted fuzzy $C$-medoids algorithm based on the proposed dissimilarities is assessed in Section \ref{subsectionevaluationw} by means of several numerical experiments.  

\section{Simulation study}\label{sectionsimulationstudy}

In this section, we carry out a set of simulations with the aim of evaluating the behaviour of the proposed algorithms in different scenarios of OTS clustering. First, we describe some procedures based on alternative distances that we consider for comparison purposes. Next, we explain how the performance of the algorithms is measured along with the corresponding simulation mechanism and results. Lastly, a sensitivity analysis is carried out to analyse how the clustering accuracy changes with respect to the set of lags ($\mathcal{L}$), and a reasonable method for selecting this set is provided. 

\subsection{Alternative metrics}

To shed light on the performance of the proposed fuzzy clustering algorithms, they were compared with some other models based on alternative dissimilarities. The considered approaches are described below. 

\begin{itemize}
	\item \textit{A procedure based on the probability mass functions}. This method considers a distance defined in the same way as $\widehat{d}_1$, but replacing the estimates $\widehat{f}_i^{(k)}$ and $\widehat{f}_{ij}^{(k)}(l)$ by the probabilities $\widehat{p}_i^{(k)}$ and $\widehat{p}_{ij}^{(k)}(l)$ in \eqref{estimatesprobabilities}, respectively, $k=1,2$. The corresponding metric is called $\widehat{d}_{PMF}$. Note that $\widehat{d}_{PMF}$ is still well defined when dealing with nominal time series, although ignoring the underlying ordering. Therefore, the performance of $\widehat{d}_{PMF}$ is an essential benchmark for the proposed metric $\widehat{d}_1$, which is specifically designed to deal with ordinal series.  
	\item \textit{Autocorrelation-based clustering}. \cite{d2009autocorrelation} proposed a distance measure between real-valued time series based on the autocorrelation function. Each time series is described by means of a vector $\big(\widehat{\rho}(l_1), \ldots, \widehat{\rho}(l_L)\big)$ whose components are the estimated autocorrelations for a given set of lags. Then, the metric is defined as the squared Euclidean distance between the vectors representing two time series. We denote this dissimilarity as $\widehat{d}_{ACF}$. Note that, although $\widehat{d}_{ACF}$ is well defined only for numerical time series, the distance can be easily computed in the ordinal case by considering the associated count time series (see Section \ref{subsectionbackground}).  
	\item \textit{Quantile-based clustering}. \cite{lafuente2016clustering} introduced a clustering method using a dissimilarity based on quantile dependence. Here, each series is replaced by a feature vector containing estimates of the so-called quantile autocovariance function for several pairs of probability levels $(\tau, \tau')\in[0,1]^2$ and a fixed set of lags. The proposed metric, denoted by $\widehat{d}_{QAF}$, is defined as the squared Euclidean distance between two vector representations. As in the case of $\widehat{d}_{ACF}$, in an ordinal context, the computation of  $\widehat{d}_{QAF}$ must be based on the corresponding count time series. Several time series clustering procedures using quantile-based features have been proposed in the literature \cite{alonso2020hierarchical, lopez2021quantile, lopez2022quantile1, lopez2022quantile2}. These methods usually show a great performance when the clusters are characterised by different nonlinear structures. 
	\item \textit{Model-based approaches relying on first-order Markov chains}. \cite{pamminger2010model} proposed two methods for clustering nominal time series based on first-order Markov chains. The first one assumes the same transition matrix for all the series in a given cluster, while the second technique allows for some degree of intra-group heterogeneity by considering the Dirichlet distribution. Both methods fit finite mixtures of Markov chains by using a Bayesian approach. Although these procedures do not directly use a distance metric, for the sake of homogeneity, we are going to refer to them as $\widehat{d}_{MC}$.  
\end{itemize}

\subsection{Experimental design and results}\label{subsectionedr}

A broad simulation study was carried out to evaluate the behaviour of the fuzzy $C$-medoids algorithm based on metrics $\widehat{d}_1$ and $\widehat{d}_2$. We intended to drive the evaluation process in a way that general conclusions on the performance of both distances can be reached. To this end, two different assessment schemes were designed. The first one includes scenarios with four different groups of OTS, and is aimed at evaluating the ability of the procedures to assign high (low) memberships if a given OTS belongs (not belongs) to a given cluster. The second one consists of scenarios formed by two different groups of OTS plus one additional OTS not belonging to any of the groups. We examine again the membership degrees of the series in the two groups, but also that the isolated series is not placed in any of the clusters with a high membership. In this case, a cutoff value is used to determine whether or not a membership degree in a given group is enough to assign the OTS to that cluster. 

\subsubsection{First assessment scheme}\label{subsubsection1as}

We considered three simple scenarios consisting of four clusters represented by the same type of generating processes, denoted by $\mathcal{C}_1$, $\mathcal{C}_2$, $\mathcal{C}_3$, and $\mathcal{C}_4$. Each one of the groups contains five 6-state OTS, which gives rise to a set of 20 OTS defining the true clustering partition. We attempted to construct scenarios with a wide variety of ordinal models commonly used in practice to deal with OTS. The generating models concerning the count process $\{C_t\}_{t \in \mathbb{Z}}$ in each group are given below for each one of the scenarios.    

\vspace*{0.1cm}

\noindent \textbf{Scenario 1}. Fuzzy clustering of OTS based on binomial AR($p$) models \cite{weiss2009new}. Let $\pi \in(0, 1)$, $\rho \in \Big(\max\big\{\frac{-\pi}{1-\pi}, \frac{1-\pi}{-\pi}\big\}, 1\Big)$, $\beta=\pi(1-\rho)$, $\alpha=\beta+\rho$. Let the count process $\{C_t\}_{t \in \mathbb{Z}}$ be defined by the recursion
\begin{equation}\label{binomialar}
	C_t=\sum_{i=1}^{p}D_{t,i}\Big(\alpha \circo_{\hspace{-0.1cm}t\hspace{0.1cm}} C_{t-i} + \beta \circo_{\hspace{-0.1cm}t\hspace{0.1cm}} \big(n-C_{t-i}\big) \Big),
\end{equation}
where the $\big(D_{t,1}, \ldots, D_{t,p}\big)$ are independent variables distributed according to MULT$(1; \phi_1, \ldots, \phi_p)$, with $\phi_1+\ldots+\phi_p=1$, and $ \circo_{\hspace{-0.1cm}t\hspace{0.1cm}}$ denotes the binomial thinning operator performed at a specific time $t$. Here, the binomial thinning operator $ \circo$ applied to a count random variable $Y$ is defined by a conditional binomial distribution, $\alpha' \circo Y\sim\text{Bin}(Y,\alpha')$, where $\alpha' \in (0, 1)$. 
    The processes considered in this scenario are binomial AR(1), for clusters $\mathcal{C}_1$ and $\mathcal{C}_2$, and binomial AR(2) models for $\mathcal{C}_3$ and $\mathcal{C}_4$,  with vectors of coefficients given by
$$\begin{array}{ll}
\mathcal{C}_1:  \, \, \,  (\alpha, \beta)=(0.70, 0.20) & \qquad \mathcal{C}_3: \, \, \,  (\alpha, \beta, \phi_1, \phi_2)=(0.76, 0.06, 0.5, 0.5) \\
\mathcal{C}_2:  \, \, \,  (\alpha, \beta)=(0.72, 0.12) & \qquad \mathcal{C}_4: \, \, \, (\alpha, \beta, \phi_1, \phi_2)=(0.91, 0.01, 0.5, 0.5) 
\end{array}$$

\noindent \textbf{Scenario 2}. Fuzzy clustering of OTS based on binomial INARCH($p$) models \cite{ristic2016binomial}. Let $\beta, \alpha_1, \ldots, \alpha_p$ be real numbers such that $\beta, \beta + \sum_{i=1}^{p}\alpha_i \in (0, 1)$, and assume that the count process $\{C_t\}_{t \in \mathbb{Z}}$ satisfies 
\begin{equation}\label{binomialinarch}
	C_t |C_{t-1}, C_{t-2},\ldots \, \, \sim \, \, \text{Bin}\bigg(n, \beta + \frac{1}{p} \sum_{i=1}^{p}\alpha_iC_{t-i}\bigg).
\end{equation}

The considered processes are binomial INARCH(1) models for $\mathcal{C}_1$ and $\mathcal{C}_2$, and binomial INARCH(2) for $\mathcal{C}_3$ and $\mathcal{C}_4$, with vectors of coefficients given by
$$\begin{array}{ll}
\mathcal{C}_1:  \, \, \,  (\alpha_1, \beta)=(0.30, 0.35) & \qquad \mathcal{C}_3:  \, \, \,  (\alpha_1, \alpha_2, \beta)=(0.1, 0.1, 0.2) \\
\mathcal{C}_2:  \, \, \,  (\alpha_1, \beta)=(0.30, 0.40) & \qquad \mathcal{C}_4: \, \, \, (\alpha_1, \alpha_2, \beta)=(0.1, 0.1, 0.4) 
\end{array}$$

\noindent \textbf{Scenario 3}. Fuzzy clustering of ordinal logit AR(1) models (see Examples 7.4.6 and 7.4.8 in \cite{weiss2018introduction}). Let $\{C_t\}_{t \in \mathbb{Z}}$ be a count process with range $\{0, 1, \ldots,n\}$, and denote by $\{\boldsymbol Y_t=(Y_{t,0}, \ldots, Y_{t, n})^\top\}_{t \in \mathbb{Z}}$ its binarization (i.e., $C_t=k$ if and only if $Y_{t,k}=1$ and $Y_{t,k'}=0$, $k' \ne k$) and by $\{\boldsymbol Y_t^*=(Y_{t,0}, \ldots, Y_{t, n-1})^\top\}_{t \in \mathbb{Z}}$ its reduced binarization. Let $\{Q_t\}_{t \in \mathbb{Z}}$ be the process formed by independent variables following a standard logistic distribution and assume that 
\begin{equation}\label{logitar}
	C_t=j \, \, \, \, \,  \text{if and only if} \, \, \, \, \, Q_t-\boldsymbol\alpha^\intercal{\boldsymbol {Y}_t^*}\boldsymbol \in [\eta_{j-1}, \eta_j). 
\end{equation}

\noindent Here, $\boldsymbol \alpha=(\alpha_1, \ldots, \alpha_n)^\top \in \mathbb{R}^n$, and $-\infty=\eta_{-1}<\eta_0<\ldots<\eta_{n-1}<\eta_n=+\infty$ are threshold parameters which can be represented by means of the vector $\boldsymbol \eta=(\eta_0, \ldots, \eta_{n-1})$. The considered processes are four 6-state ordinal logit AR(1) models with vectors of coefficients given by $ \boldsymbol \eta=(-2, -1, 0, 1, 2)$ and
$$\begin{array}{ll}
\mathcal{C}_1:  \, \, \,  \boldsymbol{\alpha}=(0.4, -0.8, 1.2, 1.6, 2)^\top & \qquad \mathcal{C}_3:  \, \, \,  \boldsymbol{\alpha}=(0.8, -1.6, 2.4, 3.2, 4)^\top \\
\mathcal{C}_2:  \, \, \,  \boldsymbol{\alpha}=(0.6, -1.2, 1.8, 2.4, 3)^\top & \qquad \mathcal{C}_4: \, \, \, \boldsymbol{\alpha}=(1, -2, 3, 4, 5)^\top 
\end{array}$$

As a preliminary step, metric two-dimensional scaling (2DS) based on both  $\widehat{d}_1$ and $\widehat{d}_2$ was carried out to gain insight about the capability of these metrics to discriminate between the underlying groups. Given a distance matrix $\bm D=(D_{ij})_{1 \leq i,j \leq s}$, a 2DS finds the points $\{(a_i, b_i), i = 1,\ldots, s\}$ minimizing the loss function called stress given by  
\begin{equation}\label{stress}
	\sqrt{\frac{\sum_{i \ne j=1}^{s}(\norm{(a_i, b_i)-(a_j, b_j)}-D_{ij})^2}{\sum_{i \ne j = 1}^{s}D_{ij}^2}}
\end{equation}
Thus, the goal is to represent the distances $D_{ij}$ in terms of Euclidean distances into a 2-dimensional space so that the original distances are preserved as well as possible. The lower the value of the stress function, the more reliable the 2DS configuration. This way, a 2DS plot provides a valuable visual representation of how the elements are located with respect to each other according to the original distances.

\begin{figure}[ht]
	\centering
	\includegraphics[width=1\textwidth]{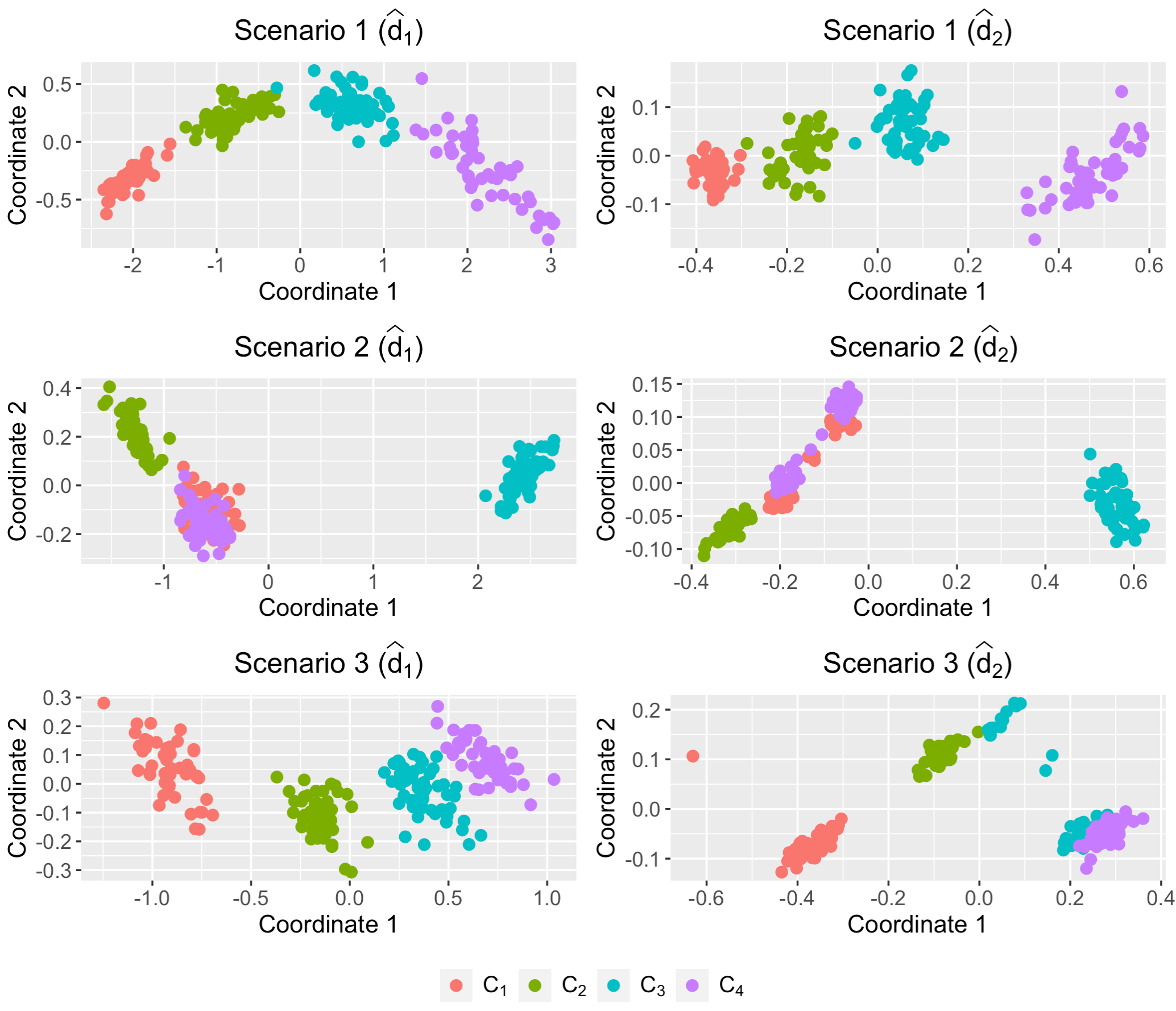}
	\caption{Two-dimensional scaling planes based on distances $\widehat{d}_1$ and $\widehat{d}_2$ between simulated time series in Scenarios 1, 2 and 3. The series length is $T = 600$.}
	\label{2dsplot}
\end{figure}

To obtain informative 2DS plots, 50 OTS of length $T = 600$ from each generating model were simulated for each scenario. 
The 2DS was carried out for each set of 200 CTS by computing the pairwise dissimilarity matrices based on $\widehat{d}_1$ and $\widehat{d}_2$. We considered the set of lags $\mathcal{L}=\{1, 2\}$ in Scenarios 1 and 2 and $\mathcal{L}=\{1\}$ in Scenario 3. The resulting plots are shown in Figure~\ref{2dsplot}, where a different colour was used for each generating process. It is worth highlighting that the $R^2$ value associated with the scaling is above 0.85 in all cases, thus concluding that the graphs in Figure~\ref{2dsplot} provide an accurate picture of the underlying representations according to both metrics. 

The reduced bivariate spaces in Figure~\ref{2dsplot} show different configurations. In Scenario 1, the metrics seem able to detect the underlying clustering partition, which is expected, since the four generating processes in this scenario are clearly dissimilar. In Scenario 2, both distances place cluster $\mathcal{C}_3$ quite far from the rest, which is reasonable in view of the coefficients defining the models. In addition, there is a high degree of overlap between clusters $\mathcal{C}_1$ and $\mathcal{C}_4$. This is logical, since the generating models of both clusters are quite similar in terms of marginal distributions and serial dependence (the properties of a binomial INARCH($p$) model can be seen in \cite{ristic2016binomial}). Concerning Scenario 3, $\widehat{d}_1$ and $\widehat{d}_2$ produce plots with substantially different structures. While $\widehat{d}_1$ is capable of successfully identifying the four groups, $\widehat{d}_2$ struggles to separate $\mathcal{C}_3$ and $\mathcal{C}_4$. This is because some of the features employed by $\widehat{d}_2$ take similar values for the processes behind these clusters (e.g., $\widehat{\text{disp}}_{d_{\text{o},1}}$ or $\widehat{\kappa}_{d_{\text{o},1}}(1)$). In sum, the plots in Figure~\ref{2dsplot} suggest that $\widehat{d}_1$ and $\widehat{d}_2$ have 
different levels of difficulty to identify the true clustering partition. 

The simulation study was carried out as follows. For each scenario, 5 OTS of length $T \in \{200, 600\}$ were generated from each process in order to execute the clustering algorithms twice and examine the effect of the series length. In all cases, the range of the count process $\{C_t\}_{t \in \mathbb{Z}}$ was set to $\{0, 1, \ldots, 5\}$, giving rise to ordinal realizations with range $\{s_0, s_1, \ldots, s_5\}$. Several values of the fuzziness parameter $m$ were considered, namely $m \in \{1.2, 1.4, 1.6, 1.8, 2\}$. The problem of selecting a proper value for $m$ has been extensively addressed in the literature, although there seems to be no consensus about the optimal way of choosing this parameter (see the discussion in Section 3.1.6 of \cite{maharaj2011fuzzy}). When $m=1$, the hard version of the fuzzy $C$-medoids algorithm is obtained, while excessively large values of $m$ result in a partition with all memberships close to $1/C$, thus having a large degree of overlap between groups. As a consequence, selecting these values for $m$ is not recommended \cite{arabie1981overlapping}. Moreover, in the context of time series clustering, several works consider a grid of values for $m$ similar to our choice \cite{d2009autocorrelation,vilar2018quantile,lopez2021quantile}. 

Given a scenario and fixed values for $m$ and $T$, 200 simulations were executed. In each trial, the fuzzy $C$-medoids algorithm based on $\widehat{d}_1$, $\widehat{d}_2$, $\widehat{d}_{PMF}$, $\widehat{d}_{ACF}$ and $\widehat{d}_{QAF}$ was applied with each value of $m$ as input. The number of clusters was set to $C=4$. The collection of lags was $\mathcal{L}=\{1, 2\}$ in Scenarios~1 and 2 and $\mathcal{L}=\{1\}$ in Scenario~3, thus considering the maximum number of lags at each scenario. The same lags were used to obtain the alternative dissimilarities, e.g. $\widehat{d}_{ACF}$ employed the two first autocorrelations in Scenarios 1 and 2. Concerning the distance $\widehat{d}_{QAF}$, several sets of probability levels were independently considered for its computation, namely $\mathcal{T}_1=\{0.1, 0.5, 0.9\}$, $\mathcal{T}_2=\{0.3, 0.5, 0.7\}$ and $\mathcal{T}_3=\{0.4, 0.8\}$. Clustering acuraccy was assessed using the fuzzy extensions of the Adjusted Rand Index (ARIF) and the Jaccard Index (JIF) introduced by \cite{campello2007fuzzy}. Both indexes are obtained by reformulating the original ones in terms of the fuzzy set theory, which allows to compare the true (hard) partition with a experimental fuzzy partition. ARIF and JIF take values in the intervals $[-1, 1]$ and $[0, 1]$, respectively, with values closer to 1 indicating a more accurate clustering solution. Note that the Bayesian method of cite{pamminger2010model} ($\widehat{d}_{MC}$) can be seen as a soft clustering procedure by treating the posterior probabilities as membership degrees. However, this approach does not involve the fuzziness parameter $m$ and, consequently, our results for $\widehat{d}_{MC}$ only include one value of ARIF and JIF for a given series length.


The average values of ARIF and JIF based on the 200 simulation trials are shown in Table~\ref{tablescenarios123}, for all metrics except for $\widehat{d}_{MC}$, and in Table~\ref{tablescenarios123dmc}, for $\widehat{d}_{MC}$. Concerning $\widehat{d}_{QAF}$, it is important to notice that only the highest ARIF and JIF are presented, regardless of the employed probability levels. From Table~\ref{tablescenarios123}, it is concluded that all distances decrease their performance when increasing the value of $m$. This is reasonable and expected, since larger values of $m$ produce a smoother boundary between the four well-separated clusters, thus making the classification fuzzier and decreasing the value of ARIF and JIF. 
\begin{table}[!ht]
	\centering
	\resizebox{12cm}{!}{\begin{tabular}{ccccccc|ccccc}  \hline   &    &  &  & ARIF  &  &  &  &  & JIF   &  \\  \hline
	Scenario 1&  & & & & & & & & & &  \\  
	& & $\widehat{d}_1$  &  $\widehat{d}_2$ &  $\widehat{d}_{PMF}$ &   $\widehat{d}_{ACF}$ &  $\widehat{d}_{QAF}$ & $\widehat{d}_1$  &   $\widehat{d}_2$ &  $\widehat{d}_{PMF}$ &   $\widehat{d}_{ACF}$ &  $\widehat{d}_{QAF}$       \\  \hline 
	$T=200$  & $m=1.2$ & 0.67 & \textbf{0.76} & 0.60 & 0.54 & 0.33  & 0.60 & \textbf{0.69} & 0.54 & 0.48 & 0.33  \\  
	& $m=1.4$ & 0.57 & \textbf{0.61} & 0.46 & 0.47 & 0.28  & 0.52 & \textbf{0.55} & 0.43 & 0.43 & 0.31  \\    
	& $m=1.6$ & \textbf{0.49} & \textbf{0.49} & 0.34 & 0.39 & 0.22  & \textbf{0.46} & 0.45& 0.35 & 0.38 & 0.27 \\
	& $m=1.8$ & 0.40 & \textbf{0.41} & 0.29 & 0.32 & 0.20  & 0.39 & \textbf{0.40} & 0.32 & 0.34 & 0.27  \\
	& $m=2.0$ & \textbf{0.34} & 0.33 & 0.23 & 0.28 & 0.16  & \textbf{0.35} & \textbf{0.35} & 0.29 & 0.31 & 0.24  \\ \hline 
	$T=600$  & $m=1.2$ & \textbf{0.92} & \textbf{0.92} & 0.91 & 0.83 & 0.59  & \textbf{0.89} & 0.88 & 0.87 & 0.78 & 0.53  \\ 
	& $m=1.4$& \textbf{0.85} & 0.83 & 0.77 & 0.73 & 0.52 & \textbf{0.80} & 0.78 & 0.70 & 0.66 & 0.47 \\
	& $m=1.6$ & \textbf{0.75} & 0.73 & 0.64 & 0.64 & 0.46  & \textbf{0.68} & 0.67 & 0.57 & 0.58 & 0.43  \\
	& $m=1.8$ & \textbf{0.65} & 0.61 & 0.51 & 0.55 & 0.37  & \textbf{0.58} & 0.55 & 0.47 & 0.50 & 0.37  \\
	& $m=2.0$ & \textbf{0.56 }& 0.52 & 0.42 & 0.46 & 0.31  & \textbf{0.51} & 0.48 & 0.41 & 0.43 & 0.34  \\ \hline 
	Scenario 2&  & & & & & & & & & &  \\  
	& & $\widehat{d}_1$  &  $\widehat{d}_2$ &  $\widehat{d}_{PMF}$ &   $\widehat{d}_{ACF}$ &  $\widehat{d}_{QAF}$ & $\widehat{d}_1$  &   $\widehat{d}_2$ &  $\widehat{d}_{PMF}$ &   $\widehat{d}_{ACF}$ &  $\widehat{d}_{QAF}$       \\  \hline
	$T=200$  & $m=1.2$ & 0.57 &\textbf{0.60} & 0.49 & 0.21 & 0.37 & 0.51 & \textbf{0.54} & 0.46 & 0.26 & 0.36 \\    
	& $m=1.4$ & \textbf{0.54} & 0.53 & 0.40 & 0.17 & 0.32 & \textbf{0.49} & 0.48 & 0.39 & 0.25 & 0.33 \\    
	& $m=1.6$ & \textbf{0.48} & 0.45 & 0.32 & 0.14 & 0.27 & \textbf{0.45} & 0.43 & 0.35 & 0.24 & 0.30 \\    
	& $m=1.8$ &\textbf{0.41} & 0.38 & 0.26 & 0.12 & 0.22 & \textbf{0.40} & 0.38 & 0.31 & 0.23 & 0.28 \\
	& $m=2.0$ & \textbf{0.35} & 0.32 & 0.21 & 0.10 & 0.19 & \textbf{0.36} & 0.34 & 0.28 & 0.22 & 0.26 \\   \hline
	$T=600$  & $m=1.2$ & 0.69 &\textbf{0.71} & 0.63 & 0.30 & 0.58 & 0.62 & \textbf{0.64} & 0.37 & 0.31 & 0.51 \\    
	& $m=1.4$ & 0.64 & \textbf{0.67} & 0.53 & 0.26 & 0.51 & 0.58 & \textbf{0.60} & 0.49 & 0.29 & 0.46 \\ 
	& $m=1.6$ &\textbf{0.59} & \textbf{0.59} & 0.45 & 0.23 & 0.45 & \textbf{0.54} & 0.53 & 0.43 & 0.28 & 0.42 \\  
	& $m=1.8$ & \textbf{0.52} & 0.51 & 0.37 & 0.18 & 0.37 & \textbf{0.48} & 0.47 & 0.38 & 0.26 & 0.37 \\  
	& $m=2.0$ & \textbf{0.46} & 0.44 & 0.31 & 0.16 & 0.33 & \textbf{0.44} & 0.42 & 0.34 & 0.25 & 0.34  \\   \hline 
	Scenario 3&  & & & & & & & & & &  \\  
	& & $\widehat{d}_1$  &  $\widehat{d}_2$ &  $\widehat{d}_{PMF}$ &   $\widehat{d}_{ACF}$ &  $\widehat{d}_{QAF}$ & $\widehat{d}_1$  &   $\widehat{d}_2$ &  $\widehat{d}_{PMF}$ &   $\widehat{d}_{ACF}$ &  $\widehat{d}_{QAF}$       \\  \hline
	$T=200$  & $m=1.2$ & \textbf{0.65} & 0.62 & 0.61 & 0.15 & 0.53 & \textbf{0.59} & 0.55 & 0.55 & 0.21 & 0.48 \\    
	& $m=1.4$ & 0.51 &\textbf{0.53} & 0.40 & 0.14 & 0.43 & 0.47 & \textbf{0.48} & 0.39 & 0.21 & 0.41 \\    
	& $m=1.6$ & 0.40 & \textbf{0.45} & 0.28 & 0.13 & 0.35 & 0.39 & \textbf{0.42} & 0.32 & 0.21 & 0.36 \\    
	& $m=1.8$ &0.32 & \textbf{0.36} & 0.22 & 0.12 & 0.28 & 0.34 & \textbf{0.36} & 0.28 & 0.21 & 0.31 \\
	& $m=2.0$ & 0.26 & \textbf{0.30} & 0.18 & 0.11 & 0.23 & 0.30 & \textbf{0.33} & 0.26 & 0.21 & 0.29 \\   \hline
	$T=600$  & $m=1.2$ & \textbf{0.91} & 0.78 & 0.90 & 0.30 & 0.70 & \textbf{0.88} & 0.72 & 0.85 & 0.30 & 0.63 \\    
	& $m=1.4$ & \textbf{0.78} & 0.72 & 0.69 & 0.28 & 0.62 & \textbf{0.72} & 0.66 & 0.62 & 0.30 & 0.56 \\ 
	& $m=1.6$ & \textbf{0.64} & \textbf{0.64} & 0.52 & 0.25 & 0.53 & \textbf{0.58} & 0.57 & 0.47 & 0.28 & 0.48 \\  
	& $m=1.8$ & 0.52 & \textbf{0.55} & 0.40 & 0.23 & 0.45 & 0.48 & \textbf{0.50} & 0.39 & 0.28 & 0.42 \\  
	& $m=2.0$ &0.43 & \textbf{0.47} & 0.33 & 0.20 & 0.38 & 0.41 & \textbf{0.44} & 0.34 & 0.26 & 0.37 \\    \hline\end{tabular}}
	\caption{Average values of ARIF and JIF obtained by the fuzzy $C$-medoids clustering algorithm based on several dissimilarities. Scenarios 1, 2 and 3. For each value of $m$ and $T$, the best result is shown in bold.}
	\label{tablescenarios123}
\end{table}

\begin{table}[!ht]
		\centering 
		\begin{tabular}{ccc|cc} \hline 
			&   \multicolumn{2}{c|}{ARIF}          &       \multicolumn{2}{c}{JIF}    \\   
			& $T=200$ &       $T=600$ & $T=200$ &       $T=600$ \\ \hline 
			Scenario 1 &   0.61       &   0.61   & 0.56  &    0.56              \\
			Scenario 2 &   0.41      &  0.39   &  0.41   & 0.40                  \\
			Scenario 3 &  0.65    &   0.66    &   0.59  &      0.61          \\ \hline 
		\end{tabular}
		\caption{Average values of ARIF and JIF obtained by the fuzzy $C$-medoids clustering algorithm based on $\widehat{d}_{MC}$. Scenarios 1, 2 and 3.}
		\label{tablescenarios123dmc}
	\end{table}

In Scenario~1, $\widehat{d}_1$ and $\widehat{d}_2$ show the best performance regardless of $m$ and $T$, with similar average values for both clustering quality indices. While the quantile-based distance $\widehat{d}_{QAF}$ displays the worst results in this scenario, $\widehat{d}_{PMF}$ and $\widehat{d}_{ACF}$ also exhibit a high clustering effectiveness. Indeed, a suitable behaviour of $\widehat{d}_{ACF}$ is here expected because of the generating processes in Scenario~1 have very different autocorrelations (note that e.g. the lag-1 autocorrelation for a binomial AR(1) process is $\alpha-\beta$). The proposed distances $\widehat{d}_1$ and $\widehat{d}_2$ attain the best average scores in Scenario~2, significantly outperforming the remaining metrics in most of the considered settings. However, their performance decreases with respect to Scenario 1. This is coherent with the 2DS plots in Figure~\ref{2dsplot}, where both metrics seem to clearly identify the true clustering structure in Scenario~1 while struggling to distinguish between clusters $\mathcal{C}_1$ and $\mathcal{C}_4$ in Scenario~2. Metrics $\widehat{d}_1$ and $\widehat{d}_2$ are also the best-performing ones in Scenario~3. The autocorrelation-based metric $\widehat{d}_{ACF}$ produces very inaccurate clustering partitions in Scenarios~2 and 3, which indicates a limited ability of the autocorrelation function to discriminate between the generating processes considered in these scenarios. Furthemore, $\widehat{d}_{QAF}$ shows always a worse behaviour than the proposed distances, thus suggesting that the treatment of OTS as count time series is not advantageous for clustering purposes. As expected, all dissimilarities improve their performance when increasing the series length, although this is generally less pronounced in Scenario~2. 

According to Table~\ref{tablescenarios123dmc}, the Bayesian clustering approach ($\widehat{d}_{MC}$) attains moderate scores in the three scenarios, but its performance does not improve when increasing the series length. As $\widehat{d}_{MC}$ does not require the fuzziness parameter $m$, a direct comparison with the results in Table~\ref{tablescenarios123} is not possible. However, since the considered scenarios are formed by well-defined clusters (i.e. the underlying clustering structure is a hard partition), it is reasonable to compare the partitions based on $\widehat{d}_{MC}$ with the ones generated by the remaining techniques when $m=1.2$ (or even when $m$ is lower than 1.2). Thus, one could state that the proposed metrics $\widehat{d}_1$ and $\widehat{d}_2$ significantly outperform the Bayesian approach in most cases. 


In order to provide a more comprehensive evaluation of the proposed clustering methods, we designed two more challenging setups, Scenarios~4 and 5, where the complexity of the original experiments is increased.

\vspace*{0.1cm}

\noindent \textbf{Scenario 4}. It consists of six clusters, $\mathcal{C}_1, \mathcal{C}_2, \ldots, \mathcal{C}_6$, such that $\mathcal{C}_1$ and $\mathcal{C}_2$ are defined as the first and second clusters in Scenario~1, respectively, $\mathcal{C}_3$ is defined as the last cluster in Scenario~3, and $\mathcal{C}_4$, $\mathcal{C}_5$ and $\mathcal{C}_6$ are binomial INARCH(3) models (see \eqref{binomialinarch}) with vectors of coefficients given by 
$$\begin{array}{l}
\mathcal{C}_4:  \, \, \,  (\alpha_1, \alpha_2, \alpha_3, \beta)=(0.1, 0.3, 0.2, 0.2) \\
\mathcal{C}_5:  \, \, \,  (\alpha_1, \alpha_2, \alpha_3, \beta)=(0.1, 0.2, 0.3, 0.2) \\
\mathcal{C}_6:  \, \, \,  (\alpha_1, \alpha_2, \alpha_3, \beta)=(0.1, 0.25, 0.25, 0.2)
\end{array}$$

Simulations in Scenario~4 were carried out by setting $C=6$ and $\mathcal{L}=\{1, 2, 3\}$, but selecting the remaining inputs (series length, series per cluster,\ldots) in the same manner as in Scenarios~1--3. Compared to the above scenarios, Scenario~4 is clearly more complex: (i) there are a larger number of clusters, (ii) three different types of ordinal processes, and (iii) the processes behind $\mathcal{C}_4$, $\mathcal{C}_5$ and $\mathcal{C}_6$ have identical marginal and one-lagged bivariate distributions, thus the series in these clusters can be well-located only by analysing higher-order dependencies. 


However, Scenario~4 still considers five series per cluster, a range with six categories ($n=5$), and $T\in\{200, 600\}$. In order to assess the performance of the different methods when varying the value of these parameters, we consider a second additional setup as described below.

\vspace*{0.1cm}

\noindent \textbf{Scenario 5}. The following random mechanism is incorporated into Scenario~4. At each simulation trial, the value of $n$ defining the range $\{s_0, \ldots, s_n\}$, the number of series in the $i$th cluster, $i=1,2,\ldots,6$, and the length of each series, are randomly selected with equiprobability from the sets $\{1, 2, \ldots, 10\}$, $\{2, 3, \ldots, 10\}$ and $\{100, 200, \ldots, 500\}$, respectively.

In sum, Scenario~5 defines a challenging setting, which inherits the complexity of Scenario~4 besides giving rise to instances with unequal series lengths and cluster sizes. Note that a different true partition must be considered at each trial to compute the values of ARIF and JIF.



The average results for these new scenarios are provided in Tables~\ref{tablescenarios45} and \ref{tablescenarios45dmc}. In Scenario~4, $\widehat{d}_{QAF}$ attains the worst average scores for all $m$ and $T$. In contrast, the proposed metrics yield again the best results, slightly improving the scores based on $\widehat{d}_{ACF}$ and somewhat more sharply the ones obtained by $\widehat{d}_{PMF}$, specially for large values of $m$. Overall, all the dissimilarities decrease their performance with respect to Scenarios~1, 2 and 3, which is reasonable due to the higher degree of complexity in Scenario~4. Average scores in Scenario~5 lead to similar conclusions. The Bayesian procedure $\widehat{d}_{MC}$ obtains moderate scores in Scenario~4, but displays a very poor performance in Scenario~5, where it gets negatively affected by the high level of variability of this scenario. 

\begin{table}[!ht]
	\centering
		\resizebox{12cm}{!}{\begin{tabular}{ccccccc|ccccc}  \hline   &    &  &  & ARIF  &  &  &  &  & JIF   &  \\  \hline
				Scenario 4	&  & & & & & & & & & &  \\  
				& & $\widehat{d}_1$  &  $\widehat{d}_2$ &  $\widehat{d}_{PMF}$ &   $\widehat{d}_{ACF}$ &  $\widehat{d}_{QAF}$ & $\widehat{d}_1$  &   $\widehat{d}_2$ &  $\widehat{d}_{PMF}$ &   $\widehat{d}_{ACF}$ &  $\widehat{d}_{QAF}$       \\  \hline 
				$T=200$	  & $m=1.2$ & 0.51 & \textbf{0.53} & 0.46 & 0.46 & 0.30  & 0.42 & \textbf{0.44} & 0.39 & 0.38 & 0.26  \\
				& $m=1.4$ & \textbf{0.45} & 0.44 & 0.35 & 0.39 & 0.24 & \textbf{0.38} & \textbf{0.38} & 0.32 & 0.34 & 0.23  \\    
				& $m=1.6$ & \textbf{0.37} & 0.35 & 0.26 & 0.32 & 0.20  & \textbf{0.33} & 0.31 & 0.26 & 0.29 & 0.21   \\    
				& $m=1.8$ & \textbf{0.30} & 0.28 & 0.20 & 0.25 & 0.16  & \textbf{0.28} & 0.27 & 0.23 & 0.25 & 0.19   \\
				& $m=2.0$ & \textbf{0.24} & 0.23 & 0.16 & 0.21 & 0.13  & \textbf{0.25} & 0.24 & 0.20 & 0.23 & 0.18  \\   \hline
				$T=600$	  & $m=1.2$ & \textbf{0.59} & \textbf{0.59} & 0.58 & 0.58 & 0.43  & \textbf{0.49} & \textbf{0.49} & \textbf{0.49} & \textbf{0.49} & 0.36  \\    
				& $m=1.4$ & \textbf{0.57} & 0.54 & 0.49 & 0.53 & 0.37  & \textbf{0.48} & 0.45 & 0.42 & 0.44 & 0.32  \\
				& $m=1.6$ & \textbf{0.50} & 0.46 & 0.39 & 0.45 & 0.31  & \textbf{0.42} & 0.39 & 0.35 & 0.38 & 0.28  \\
				& $m=1.8$ & \textbf{0.42} & 0.39 & 0.31 & 0.38 & 0.26  & \textbf{0.36} & 0.34 & 0.30 & 0.33 & 0.25  \\
				& $m=2.0$ & \textbf{0.35} & 0.33 & 0.25 & 0.31 & 0.22  & \textbf{0.31} & 0.30 & 0.26 & 0.29 & 0.23  \\ \hline
				Scenario 5	&  & & & & & & & & & &  \\  
				& & $\widehat{d}_1$  &  $\widehat{d}_2$ &  $\widehat{d}_{PMF}$ &   $\widehat{d}_{ACF}$ &  $\widehat{d}_{QAF}$ & $\widehat{d}_1$  &   $\widehat{d}_2$ &  $\widehat{d}_{PMF}$ &   $\widehat{d}_{ACF}$ &  $\widehat{d}_{QAF}$       \\  \hline 
				Variable $T$	  & $m=1.2$ & 0.52 & \textbf{0.56} & 0.51 & 0.50 & 0.37  & 0.45 & \textbf{0.49} & 0.45 & 0.43 & 0.33  \\ 
				& $m=1.4$ & \textbf{0.49} & \textbf{0.49} & 0.41 & 0.43 & 0.29  & \textbf{0.44} & 0.43 & 0.38 & 0.39 & 0.29  \\    
				& $m=1.6$ & \textbf{0.43} & 0.41 & 0.33 & 0.37 & 0.26  & \textbf{0.39} & 0.38 & 0.33 & 0.35 & 0.27  \\    
				& $m=1.8$ & \textbf{0.36} & 0.33 & 0.26 & 0.30 & 0.20  & \textbf{0.35} & 0.33 & 0.28 & 0.30 & 0.24  \\
				& $m=2.0$ & \textbf{0.30} & 0.27 & 0.21 & 0.25 & 0.17 & \textbf{0.31} & 0.29 & 0.26 & 0.27 & 0.22  \\  \hline
		\end{tabular}} 
		\caption{Average values of ARIF and JIF obtained by the fuzzy $C$-medoids clustering algorithm based on several dissimilarities. Scenarios 4 and 5. For each value of $m$ and $T$, the best result is shown in bold.}
		\label{tablescenarios45}
\end{table}

\begin{table}[!ht]
		\centering 
		\begin{tabular}{lcc|cc} \hline 
			&   \multicolumn{2}{c|}{ARIF}          &       \multicolumn{2}{c}{JIF}    \\   \hline
			Scenario 4    & $T=200$ &      $T=600$  & $T=200$ &      $T=600$ \\ 
			  &   0.447      &   0.434  &    0.384  &    0.374             \\ \hline
			Scenario 5 &  \multicolumn{2}{c|}{Variable $T$} & 
   \multicolumn{2}{c}{Variable $T$} \\
    & \multicolumn{2}{c|}{0.160}    &   \multicolumn{2}{c}{0.230}        \\ \hline 
		\end{tabular}
		\caption{Average values of ARIF and JIF obtained by the fuzzy $C$-medoids clustering algorithm based on $\widehat{d}_{MC}$. Scenarios 4 and 5.}
		\label{tablescenarios45dmc}
\end{table}

Overall, the previous analyses showed the great performance of $\widehat{d}_1$ and $\widehat{d}_2$ to perform clustering of OTS when the true partition is formed by well-separated clusters. The superiority of both metrics with respect to distances ignoring the ordinal nature of the series ($\widehat{d}_{PMF}$ and $\widehat{d}_{MC}$) and classical metrics in clustering of real-valued time series ($\widehat{d}_{ACF}$ and $\widehat{d}_{QAF}$) was corroborated in scenarios characterized by well-known types of ordinal processes and different degrees of complexity. This highlights the importance of constructing dissimilarities specifically designed to deal with ordinal series.

\subsubsection{Second assessment scheme}\label{subsubsection2as}

A second simulation experiment was conducted to analyze the effect of isolated series, whose presence introduces certain degree of ambiguity and increases the fuzzy nature of the clustering task. Two new scenarios consisting of two well-separated clusters of 5 OTS each and a single isolated series arising from a different process are defined as follows.

\vspace*{0.1cm}

\noindent \textbf{Scenario 6}. A set of 11 OTS, where five series (cluster $\mathcal{C}_1$) are generated from a binomial AR(1) process with coefficients $(\alpha, \beta)=(0.52, 0.12)$, five series (cluster $\mathcal{C}_2$) come from a binomial AR(2) process with coefficients $(\alpha, \beta, \phi_1, \phi_2)=(0.42, 0.07, 0.1, 0.9)$, and one isolated series is generated from an ordinal logit AR(1) model with vectors of coefficients given by $\boldsymbol \eta=(-2, -1, 0, 1, 2)$ and $\boldsymbol \alpha=(0.5, -1, 1.5, 2, 2.5)^\top$.

\vspace*{0.1cm}

\noindent \textbf{Scenario 7}. Defined in the same way as Scenario~6, but with different generating models for clusters $\mathcal{C}_1$ and $\mathcal{C}_2$. Here, $\mathcal{C}_1$ and $\mathcal{C}_2$ are formed by OTS generated from binomial INARCH(2) models with vectors of coefficients $(\alpha_1, \alpha_2, \beta)=(0.1, 0.1, 0.1)$ and $(\alpha_1, \alpha_2, \beta)=(0.5, 0.1, 0.1)$, respectively. 

\vspace*{0.1cm}


The values for $n$, $T$, and the number of simulation trials were fixed as in Scenarios~1--3. The number of clusters and the collection of lags were set to $C=2$ and $\mathcal{L}=\{1, 2\}$, respectively. Assessment was performed in a different way. We computed the proportion of times that: the five series from  $\mathcal{C}_1$ grouped together in one group, the five series from $\mathcal{C}_2$ clustered together in another group, and the isolated series had a relatively high membership degree with respect to each of the groups. To this aim, a cutoff point must be determined to conclude when a series is assigned to a specific cluster. We decided to use the cutoff value of 0.7, i.e. the $i$th OTS was placed into the $c$th cluster if $u_{ic} > 0.7$. On the contrary, a time series was considered to simultaneously belong to both clusters if its membership degrees were both below 0.7. The use of a cutoff value to assess fuzzy clustering algorithms has already been considered in prior works \cite{maharaj2011fuzzy, d2012wavelets, lopez2022quantile1} (arguments for this choice are given in \cite{maharaj2011fuzzy}).

\begin{figure}[ht]
		\centering
		\includegraphics[width=0.8\textwidth]{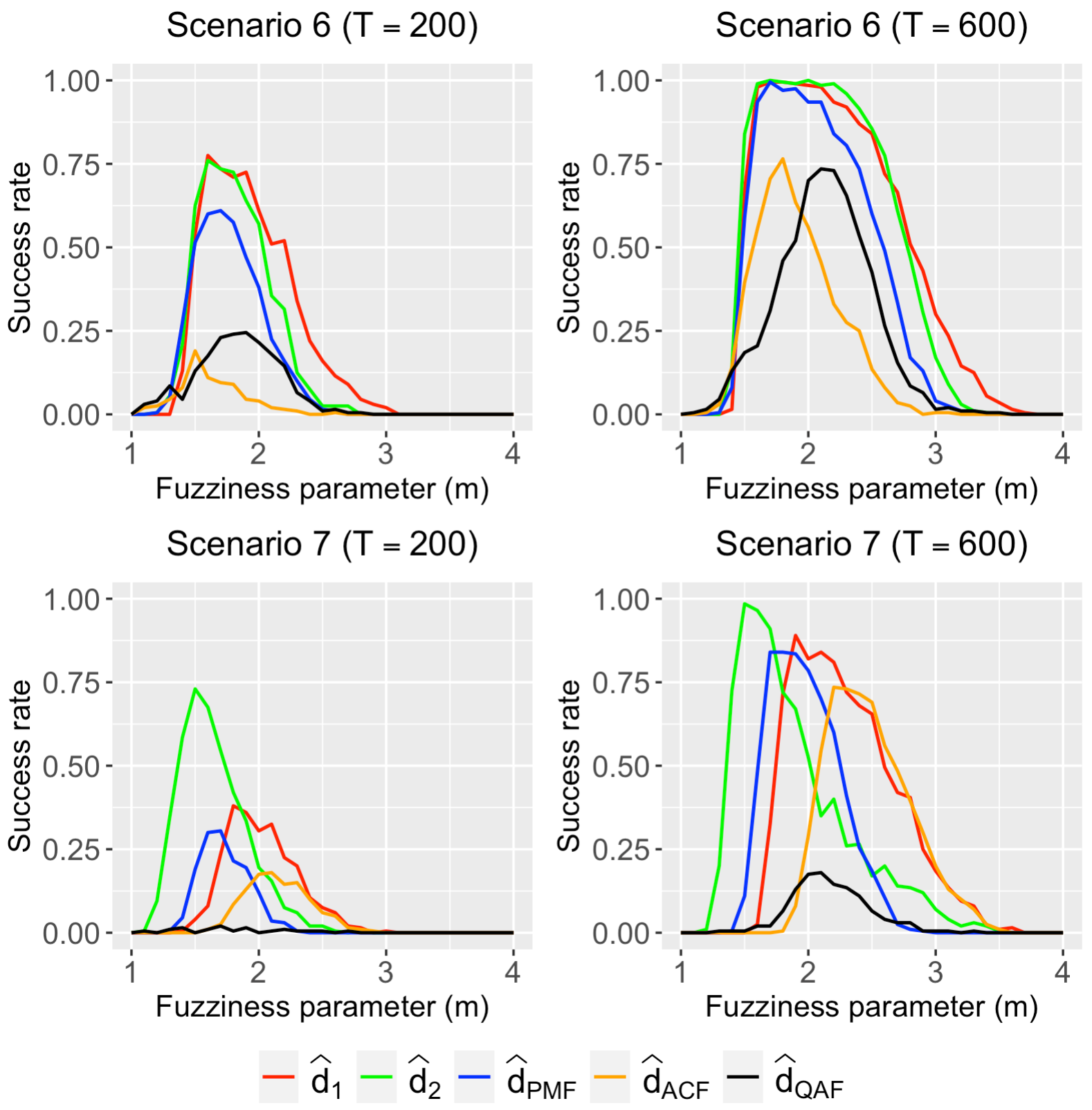}
		\caption{Rates of correct classification as function of $m$ obtained by the fuzzy $C$-medoids clustering algorithm based on several dissimilarities for a cutoff of 0.7. Scenarios 6 and 7.}
		\label{curvesscenarios67}
\end{figure}

Note that this evaluation criterion is very sensitive to the selection of $m$, since a single series with membership degrees failing to fulfil the required condition results in an incorrect classification. In fact, the different metrics could achieve their best behaviour for rather different values of $m$. For this reason, we decided to run the clustering algorithms for a grid of values for $m$ on the interval $(1, 4]$. Figure~\ref{curvesscenarios67} contains the curves of rates of correct classification as a function of $m$ for $\widehat{d}_1$, $\widehat{d}_2$, $\widehat{d}_{PMF}$, $\widehat{d}_{ACF}$ and $\widehat{d}_{QAF}$. The approach based on $\widehat{d}_{MC}$ showed a very poor performance and their results are here omitted. 

Plots in Figure~\ref{curvesscenarios67} confirm that the fuzziness parameter dramatically affects the clustering performance. In all cases, low and high values of $m$ produce poor rates of correct classification since partitions with all memberships close to 1 or to $1/2$ are respectively generated, thus resulting in failed trials. By contrast, moderate values of $m$ generally result in higher clustering effectiveness, although the optimal range varies for each distance. In Scenario~6, the proposed distances $\widehat{d}_1$ and $\widehat{d}_2$ attain the best results, outperforming the alternative metrics for most values of $m$, especially when $T=600$. Metric $\widehat{d}_{PMF}$ also shows a high clustering accuracy in this scenario, while $\widehat{d}_{ACF}$ and $\widehat{d}_{QAF}$ exhibit worse behaviour, particularly when $T=200$. Distance $\widehat{d}_2$ clearly leads to the best performing approach in Scenario~7 when $T=200$. A different situation happens when $T=600$, with $\widehat{d}_1$, $\widehat{d}_2$, and $\widehat{d}_{PMF}$ attaining high scores for several values of $m$. The quantile-based metric $\widehat{d}_{QAF}$ behaves very poorly in this scenario. In all cases, increasing the series length results in better rates of correct classification. Note that our results account for the importance of a suitable selection of $m$, although this issue is not addressed here because there are several procedures available in the literature for this purpose.

Rigorous comparisons based on Figure~\ref{curvesscenarios67} can be made by computing: (i) the maximum value of each curve, and (ii) the area under each curve, denoted by AUFC (area under the fuzziness curve), which was already used by \cite{lopez2022quantile1}. The values for both quantities are given in Table~\ref{tablescenarios67} and clearly corroborate the great performance of $\widehat{d}_1$ and $\widehat{d}_2$ when dealing with data sets including  series whose dynamic pattern does not belong to one specific group. In terms of AUFC, both metrics substantially outperform the rest in all cases. 

\begin{table}[!ht]
		\centering
		\resizebox{8cm}{!}{\begin{tabular}{ccccccc}  \hline
				Scenario 6	&  & & & & &  \\  
				& & $\widehat{d}_1$  &  $\widehat{d}_2$ &  $\widehat{d}_{PMF}$ &   $\widehat{d}_{ACF}$ &  $\widehat{d}_{QAF}$ \\  \hline
				$T=200$	  & Maximum & \textbf{0.76} & 0.73 &  0.62 & 0.14 & 0.25  \\ 
				& AUFC & \textbf{0.63} & 0.53 & 0.40 & 0.08 & 0.19  \\ \hline
				$T=600$	  & Maximum & 0.99 & \textbf{1.00} &  0.97 & 0.69 & 0.75  \\ 
				& AUFC & \textbf{1.34} & 1.31 & 1.06 & 0.54 & 0.63  \\ \hline 
				Scenario 7	&  & & & & &   \\  
				& & $\widehat{d}_1$  &  $\widehat{d}_2$ &  $\widehat{d}_{PMF}$ &   $\widehat{d}_{ACF}$ &  $\widehat{d}_{QAF}$  \\  \hline 
				$T=200$	  & Maximum & 0.38 & \textbf{0.75} & 0.35 & 0.19 & 0.03   \\
				& AUFC & 0.24 & \textbf{0.43} & 0.14 & 0.11 & 0.01  \\ \hline
				$T=600$	  & Maximum & 0.92 & \textbf{0.98}& 0.88 & 0.78 & 0.18  \\ 
				& AUFC & \textbf{0.86} & 0.79 & 0.62 & 0.61 & 0.12  \\ \hline 
		\end{tabular}}
		\caption{Maximum rates of correct classification and AUFC obtained by the fuzzy $C$-medoids clustering algorithm based on several distances for a cutoff value of 0.7. Scenarios 6 and 7. The best results are shown in bold.}
		\label{tablescenarios67}
\end{table}

Next step was to analyse the effect of the selected cutoff. A higher cutoff relaxes the condition for the membership degrees of the isolated series in order to be correctly classified, but establishes harder requirements for the membership degrees of the remaining series. The opposite happens with a lower cutoff. Thus, the experiments were repeated by fixing the cutoff values at 0.8 and 0.6, and the results are shown in Table~\ref{tablescenarios67additional}. For all metrics, the maximum rates of correct classification are very similar to the ones obtained in Table~\ref{tablescenarios67} with the cutoff at 0.7, but the AUFC values are dramatically different in most cases, thus accounting for the heavy influence of the cutoff in the evaluation mechanism. Specifically, lower and higher values are obtained when 0.8 and 0.6 are respectively used as cutoff. In any case, the proposed distances still outperform the alternative metrics in all settings.

\begin{table}[!ht]
		\centering
		\resizebox{11cm}{!}{\begin{tabular}{ccccccc}  \hline
				Scenario 6	&  & & & & &  \\  
				& & $\widehat{d}_1$  &  $\widehat{d}_2$ &  $\widehat{d}_{PMF}$ &   $\widehat{d}_{ACF}$ &  $\widehat{d}_{QAF}$ \\  \hline
				$T=200$	  & Maximum & \textbf{0.79} (\textbf{0.77}) & 0.78 (0.73)  & 0.61 (0.62)  & 0.13 (0.17) & 0.25 (0.28)  \\ 
				& AUFC & \textbf{0.37} (\textbf{1.23})  & 0.32 (1.03) & 0.24 (0.80)  & 0.05 (0.19)  & 0.11 (0.38)  \\ \hline
				$T=600$	  & Maximum & \textbf{1.00} (\textbf{1.00}) & \textbf{1.00} (\textbf{1.00})  &  \textbf{1.00} (\textbf{1.00}) & 0.66 (0.73) & 0.73 (0.74)  \\ 
				& AUFC & \textbf{0.85} (\textbf{2.78}) & 0.80 (2.67) & 0.67 (2.19) & 0.32 (1.09) & 0.40 (1.29)  \\ \hline 
				Scenario 7	&  & & & & &   \\  
				& & $\widehat{d}_1$  &  $\widehat{d}_2$ &  $\widehat{d}_{PMF}$ &   $\widehat{d}_{ACF}$ &  $\widehat{d}_{QAF}$  \\  \hline 
				$T=200$	  & Maximum & 0.40 (0.47) & \textbf{0.70} (\textbf{0.72}) & 0.29 (0.36) & 0.21 (0.23) &  0.02 (0.01) \\
				& AUFC & 0.17 (0.56) & \textbf{0.26} (\textbf{0.85}) & 0.08 (0.30) & 0.08 (0.26)  & 0.01 (0.02)   \\ \hline
				$T=600$	  & Maximum & 0.86 (0.92) & \textbf{0.98} (\textbf{1.00}) & 0.88 (0.93) & 0.72 (0.76) & 0.21 (0.23)  \\ 
				& AUFC & \textbf{0.51} (\textbf{1.76})  & 0.48 (1.66) & 0.39 (1.31) & 0.34 (1.23) & 0.07 (0.26)  \\ \hline 
		\end{tabular}}
		\caption{Maximum rates of correct classification and AUFC obtained by the fuzzy $C$-medoids clustering algorithm based on several dissimilarities for cutoff values of 0.8 and 0.6 (in brackets). Scenarios 6 and 7. The best results are shown in bold.}
		\label{tablescenarios67additional}
\end{table}

\begin{figure}[!ht]
		\centering
		\includegraphics[width=0.8\textwidth]{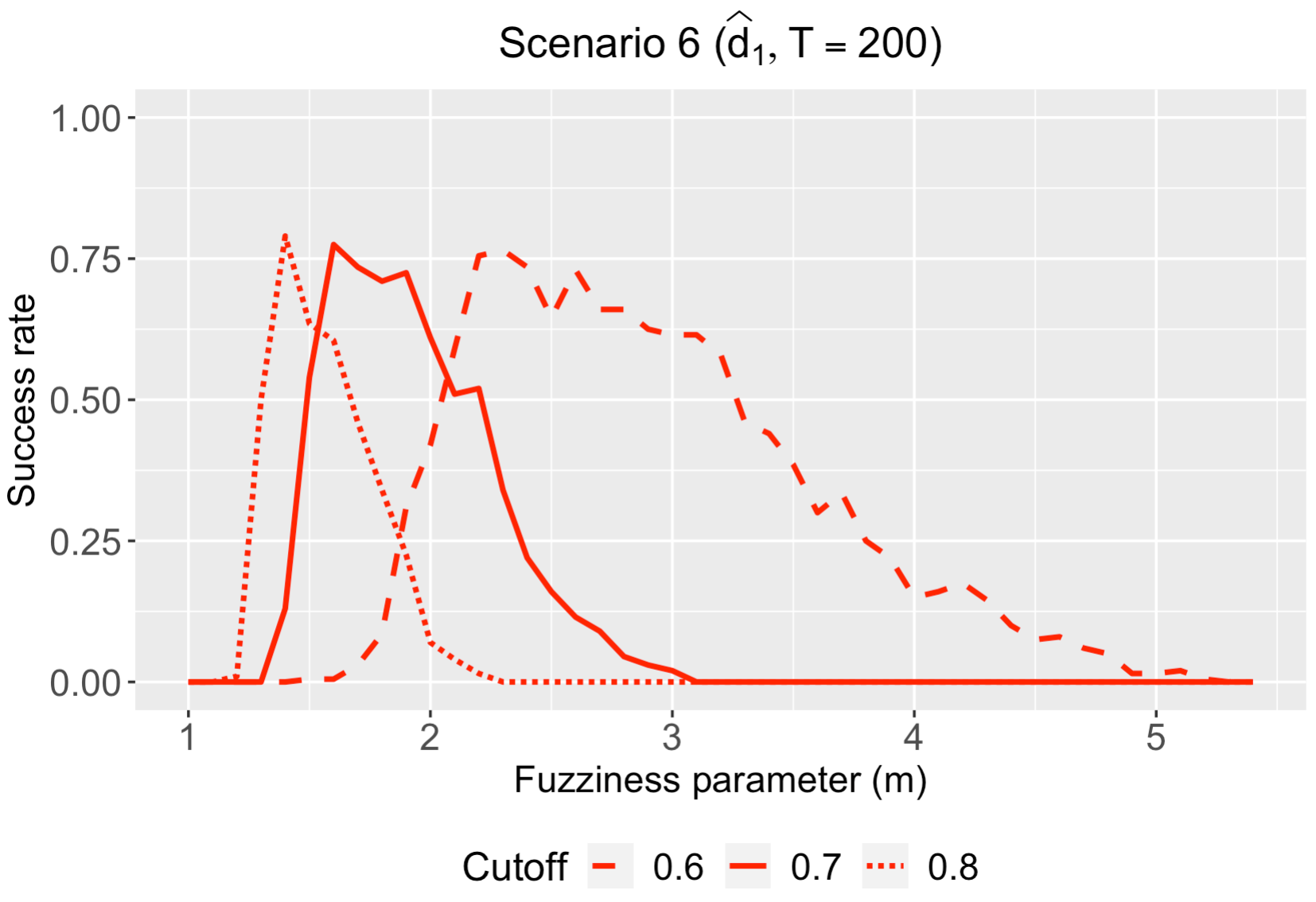}
		\caption{Rates of correct classification (as a function of $m$) obtained by the fuzzy $C$-medoids clustering algorithm based on $\widehat{d}_1$ for cutoff values of 0.6 (dashed line), 0.7 (solid line) and 0.8 (dotted line). Scenario 6 with $T=200$.}
		\label{3curvescutoff}
\end{figure}

To better understand the influence of the cutoff, we fixed the Scenario~6, the distance $\widehat{d}_{1}$ and $T=200$, and then examine the rates of correct classification with respect to $m$ for the cutoff values 0.6, 0.7 and 0.8. The obtained curves are displayed in Figure~\ref{3curvescutoff}. It is observed that the larger the cutoff value, the more concentrated and shifted to the left is the corresponding curve, which can be explained as follows. Low values of $m$ imply that the maximum membership degree is close to one for all series, thus making it easier to exceed the cutoff value, which in turn leads to misclassify the isolated series and correctly classify the series in the regular clusters. Indeed, if a high cutoff (e.g. 0.8) is used, then the isolated series are still well-classified for small values of $m$, but they would be misclassified in many trials if a smaller cutoff (e.g. 0.7 or 0.6) is used or when $m$ increases. By contrast, moderate and large values of $m$ move progressively the membership degrees towards 0.5, thus producing the opposite effect: failures with series in regular clusters and successes with the isolated series. However, it is worthy remarking that even large values of $m$ frequently generate maximum membership degrees above 0.6 for the non-isolated series, which justifies that the curve for a cutoff 0.6 is nonzero for a much broader range of values of $m$ and a rather large value for the corresponding AUFC. Additional experiments showed that the situation illustrated in Figure~\ref{3curvescutoff} is also observed for alternative values of $T$, in Scenario~7, and with the distance $\widehat{d}_2$.

In sum, the experiments from this section showed the clustering effectiveness of the proposed metrics also when series with a certain level of ambiguity are included in the data set subjected to clustering. Furthermore, the higher values of the AUFC attained by both $\widehat{d}_1$ and $\widehat{d}_2$ with respect to the alternative distances indicate a greater robustness of these distances to the choice of $m$. This is a nice property since the optimal selection of this parameter is still an open problem in the fuzzy clustering literature. 

\subsection{Evaluation of the weighted approach}\label{subsectionevaluationw}

The weighted fuzzy $C$-medoids model based on the proposed distances (see \eqref{wfcm}) was evaluated by considering the same scenarios, simulation parameters, and performance measures as in Sections~\ref{subsubsection1as} and \ref{subsubsection2as}. The corresponding average results are shown in Table~\ref{tablewvsnw}. For the sake of simplicity and homogeneity, we decided to show only the values of the ARIF index for Scenarios~1 to 5, and the rates of correct classification associated with $m \in \{1.2, 1.4, 1.6, 1.8, 2\}$ for Scenarios 6 and 7. The weighted versions using $\widehat{d}_1$ and $\widehat{d}_2$ are denoted by $\widehat{d}_{1, W}$ and $\widehat{d}_{2, W}$, respectively. To rigorously compare the weighted and non-weighted approaches, statistical tests based on the 200 trials were carried out, namely the Wilcoxon signed-rank test in Scenarios~1--5 and the McNemar test to compare two proportions in Scenarios~6 and 7. The tests were executed for each combination of scenario, metric, and values of $m$ and $T$, by considering paired-sample data and applying Bonferroni corrections for multiple comparisons. As regards the Wilcoxon signed-rank test, the alternative hypothesis stated that the average ARIF of the differences between the weighted and non-weighted versions is greater than 0.03. Asterisks in Table~\ref{tablewvsnw} indicate significant results at level $\alpha=0.05$.

\begin{table}[!ht]
		\centering
		\resizebox{8cm}{!}{\begin{tabular}{ccccccc}  \hline
				Scenario 1 & & $m=1.2$  &  $m=1.4$  &  $m=1.6$  &   $m=1.8$  &  $m=2$   \\  \hline 
				$T=200$	  & $\widehat{d}_{1, W}$ & 0.67 & 0.58 & 0.49 & 0.40 & 0.35   \\
				& $\widehat{d}_{2, W}$  & 0.81$^*$ & 0.69$^*$  & 0.58$^*$  & 0.47$^*$   &  0.40$^*$  \\  \hline
				$T=600$	  & $\widehat{d}_{1, W}$ & 0.93  & 0.85 & 0.76 & 0.64 &  0.55  \\ 
				& $\widehat{d}_{2, W}$ & 0.96 & 0.89$^*$ & 0.79$^*$ & 0.69$^*$ &  0.59$^*$ \\ \hline  
				Scenario 2 & & $m=1.2$  &  $m=1.4$  &  $m=1.6$  &   $m=1.8$  &  $m=2$   \\  \hline 
				$T=200$	  & $\widehat{d}_{1, W}$ & 0.57  & 0.54  & 0.47 & 0.41 & 0.36    \\
				& $\widehat{d}_{2, W}$  & 0.60  & 0.56 & 0.49$^*$ & 0.43$^*$  & 0.37$^*$   \\  \hline
				$T=600$	  & $\widehat{d}_{1, W}$ & 0.70  & 0.64  & 0.59 & 0.52 & 0.47  \\ 
				& $\widehat{d}_{2, W}$ & 0.72 & 0.66 & 0.60 & 0.54 & 0.49$^*$  \\ \hline  
				Scenario 3 & & $m=1.2$  &  $m=1.4$  &  $m=1.6$  &   $m=1.8$  &  $m=2$   \\  \hline 
				$T=200$	  & $\widehat{d}_{1, W}$ & 0.69  & 0.56 & 0.44 & 0.35 & 0.29   \\
				& $\widehat{d}_{2, W}$  & 0.67$^*$ & 0.64$^*$ & 0.55$^*$ & 0.45$^*$ & 0.38$^*$  \\  \hline
				$T=600$	  & $\widehat{d}_{1, W}$ & 0.92  & 0.81 & 0.68  & 0.56 & 0.47   \\ 
				& $\widehat{d}_{2, W}$ & 0.81 & 0.76$^*$   & 0.68$^*$  & 0.59$^*$  & 0.51$^*$    \\ \hline  
				Scenario 4 & & $m=1.2$  &  $m=1.4$  &  $m=1.6$  &   $m=1.8$  &  $m=2$   \\  \hline 
				$T=200$	  & $\widehat{d}_{1, W}$ & 0.51  & 0.46  & 0.38 & 0.31 & 0.25    \\
				& $\widehat{d}_{2, W}$  & 0.53 & 0.47  & 0.40$^*$ & 0.35$^*$ & 0.25  \\  \hline
				$T=600$	  & $\widehat{d}_{1, W}$ &  0.59 & 0.57 & 0.50 & 0.43 & 0.36 \\ 
				& $\widehat{d}_{2, W}$ &  0.61 & 0.57 & 0.53$^*$ & 0.45$^*$ & 0.36  \\ \hline  
				Scenario 5 & & $m=1.2$  &  $m=1.4$  &  $m=1.6$  &   $m=1.8$  &  $m=2$   \\  \hline 
				Variable $T$  & $\widehat{d}_{1, W}$ & 0.52  & 0.49 & 0.44 & 0.35 & 0.30   \\
				& $\widehat{d}_{2, W}$  & 0.56  & 0.50  & 0.47$^*$ & 0.39$^*$ & 0.29  \\  \hline
				Scenario 6 & & $m=1.2$  &  $m=1.4$  &  $m=1.6$  &   $m=1.8$  &  $m=2$   \\  \hline 
				$T=200$	  & $\widehat{d}_{1, W}$ &   0.00 & 0.14 & 0.57 & 0.57  & 0.42    \\
				& $\widehat{d}_{2, W}$  & 0.05  & 0.36$^*$  & 0.61  & 0.54  & 0.45$^*$   \\  \hline
				$T=600$	  & $\widehat{d}_{1, W}$ & 0.00  & 0.06  & 0.80  & 0.81  & 0.81   \\ 
				& $\widehat{d}_{2, W}$ & 0.08$^*$  & 0.62$^*$ & 0.90$^*$ & 0.76 & 0.75  \\ \hline   
				Scenario 7 & & $m=1.2$  &  $m=1.4$  &  $m=1.6$  &   $m=1.8$  &  $m=2$   \\  \hline 
				$T=200$	  & $\widehat{d}_{1, W}$ & 0.00  & 0.02  & 0.13  & 0.35  & 0.37    \\
				& $\widehat{d}_{2, W}$  & 0.25$^*$  & 0.63$^*$  & 0.70  & 0.55$^*$  & 0.31$^*$   \\  \hline
				$T=600$	  & $\widehat{d}_{1, W}$ & 0.00  & 0.01  & 0.08  & 0.88  & 0.85   \\ 
				& $\widehat{d}_{2, W}$ & 0.26$^*$  & 0.76  & 0.95  & 0.89$^*$  & 0.71$^*$   \\ \hline 
		\end{tabular}}
		\caption{Average values of ARIF (Scenarios~1 to 5) and rates of correct classification (Scenarios~6 and 7) obtained by the weighted fuzzy $C$-medoids clustering algorithm based on $\widehat{d}_{1}$ and $\widehat{d}_{2}$. An asterisk indicates that the weighted approach is significantly better than its non-weighted counterpart at a significance level $\alpha=0.05$.}
		\label{tablewvsnw}
\end{table}

According to Table~\ref{tablewvsnw}, the weighted fuzzy $C$-medoids model based on $\widehat{d}_1$ achieves higher average scores than its non-weighted counterpart in some settings (e.g., Scenario~3 with $T=600$), but the differences are always non-significant. This is because of the groups in Scenarios~1--7 can be distinguished by both the marginal distributions and the serial patterns. Thus, regarding that the marginal features can be directly obtained from the joint ones, it is expected that $\widehat{d}_{1, W}$ and $\widehat{d}_{1}$ show similar discriminatory power for any value of $\beta$. A better performance of the weighted approach is expected when clusters have similar marginal distributions but different dependence patterns. In fact, the strongest improvements (although not significant) are given in Scenario~3, whose clusters present the highest amount of similarity between marginal distributions. By contrast, the weighted algorithm based on $\widehat{d}_2$ yields significant improvements in several cases, especially for moderate to large values of $m$. In fact, it leads to the highest scores among the four proposed ones ($\widehat{d}_{1}$, $\widehat{d}_{2}$, $\widehat{d}_{1, W}$, and $\widehat{d}_{2, W}$) in some settings (e.g., Scenario 3 with $T=200$). Note that, as expected, giving weights to the marginal and bivariate components of $\widehat{d}_{1}$ and $\widehat{d}_{2}$ never results in a worse performance. Concerning Scenarios 1--5, similar results were obtained by using JIF to assess the clustering quality.

\begin{figure}[!ht]
		\centering
	\includegraphics[width=0.7\textwidth]{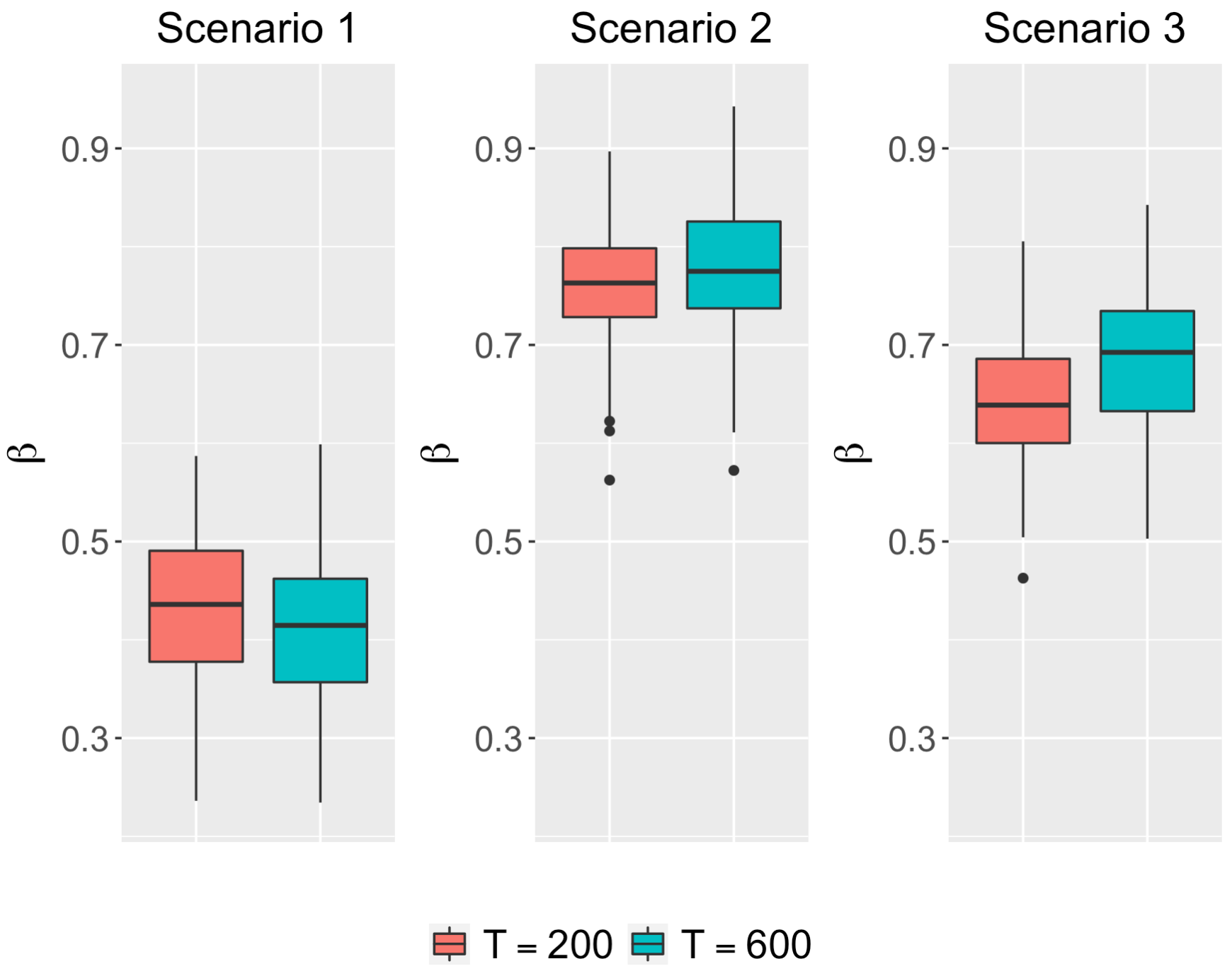}
		\caption{Distribution of the final value for $\beta$ produced by the weighted fuzzy $C$-medoids algorithm based on $\widehat{d}_2$. Scenarios 1, 2 and 3 with $m=2$.}
		\label{boxplots}
\end{figure}

Boxplots in Figure~\ref{boxplots} show the distribution of the weights $\beta$ returned by the algorithm based on $\widehat{d}_{2, W}$ in Scenarios~1--3 with $m=2$ (where the weighted approach outperforms the standard one). In Scenario~1, the algorithm usually leads to values of $\beta$ below 0.5, indicating that the bivariate component, $\widehat{d}_{2, B}$, plays a more important role than the marginal one, $\widehat{d}_{2, M}$. Regarding that the four processes in Scenario~1 clearly differ in both marginal and serial dependence structures, the lower weight received by $\widehat{d}_{2, M}$ is explained by the four terms defining this component, while $\widehat{d}_{2, B}$ only contributes with two terms. On the contrary, $\beta$ takes values above 0.5 in Scenarios~2 and 3. In Scenario 2, the higher weight for $\widehat{d}_{2, M}$ is justified by the fact that the four groups have different marginal features, while the serial features $\widehat{\kappa}_{d_{\text{o},1}}(1)$ and $\widehat{\kappa}_{d_{\text{o},1}}(2)$ take similar values on the pairs of clusters $(\mathcal{C}_1, \mathcal{C}_2)$ and $(\mathcal{C}_3, \mathcal{C}_4)$. An analogous situation happens in Scenario~3, albeit to a lesser extent. It is also noticeable the  low variability of $\beta$ in all settings. For instance, in Scenario~2 with $T=200$, $\beta$ moves from 0.7 to 0.8 more than 50\% of the times, which indicates that the optimal weight is approximated with high accuracy. Although not shown in the article for the sake of simplicity, similar boxplots are obtained for other values of $m$.

\subsection{Analysing clustering effectiveness with respect to selected lags}\label{subsectionsetl}


To examine the effect of a misspecification of the set of lags $\mathcal{L}$ required to compute $\widehat{d}_1$ and $\widehat{d}_2$, a sensitivity analysis was performed by considering Scenarios~1--3 and five different collections of lags, namely $\mathcal{L}_i=\{1,2,\ldots,i\}$, for $i=1, \ldots, 5$. The  average ARIF attained with $\widehat{d}_1$ and $\widehat{d}_2$ are given in Tables~\ref{tablel1} and \ref{tablel2}, respectively. For the sake of simplicity, only the results for $m=1.6$ are presented. The theoretical set of lags at each scenario is given in parentheses.
\begin{table}[!ht]
	\centering
	\textcolor{black}{
		\resizebox{9cm}{!}{
			\begin{tabular}{lcccccccc}   \hline
				& & \multicolumn{3}{c}{$T=200$} & & \multicolumn{3}{c}{$T=600$} \\ \cline{3-5} \cline{7-9} 
				Set &  & S1 ($\mathcal{L}_2$) & S2 ($\mathcal{L}_2$) & S3 ($\mathcal{L}_1$) & & S1 ($\mathcal{L}_2$) & S2 ($\mathcal{L}_2$) & S3 ($\mathcal{L}_1$) \\   \hline
				$\mathcal{L}_1$ &  & 0.47  & 0.46   & 0.40   &  & 0.75 & 0.59   & 0.64    \\ 
				$\mathcal{L}_2$ & & 0.49   & 0.48  & 0.39    & &  0.75 & 0.59  & 0.63   \\ 
				$\mathcal{L}_3$ & &  0.49 & 0.47  & 0.39    & &  0.76 & 0.59  & 0.63   \\ 
				$\mathcal{L}_4$ & & 0.49    & 0.47  & 0.39   & & 0.75  & 0.58 & 0.62   \\ 
				$\mathcal{L}_5$ &  & 0.49  & 0.47  & 0.39  & &  0.76   & 0.58  & 0.62   \\ 
				\hline \\
			\end{tabular}
		}
	}
\caption{Average ARIF based on $\widehat{d}_1$ for $m=1.6$ and different sets of lags ($\mathcal{L}_j=\{1,\ldots,j\}, j=1,\ldots,5$) in Scenarios~1, 2 and 3, denoted by S1, S2 and S3, respectively. The theoretical set of lags at each scenario is indicated in brackets.}
	\label{tablel1}
\end{table}
\begin{table}[!ht]
	\centering
	\textcolor{black}{
		\resizebox{9cm}{!}{
			\begin{tabular}{lcccccccc}   \hline
				& & \multicolumn{3}{c}{$T=200$} & & \multicolumn{3}{c}{$T=600$} \\ \cline{3-5} \cline{7-9} 
				Set &  & S1 ($\mathcal{L}_2$) & S2 ($\mathcal{L}_2$) & S3 ($\mathcal{L}_1$) & & S1 ($\mathcal{L}_2$) & S2 ($\mathcal{L}_2$) & S3 ($\mathcal{L}_1$) \\   \hline
				$\mathcal{L}_1$ &  & 0.47  &  0.49   & 0.45 & &  0.70 &  0.62  & 0.64   \\ 
				$\mathcal{L}_2$ & &  0.49   & 0.45  & 0.40   & &  0.73  & 0.59  & 0.61  \\ 
				$\mathcal{L}_3$ & & 0.50  & 0.42  & 0.35   & &  0.72  & 0.58  & 0.56   \\ 
				$\mathcal{L}_4$ & & 0.49   & 0.40  & 0.32   & & 0.71   & 0.54 & 0.52  \\ 
				$\mathcal{L}_5$ &  &  0.48  & 0.37   & 0.30   & &  0.71  & 0.53  & 0.49   \\ 
				\hline \\
			\end{tabular}
		}
	}
	\caption{Average ARIF based on $\widehat{d}_2$ for $m=1.6$ and different sets of lags ($\mathcal{L}_j=\{1,\ldots,j\}, j=1,\ldots,5$) in Scenarios~1, 2 and 3, denoted by S1, S2 and S3, respectively. The theoretical set of lags at each scenario is indicated in brackets.}
	\label{tablel2}
\end{table}

Results in Table~\ref{tablel1} suggest that metric $\widehat{d}_1$ is clearly robust to the choice of $\mathcal{L}$. In fact, in all scenarios and for both values of $T$, no significant changes are observed in the average scores of the ARIF. A similar conclusion follows from Table~\ref{tablel2} for $\widehat{d}_2$ in Scenario~1. However, in Scenarios~2 and 3, $\widehat{d}_2$ slightly decreases its performance as more lags are added to $\mathcal{L}$, i.e.  including unnecessary features (noise) in the time series representation 
negatively affects the clustering performance. Note that $\widehat{d}_2$ achieves the highest ARIF in Scenario~2 with $\mathcal{L}_1$, although $\mathcal{L}_2$ is here the theoretical situation. This is because $\kappa_{d_{\text{o},1}}(2)$ takes similar values for clusters $\mathcal{C}_3$ and $\mathcal{C}_4$ in this scenario. Thus,
including $\widehat{\kappa}_{d_{\text{o},1}}(2)$ helps to separate $\mathcal{C}_1$ and $\mathcal{C}_2$ from $\mathcal{C}_3$ and $\mathcal{C}_4$, but makes it harder to distinguish between the series of clusters $\mathcal{C}_3$ and $\mathcal{C}_4$, which results in a lower accuracy. 


In sum, even for $\widehat{d}_2$, small deviations from the nominal lag order do not have a substantial impact on the clustering accuracy. Thus, while the optimal lag selection is a critical issue in modelling and forecasting problems, the clustering approaches based on both $\widehat{d}_1$ and $\widehat{d}_2$ exhibit a reasonable robustness to a non-optimal choice of $\mathcal{L}$. This is a particularly nice property in our setting because the proposed algorithms are model-free and no single lag selection procedure has been proven to perform properly with all time series models.

The mentioned robustness property justifies to select $\mathcal{L}$ through a simple and automatic procedure, mainly satisfying two properties: applicability without prior assumptions about the generating models and computational efficiency. To this aim, we propose a criterion based on assessing serial dependence at several lags for each OTS. Specifically, we consider the partial Cohen's $\kappa$ at lag $l$, denoted by $\kappa^p_{d_{\text{o},1}}(l)$, which is defined in an analogous way to the partial autocorrelation in the real-valued setting. In practice, the sample counterparts $\widehat{\kappa}^p_{d_{\text{o},1}}(1), \widehat{\kappa}^p_{d_{\text{o},1}}(2), \widehat{\kappa}^p_{d_{\text{o},1}}(3), \ldots$ can be computed from $\widehat{\kappa}_{d_{\text{o},1}}(1), \widehat{\kappa}_{d_{\text{o},1}}(2), \widehat{\kappa}_{d_{\text{o},1}}(3) \ldots$ via the Durbin-Levinson algorithm \cite{levinson1949wiener, durbin1960fitting}, just as the partial autocorrelations are obtained from the autocorrelations. Using $\widehat{\kappa}^p_{d_{\text{o},1}}(l)$ instead of $\widehat{\kappa}_{d_{\text{o},1}}(l)$, the significant lags are free of quantifying dependence explained by shorter lags, which could be helpful to identify the maximum significant lag. On the other hand, according to Theorem~7.2.1 in \cite{weiss2019distance}, $\kappa^p_{d_{\text{o},1}}(l)$ has the same asymptotic distribution as $\widehat{\kappa}_{d_{\text{o},1}}(l)$ under serial independence. Based on these arguments, given the set $\mathbb{S}=\{X_{T_1}^{(1)}, \ldots, X_{T_s}^{(s)}\}$ of OTS subject to clustering, we propose to select $\mathcal{L}$ as follows.

\begin{enumerate}
\item[1.] Fix a global significance level $\alpha>0$ and a maximum lag $L_{\text{Max}}\in\mathbb{N}$. Adjust the significance level in a suitable way, obtaining the corrected significance level $\alpha'$.  
\item[2.] For each series $X_{T_i}^{(i)}$ in $\mathbb{S}$:
\begin{enumerate}
\item[2.1.] Use the sample version of the ordinal Cohen's $\kappa$ to test for serial independence at all lags up to $L_{\text{Max}}$. Specifically, for $l=1,2, \ldots, L_{\text{Max}}$, the null hypothesis is rejected if
\begin{equation}\label{testselectionl}
\Bigg | \frac{\sqrt{T_i}\widehat{\text{disp}}^{(i)}_{d_{\text{o},1}}\big(\widehat{\kappa}^p_{d_{\text{o},1}}(l)^{(i)}+1/T_i\big)}{2\sqrt{\sum_{k,l=0}^{n-1}\big(\widehat{f}^{(i)}_{\min\{k, l\}}-\widehat{f}^{(i)}_k\widehat{f}^{(i)}_l\big)^2}} \Bigg | > z_{1-\alpha'/2},
\end{equation}
\noindent where the superscript $(i)$ indicates that the estimates are computed with respect to the $i$th series, and $z_\theta$ is the $\theta$-quantile of the standard normal distribution.
\item[2.2.] Record the maximum significant lag, $L^{(i)}$, according to \eqref{testselectionl}.
\end{enumerate}
\item[3.] Consider $L^*=\max\{L^{(1)}, L^{(2)}, \ldots, L^{(s)}\}$ and define $\mathcal{L}=\{1, 2, \ldots, L^*\}$. 
\end{enumerate}

Some remarks concerning the previous procedure are given below. The global significance level $\alpha$ is corrected in Step~1 to address the problem of multiple comparisons, since $sL_{\text{Max}}$ statistical tests are simultaneously performed. The use of a conservative rule (e.g., the Bonferroni correction) is recommended, since frequently a few lags are sufficient to characterize the serial dependence. Nonetheless, other less conservative procedures ensuring that the family-wise error rate is at most $\alpha$ could be employed. 
In Step~3, $L^*$ is the highest lag within $\mathcal{L}$. By construction, $L^*$ is necessarily a significant lag for one or several series, although indeed some series might not exhibit significant serial dependence at $L^*$ or lower lags. However, this is not an issue because the corresponding estimated features are expected to be close to zero for these series.

Our proposal to select $\mathcal{L}$ was examined via simulation by considering the series in Scenarios~1--3, and setting $L_{\text{Max}}=5$ and $\alpha=0.05$ (with the Bonferroni correction). Based on 1000 simulation trials for each $T\in\{200, 600\}$, Table~\ref{tablepercentagel} provides the proportion of times that each set of lags was selected. It is observed that the proposed method works reasonably well, although the results differ among the considered scenarios. In Scenario~3, the theoretical set ($\mathcal{L}_1$) is selected almost 100\% of the times regardless of the series length. The most often chosen set in Scenario~1 is also the theoretical one ($\mathcal{L}_2$), but here $\mathcal{L}_3$, $\mathcal{L}_4$ and $\mathcal{L}_5$ are selected a non-negligible number of times. It is worth to recall that this is not a problem in our setting since the clustering effectiveness for both metrics $\widehat{d}_1$ and $\widehat{d}_2$ is approximately the same for all  $\mathcal{L}_i$, $i=1,\ldots,5$ (see Tables~\ref{tablel1} and \ref{tablel2}). Lastly, in Scenario~2, the series length has a substantial impact on the selection of $\mathcal{L}$. When $T=200$, $\mathcal{L}_1$ is erroneously chosen most of the trials. Basically, this series length is too short to detect the serial dependence exhibited by the series within clusters $\mathcal{C}_3$ and $\mathcal{C}_4$, generated by processes with coefficients $\alpha_1$ and $\alpha_2$ very close to zero. Again, the proposed clustering algorithms do not get negatively affected in Scenario~1 when only the first lag is considered (see Tables~\ref{tablel1} and \ref{tablel2}). Note that, when $T=600$, the proper set $\mathcal{L}_2$ is usually selected because the power of the corresponding tests substantially increases. Analogous conclusions were obtained using $\alpha=0.01$ and $\alpha=0.10$.

\begin{table}[!ht]
	\centering
\resizebox{8.5cm}{!}{\begin{tabular}{lcccccc} \hline 
			&       & $\mathcal{L}_1$ & $\mathcal{L}_2$ & $\mathcal{L}_3$ & $\mathcal{L}_4$ & $\mathcal{L}_5$ \\ \hline 
			Scenario 1 ($\mathcal{L}_2$) & $T=200$ & 0.000   & \textbf{0.462} & 0.285 &  0.111   &    0.142 \\
			& $T=600$ &   0.000   &    \textbf{0.595} & 0.230   &  0.104   &    0.071 \\ \hline  
			Scenario 2 ($\mathcal{L}_2$) & $T=200$ & \textbf{0.693}   &    0.255 & 0.023   &  0.017   &    0.012 \\
			& $T=600$    &    0.244 & \textbf{0.729} &  0.015   &    0.004 & 0.008 \\ \hline
			Scenario 3 ($\mathcal{L}_1$)& $T=200$ & \textbf{0.965}   & 0.007 &  0.010   &  0.005   &    0.013 \\
			& $T=600$ & \textbf{0.969} & 0.013 & 0.007  &  0.007   &    0.004  \\ \hline
	\end{tabular}}
	\caption{Proportion of times that each set $\mathcal{L}_i$ was selected according to the proposed criterion using $\alpha=0.05$ and Bonferroni correction. Scenarios~1, 2 and 3 (theoretical set of lags in brackets). For each scenario and value of $T$, the largest rate is shown in bold.}
	\label{tablepercentagel}
\end{table}

\section{Applications.}\label{sectionapplications}

This section is devoted to show two real-data applications of the proposed clustering procedures. 
In both cases, we first describe the database along with some exploratory analyses and, afterwards, we show the results of applying the clustering algorithms.

\subsection{Fuzzy clustering of European countries in terms of credit ratings}\label{subsectionapplication1}


\subsubsection{Data set and exploratory analyses}\label{subsubsectiondaea1}

Let us consider the financial database introduced in Section~\ref{subsectionmotivating} and formerly employed by \cite{weiss2019distance}, which contains monthly credit ratings according to S\&P for the UK and the 27 countries of the EU, namely Austria (AT), Belgium (BE), Bulgaria (BG), Cyprus (CY), Czechia (CZ), Germany (DE), Denmark (DK), Estonia (EE), Spain (ES), Finland (FI), France (FR), Greece (GR), Croatia (HR), Hungary (HU), Ireland (IE), Italy (IT), Lithuania (LT), Luxembourg (LU), Latvia (LV), Malta (MT), Netherlands (NL), Poland (PL), Portugal (PT), Romania (RO), Sweden (SE), Slovenia (SL), and Slovakia (SK). The sample period spans from January 2000 to December 2017, thus resulting serial realizations of length $T=216$. The range of the OTS consists of $n+1=23$ states, $s_0, \ldots, s_{22}$, representing the different credit scores (see Section~\ref{subsectionmotivating}). As stated in \cite{weiss2019distance}, the profiles of the 28 OTS show quite different shapes, including constant trajectories but also paths with up or down movements (financial crisis). As an example, the OTS for Estonia and Slovakia are represented in Figure~\ref{estoniaslovakia}. Applying clustering on this data set could lead to meaningful groups of countries sharing similar risk profiles, monetary policy, or even government reliability. Moreover, since our approach produces  fuzzy solutions, some countries exhibiting a vague behaviour in terms of credit ratings could be identified. 

\begin{figure}[!ht]
		\centering
		\includegraphics[width=1\textwidth]{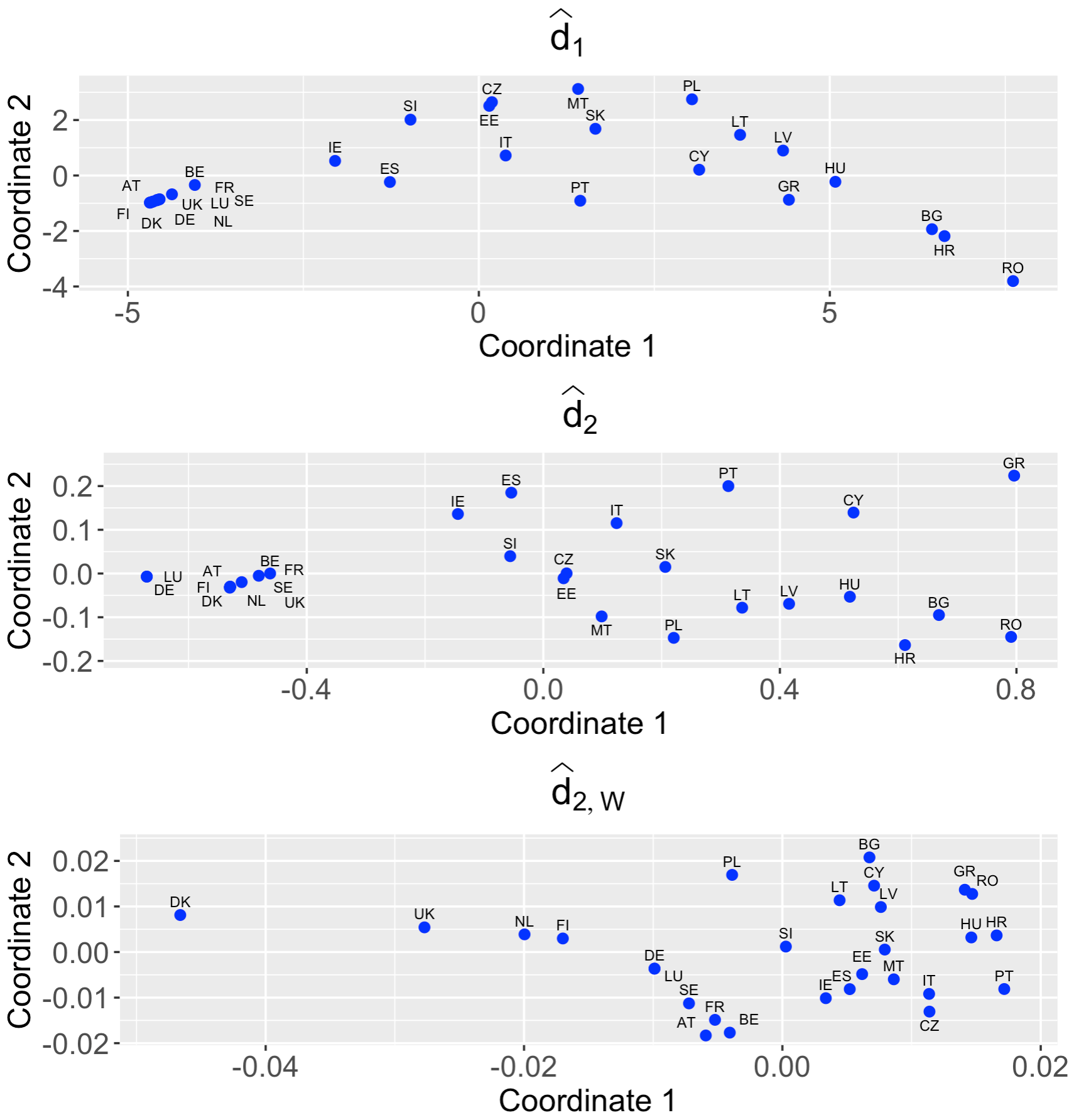}
		\caption{Two-dimensional scaling planes based on distances $\widehat{d}_1$ (top), $\widehat{d}_2$ (middle) and $\widehat{d}_{2, W}$ with $\beta=0.14$ (bottom) for the monthly credit ratings of 28 European countries.}
		\label{2dscr}
\end{figure}

As a preliminary exploratory step, we performed a 2DS based on the pairwise dissimilarity matrices calculated by using $\widehat{d}_1$ and $\widehat{d}_2$. 
The required set of lags $\mathcal{L}$ was determined by means of the procedure proposed in Section~\ref{subsectionsetl} (with $\alpha=0.05$, $L_{\text{Max}}=10$ and the Bonferroni correction), resulting $\mathcal{L}=\{1\}$. It is worth remarking that only the first lag was also selected by using alternative rules for correcting the significance level (e.g., Holm or Hommel corrections) and different values of $\alpha$. The 2DS plots based on $\widehat{d}_1$ and $\widehat{d}_2$ with $\mathcal{L}=\{1\}$ are displayed in the top ($R^2=0.90$) and middle ($R^2=0.95$) panels of Figure~\ref{2dscr}, respectively. 

From Figure~\ref{2dscr} follows that the clustering algorithms based on $\widehat{d}_1$ and $\widehat{d}_2$ give rise to quite similar configurations. There is a compact group of 10 countries clearly separated from the rest and formed by AT, BE, DE, DK, FI, FR, LU, NL, SE, and UK. Interestingly, these countries are usually characterized for having strong economies (e.g., high average income, low inflation rates \ldots). The remaining countries are more spread-out, forming poorly-separated groups, which suggests that a fuzzy approach could be particularly useful to get meaningful conclusions from this database. Note that the both plots show some interesting differences. For instance, while GR is located close to other countries in Eastern Europe when employing $\widehat{d}_1$, it constitutes an isolated point (a potential outlier) when $\widehat{d}_2$ is considered. In a clustering context, these isolated points often require an individual analysis, since their presence can negatively affect the performance of standard algorithms. 

\subsubsection{Application of clustering algorithms and results}\label{subsubsectionaca1}

Two important parameters must be set in advance before executing the clustering algorithms, namely the number of clusters, $C$, and the fuzziness parameter, $m$. Note that the latter parameter highly influences the quality of the obtained clustering partition as seen in Section \ref{sectionsimulationstudy}. The selection of $C$ and $m$ was done simultaneously by means of a procedure proposed by \cite{lopez2022quantile1}, which is based on two steps: (i) fixing a grid of values for the pair $(C, m)$, and (ii) choosing the pair leading to the minimum value of a measure relying on four internal clustering validity indices, namely the Xie-Beni index \cite{xie1991validity}, the Kwon index \cite{kwon1998cluster}, and the indices proposed in \cite{tang2005improved} and in \cite{bensaid1996validity}. These indices measure the degree of compactness of a given clustering solution and, in all cases, the lower the value of the index, the better the quality of the partition. In particular, for fixed $C$ and $m$, \cite{lopez2022quantile1} consider the average of a standardized version of these indices, thus bringing them to the same scale. In this application, the grid was constructed by setting $C \in \{1, 2, \ldots, 7\}$ and $m \in \{1.1, 1.2, \ldots, 4\}$, and the selected values were $(C, m)=(3, 1.9)$ for $\widehat{d}_1$ and $(C, m)=(3, 2.1)$ for $\widehat{d}_2$. Hence, both metrics identify an underlying partition with the same number of groups, which is coherent with the similarity of both plots in Figure~\ref{2dscr}. 

Table~\ref{tablefuzzy1} contains the membership degrees produced by the fuzzy $C$-medoids clustering algorithm based on both metrics. Superscripts 1 and 2 on the first column indicate the medoid countries according to $\widehat{d}_1$ and $\widehat{d}_2$, respectively. To ease interpretation, only one, two or three membership degrees for the $i$th country were highlighted in grey according to the following criterion: (i) only the $j$th membership is shaded gray when $u_{ij}>0.5$ and $u_{ik}<0.3$, for $k\ne j$, (ii) the three membership degrees are highlighted if $u_{ij}>0.25$, for all $j \in \{1, 2, 3\}$, and (iii) otherwise, one of them is below 0.25 and the remaining two are reasonably spread-out, so the latter ones are highlighted. These criteria basically provide a simple way of interpreting the fuzzy solutions produced by both metrics.

\begin{table}[!t]
	\centering
	\resizebox{9.5cm}{!}{\begin{tabular}{c|ccc|ccc}
		\hline
		 &  & $\widehat{d}_1$&  &  & $\widehat{d}_2$ &   \\ \hline 
		Country & $\mathcal{C}_1$ & $\mathcal{C}_2$ & $\mathcal{C}_3$ & $\mathcal{C}_1$ & $\mathcal{C}_2$ & $\mathcal{C}_3$   \\ 
		\hline
		AT & \cellcolor[gray]{0.8}{0.909} & 0.044 & 0.048 & \cellcolor[gray]{0.8}{0.996} & 0.003 & 0.001    \\
		BE & \cellcolor[gray]{0.8}{0.628} & 0.176 & 0.197 & \cellcolor[gray]{0.8}{0.760} & 0.149 & 0.091  \\
		BG & 0.199 & \cellcolor[gray]{0.8}{0.376} & \cellcolor[gray]{0.8}{0.426} & 0.133 & 0.265 & \cellcolor[gray]{0.8}{0.603}  \\
		CY & 0.146 & \cellcolor[gray]{0.8}{0.452} & \cellcolor[gray]{0.8}{0.402} & 0.132 & 0.288 & \cellcolor[gray]{0.8}{0.580}  \\ 
		CZ & 0.184 & \cellcolor[gray]{0.8}{0.524} & 0.292 & 0.034 & \cellcolor[gray]{0.8}{0.913} & 0.054  \\
		DE$^2$ & \cellcolor[gray]{0.8}{0.958} & 0.020 & 0.022 & \cellcolor[gray]{0.8}{1.000} & 0.000 & 0.000  \\ 
		DK & \cellcolor[gray]{0.8}{0.981} & 0.009 & 0.010 & \cellcolor[gray]{0.8}{0.999} & 0.001 & 0.000  \\
		EE$^2$& 0.174 & \cellcolor[gray]{0.8}{0.547} & 0.280 & 0.000 & \cellcolor[gray]{0.8}{1.000}& 0.000  \\
		ES & \cellcolor[gray]{0.8}{0.342} & \cellcolor[gray]{0.8}{0.309} & \cellcolor[gray]{0.8}{0.349} & 0.218 & \cellcolor[gray]{0.8}{0.543} & 0.239  \\ 
		FI & \cellcolor[gray]{0.8}{0.925} & 0.036 & 0.039 & \cellcolor[gray]{0.8}{0.996} & 0.002 & 0.002  \\ 
		FR & \cellcolor[gray]{0.8}{0.837} & 0.077 & 0.086 & \cellcolor[gray]{0.8}{0.812} & 0.116 & 0.072  \\ 
		GR & 0.212 & \cellcolor[gray]{0.8}{0.385} & \cellcolor[gray]{0.8}{0.403} & 0.176 & \cellcolor[gray]{0.8}{0.320} & \cellcolor[gray]{0.8}{0.504}  \\ 
		HR & 0.204 & \cellcolor[gray]{0.8}{0.372} & \cellcolor[gray]{0.8}{0.424} & 0.123 & 0.249 & \cellcolor[gray]{0.8}{0.628}  \\
		HU & 0.170 & \cellcolor[gray]{0.8}{0.408} & \cellcolor[gray]{0.8}{0.422} & 0.087 & 0.197 & \cellcolor[gray]{0.8}{0.716}  \\
		IE & \cellcolor[gray]{0.8}{0.416} & \cellcolor[gray]{0.8}{0.296} & \cellcolor[gray]{0.8}{0.289} & 0.269 & \cellcolor[gray]{0.8}{0.511} & 0.220  \\
		IT & 0.159 & \cellcolor[gray]{0.8}{0.457} & \cellcolor[gray]{0.8}{0.383} & 0.132 & \cellcolor[gray]{0.8}{0.601} & 0.268  \\
		LT & 0.161 & \cellcolor[gray]{0.8}{0.496} & \cellcolor[gray]{0.8}{0.343} & 0.085 & 0.244 & \cellcolor[gray]{0.8}{0.672}  \\ 
		LU & \cellcolor[gray]{0.8}{0.958} & 0.020 & 0.022 & \cellcolor[gray]{0.8}{1.000} & 0.000 & 0.000  \\
		LV$^2$ & 0.165 & \cellcolor[gray]{0.8}{0.452} & \cellcolor[gray]{0.8}{0.383} & 0.000 & 0.000 & \cellcolor[gray]{0.8}{1.000}  \\ 
		MT & 0.143 & \cellcolor[gray]{0.8}{0.619} & 0.238 & 0.141 & \cellcolor[gray]{0.8}{0.562} & 0.297  \\
		NL$^1$ & \cellcolor[gray]{0.8}{1.000} & 0.000 & 0.000 & \cellcolor[gray]{0.8}{0.995} & 0.002 & 0.003  \\
		PL & 0.185 & \cellcolor[gray]{0.8}{0.488} & \cellcolor[gray]{0.8}{0.328} & 0.128 & \cellcolor[gray]{0.8}{0.408} & \cellcolor[gray]{0.8}{0.464}  \\
		PT$^1$ & 0.000 & 0.000 & \cellcolor[gray]{0.8}{1.000} & 0.150 & \cellcolor[gray]{0.8}{0.393} & \cellcolor[gray]{0.8}{0.457}  \\   
		RO & 0.220 & \cellcolor[gray]{0.8}{0.356} & \cellcolor[gray]{0.8}{0.423} & 0.158 & 0.290 & \cellcolor[gray]{0.8}{0.551}  \\ 
		SE & \cellcolor[gray]{0.8}{0.954} & 0.022 & 0.024 & \cellcolor[gray]{0.8}{0.992} & 0.001 & 0.007  \\ 
		SI & 0.245 & \cellcolor[gray]{0.8}{0.440} & \cellcolor[gray]{0.8}{0.315} & 0.150 & \cellcolor[gray]{0.8}{0.682} & 0.168  \\ 
		SK$^1$ & 0.000 & \cellcolor[gray]{0.8}{1.000} & 0.000 & 0.120 & \cellcolor[gray]{0.8}{0.485} & \cellcolor[gray]{0.8}{0.396}  \\ 
		UK & \cellcolor[gray]{0.8}{0.953} & 0.023 & 0.025 & \cellcolor[gray]{0.8}{0.908} & 0.056 & 0.036  \\
		\hline
	\end{tabular}}
	\caption{Membership degrees of 28 European countries produced by the fuzzy $C$-medoids clustering algorithm based on metrics $\widehat{d}_1$ and $\widehat{d}_2$ for a 3-cluster partition. The superscripts 1 and 2 are used to indicate the medoid countries according to $\widehat{d}_1$ and $\widehat{d}_2$, respectively. For each country, the corresponding memberships were highlighted according to their values.}
	\label{tablefuzzy1}
\end{table}

Both clustering partitions in Table~\ref{tablefuzzy1} are consistent with the corresponding 2DS plots in Figure~\ref{2dscr}. Cluster $\mathcal{C}_1$ contains the ten countries constituting the well-separated group in the left part of the graphs, all with membership degrees above 0.5. Therefore, both metrics are capable of properly detecting the group including the strongest economies in Europe. On the other hand, clusters $\mathcal{C}_2$ and $\mathcal{C}_3$ are formed by countries exhibiting scattered membership degrees, which was expected since no clear clustering structure is observed for these countries in Figure~\ref{2dscr}. However, a quick glance at the distribution of the highlighted membership degrees allows us to conclude that both groups exhibit a much less degree of overlap in the partition produced by $\widehat{d}_2$. For instance, with $\widehat{d}_2$, cluster $\mathcal{C}_3$ groups together several countries located in Eastern Europe with high membership degrees, namely BG, CY, HR, HU, LT, LV, and RO. Note that both partitions contain some countries exhibiting a substantially fuzzy behaviour. For instance, the membership degrees of ES are all close to $\frac{1}{3}$ in the partition generated by $\widehat{d}_1$, thus suggesting equidistance from the three clusters. 
This fact is not surprising since ES is known to have a promising economy, but far less powerful than the ones of the countries in cluster $\mathcal{C}_1$. Analogous conclusions can be obtained for other countries whose membership degrees are evenly distributed between the 3 groups. Indeed, these insights can be reached due to the fuzzy nature of the partitions, remaining obscured with crisp partitions. Therefore, this example illustrates the usefulness of the fuzzy paradigm when performing clustering of ordinal series in real databases. 

\begin{figure}[!ht]
	\centering
	\includegraphics[width=1\textwidth]{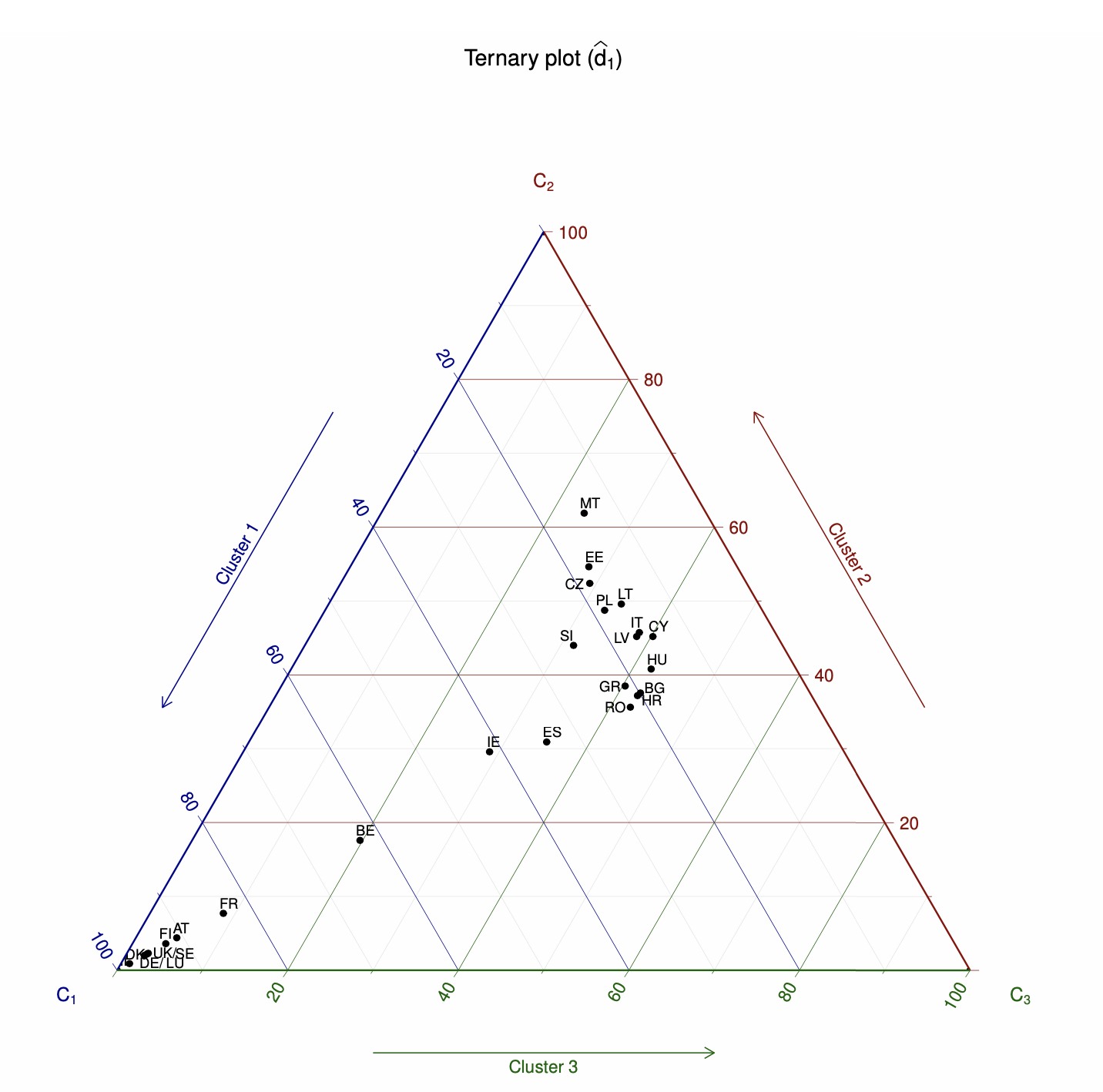}
	\caption{Ternary plot associated with the 3-cluster solution produced by distance $\widehat{d}_1$ in the data set of credit ratings.}
	\label{tp1}
\end{figure}

\begin{figure}[!ht]
	\centering
	\includegraphics[width=1\textwidth]{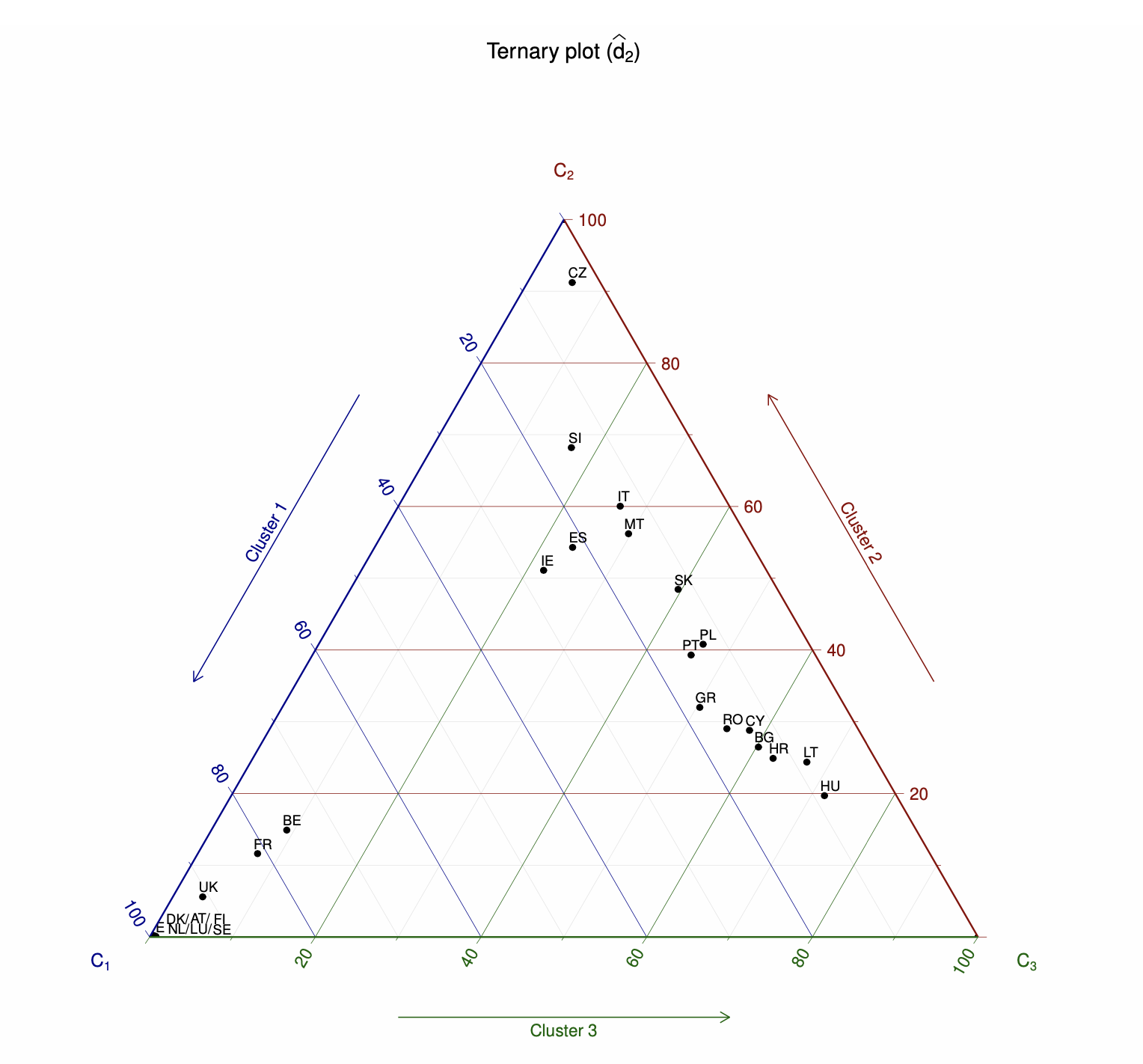}
	\caption{Ternary plot associated with the 3-cluster solution produced by distance $\widehat{d}_2$ in the data set of credit ratings.}
	\label{tp2}
\end{figure}

To gain greater insights into the clustering solutions returned by $\widehat{d}_1$ and $\widehat{d}_2$, the corresponding ternary plots are given in Figures~\ref{tp1} and \ref{tp2}, respectively. The medoids are placed in the vertexes, while the position of the rest of objects is determined by their vector of membership degrees. Note that the ternary plot based on $\widehat{d}_1$ clearly suggests a higher degree of overlap between $\mathcal{C}_2$ and $\mathcal{C}_3$.


\begin{rem}
\label{remWeightedd2}  
\textit{Clustering based on $\widehat{d}_{2, W}$}. We also run the weighted fuzzy $C$-medoids algorithm based on $\widehat{d}_{2}$. As with the unweighted approach, we set $C=3$ and $m=2.1$. The algorithm returned the value $\beta=0.14$, which indicates that the clustering partition is mostly driven by the serial component ($\widehat{d}_{2, B}$). The 2DS plot based on the combined metric $0.14^2\widehat{d}_{2, M}+0.86^2\widehat{d}_{2, B}$ is displayed in the bottom panel of Figure~\ref{2dscr}. Note that the ten richest countries are still separated from the remaining ones, but showing a much larger degree of dispersion. On the contrary, the dispersion among the rest of the countries clearly decreases. Overall, the resulting partition presents a higher degree of fuzziness than the one provided by $\widehat{d}_{2}$, thus making it more difficult to interpret the clusters. Hence, even though the weighted procedure leads to the partition with the best trade-off between intra-cluster compactness and inter-cluster separation, the unweighted approach provides more meaningful groups in the context of the current application.
\end{rem}

\subsection{Fuzzy clustering of Austrian wage mobility data}\label{subsectionapplication2}


\subsubsection{Data set and exploratory analyses}\label{subsubsectiondaea2}

The second case study is related to the nonsupervised classification of Austrian workers in terms of wage mobility. The database consists of 9402 time series for men entering the labor market in 1975 to 1980 at an age of at most 25 years. The series represent gross monthly wages in May of successive years and exhibit individual lengths ranging from 2 to 32 years with the median length being equal to 22. This time series data set is available through the R package \textbf{bayesMCClust} (object \textit{MCCExtExampleData}) \cite{bayesmcclustpackage}, and it was originally taken from the Austrian Social Security Database (ASSD) \cite{zweimuller2009austrian}. It is worth highlighting that a slightly modified version of this data collection was used in \cite{pamminger2010model} to perform clustering of categorical series. Therefore, the application presented in this section involves a case study which has already been established in the time series clustering literature. 

Following \cite{pamminger2010model}, the gross monthly wage is divided into six categories labelled by the integers from 0 to 5. Category zero corresponds to zero-income or nonemployment (which is not equivalent to be out of labour force). Categories one to five correspond to the quintiles of the income distribution, which are determined for each year from all nonzero wages observed in that year for the population of all male employees in Austria. As it is stated in \cite{pamminger2010model}, the consideration of wage categories has the advantage that no inflation adjustment has to be made and circumvents the problem that, in Austria, the recorded wages are right-censored. Note that, as a natural ordering exists in the set of wage categories, the series under study can be directly treated as OTS. Table~\ref{percentagecategories} provides the relative frequencies of the different states in the database, indicating that the wage categories are approximately uniformly distributed, with state 1 appearing a slightly higher number of times than the remaining ones.  

\begin{table}[!ht]
	\centering 
	\begin{tabular}{lcccccc} \hline 
		State          & 0 & 1 & 2 & 3 & 4 & 5 \\ \hline 
		Relative frequency & 0.172  & 0.208  &  0.146 &  0.140 & 0.161  & 0.173 \\ \hline 
	\end{tabular}
\caption{Relative frequencies of the different states in the database of Austrian wage mobility.}
\label{percentagecategories}
\end{table}

\subsubsection{Application of clustering algorithms and results}\label{subsubsectionaca2}

The fuzzy $C$-medoids algorithm based on $\widehat{d}_1$ and $\widehat{d}_2$ was applied to the data set of Austrian wage mobility. Here we considered  $\mathcal{L}=\{1\}$ due to the short length of the series in the collection (the first lag resulted significant according to the hypothesis test presented in Section~\ref{subsectionsetl}). Selection of $C$ and $m$ was carried out by using the same procedure as in the previous analysis, resulting in $(C, m)=(2, 2.0)$ for $\widehat{d}_1$ and $(C, m)=(3, 1.8)$ for $\widehat{d}_2$. Thus, both metrics identify underlying partitions with a different number of groups. 


The large number of series in this database makes it unfeasible to show the resulting fuzzy partitions. However, given a metric, the properties of the series in each group can be summarized by independently analyzing the estimated features within the group. We started by examining the $\widehat{d}_1$-based partition. First, each series was assigned to the cluster with the highest membership degree, thus obtaining a crisp partition with 4660 series in the first group ($\mathcal{C}^1_1$) and 4742 series in the second group ($\mathcal{C}^1_2$). Based on this partition, we computed the vectors $\overline{\boldsymbol f}^i=\big(\overline{f}^i_0, \overline{f}^i_1, \ldots, \overline{f}^i_4\big)$ and the matrices $\overline{\boldsymbol F}^i=\big(\overline{f}^i_{j-1k-1}(1)\big)_{1 \leq j, k \leq 5}$, for $i=1,2$, whose elements are the averages of the corresponding features over all series in the $i$th cluster. The resulting values were
$$
\begin{array}{ll}
\overline{\boldsymbol f}^1=(0.28, 0.73, 0.91, 0.97, 0.99), & \, \, \overline{\boldsymbol F}^1=\begin{pmatrix}
	0.15 & 0.24 & 0.26 & 0.28 & 0.28 \\ 
	0.24 & 0.63 & 0.70 & 0.72 & 0.73 \\
	0.26 & 0.68 & 0.86 & 0.90 & 0.91 \\
	0.28 & 0.70 & 0.89 & 0.96 & 0.97 \\
	0.28 & 0.70 & 0.90 & 0.97 & 0.99 
 \end{pmatrix}
\end{array}
$$
$$
\begin{array}{ll}
\overline{\boldsymbol f}^2=(0.08, 0.17, 0.29, 0.48, 0.73), & \, \, \overline{\boldsymbol F}^2=\begin{pmatrix}
	0.02 & 0.03 & 0.03 & 0.04 & 0.06 \\ 
	0.03 & 0.09 & 0.12 & 0.14 & 0.15 \\
	0.04 & 0.10 & 0.19 & 0.25 & 0.27 \\
	0.05 & 0.11 & 0.22 & 0.39 & 0.46 \\
	0.06 & 0.13 & 0.24 & 0.43 & 0.68 
 \end{pmatrix}
\end{array}
$$

The average vectors $\overline{\boldsymbol f}^1$ and $\overline{\boldsymbol f}^2$ are very different, indicating that the series in cluster $\mathcal{C}^1_2$ generally take larger states than the ones in $\mathcal{C}^1_1$. The average matrices $\overline{\boldsymbol F}^1$ and $\overline{\boldsymbol F}^2$ also reveal clearly dissimilar dependence structures in both groups, although their values are rather difficult to interpret. Hence, to shed light on the behaviour pattern at each cluster, we decided to compute the average values of the $\widehat{d}_2$-based features according to the clustering partition returned by $\widehat{d}_1$. The new features are provided in the upper part of Table~\ref{featuresd2bygroup}. 

\begin{table}[!ht]
\centering
\begin{tabular}{cc|ccccc}
\hline
 Method    & Cluster &  $\frac{\widehat{\text{loc}}_{d_{\text{o},1}}}{5}$    & $\frac{2\widehat{\text{disp}}_{d_{\text{o},1}}}{5}$ & $\frac{\widehat{\text{asym}}_{d_{\text{o},1}}}{5}$ & $\frac{\widehat{\text{skew}}_{d_{\text{o},1}}}{5}$ & $\widehat{\kappa}_{d_{\text{o},1}}(1)$ \\ \hline 
$\widehat{d}_1$ & $\mathcal{C}^1_1$       &    0.50   & 0.38 & 0.31 & 0.44 & 0.17      \\
     & $\mathcal{C}^1_2$       &     0.80 & 0.55 & 0.33 & -0.49 & 0.41     \\ \hline 
$\widehat{d}_2$ & $\mathcal{C}^2_1$       &    0.52  & 0.27 & 0.45 & 0.57  & 0.01     \\
    & $\mathcal{C}^2_2$       &    0.89  & 0.47 & 0.46 & -0.65 & 0.43    \\
     & $\mathcal{C}^2_3$       &    0.55 & 0.62 & 0.08 & 0.03 & 0.41    \\ \hline 
\end{tabular}
\caption{Average values of the $\widehat{d}_2$-based features in each group concerning the clustering solutions produced by $\widehat{d}_1$ and $\widehat{d}_2$ in the data set of Austrian wage mobility.}
\label{featuresd2bygroup}
\end{table}

As expected from the values of $\overline{\boldsymbol f}^1$ and $\overline{\boldsymbol f}^2$, the averages for the first four $\widehat{d}_2$-based measures indicate that both groups are clearly different in terms of marginal distributions. In particular, the series in $\mathcal{C}^1_1$ exhibit  positive skewness (tendency to lower wage categories) in contrast to the negative skewness usually displayed by the series in $\mathcal{C}^1_2$. Concerning the serial behaviour, cluster $\mathcal{C}^1_1$ is associated with a lower degree of positive dependence than $\mathcal{C}^1_2$. Based on previous considerations, a description of the Austrian labour market in terms of social mobility could be provided. Indeed, individuals with higher wages ($\mathcal{C}^1_2$) experience a lower degree of social mobility (i.e., a decline in income), since high states tend to be followed by high states. On the contrary, a more pronounced level of social mobility is observed for employees with lower salaries ($\mathcal{C}^1_1$), which indicates that these individuals are more likely to get a promotion. 

\begin{figure}[!ht]
	\centering
	\includegraphics[width=1\textwidth]{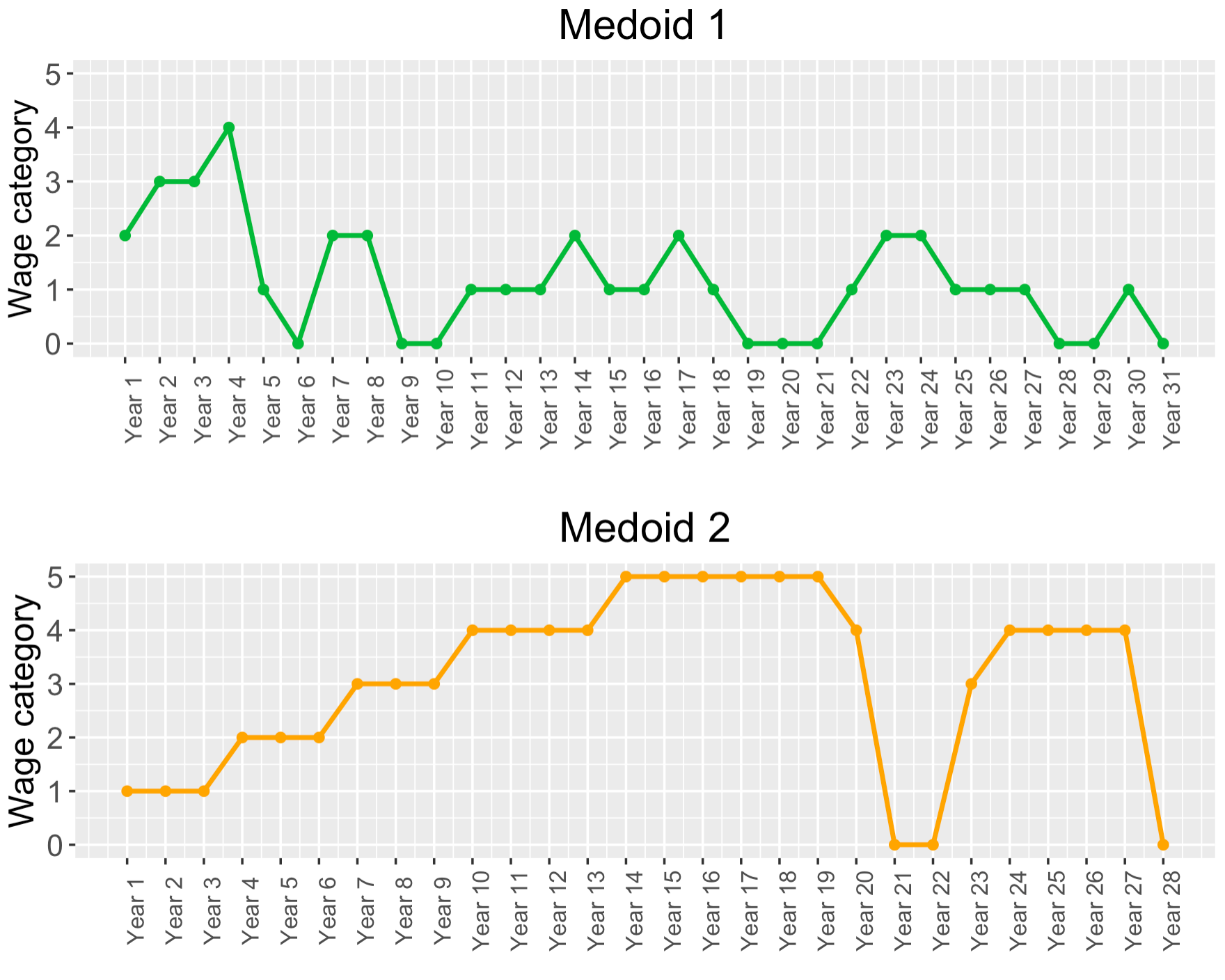}
	\caption{Medoid time series according to the 2-cluster solution produced by distance $\widehat{d}_1$ in the data set of Austrian wage mobility.}
	\label{medoidsd1}
\end{figure}

Figure~\ref{medoidsd1} shows the medoid series with $\widehat{d}_1$. Although the identification of clear patterns in this kind of graphs is usually challenging, some interesting insights can be obtained from both plots. For instance, the medoid of cluster $\mathcal{C}^1_2$ puts more weight on higher categories besides displaying a stronger tendency to generate long runs (positive dependence). Note that these considerations are consistent with the average features displayed in the upper part of Table~\ref{featuresd2bygroup}.   
 
Similar analyses can be carried out by considering the clustering solution produced by $\widehat{d}_2$. The crisp version of this partition includes 2878, 3121 and 3403 series in the first ($\mathcal{C}^2_1$), second ($\mathcal{C}^2_2$)  and third  ($\mathcal{C}^2_3$) groups, respectively. The average values of the $\widehat{d}_2$-based features with respect to each group are provided in the lower part of Table~\ref{featuresd2bygroup}. Note that this 3-cluster solution can be interpreted as a refinement of the above partition. In fact, cluster $\mathcal{C}^2_1$ is again associated with high skewness and low positive dependence (thus high social mobility). A similar reasoning can be made for the second group. Cluster $\mathcal{C}^2_3$ represents individuals with middle income and a low level of mobility. The clear connection between both clustering partitions becomes evident in the confusion matrix given in Table~\ref{confusionmatrix}, where: (i) all the series in  $\mathcal{C}^2_2$ belong to $\mathcal{C}^1_2$, (ii) only 45 series of $\mathcal{C}^2_1$ ($1.56\%$) fall outside $\mathcal{C}^1_1$, and (iii) the additional group identified by $\widehat{d}_2$, $\mathcal{C}^2_3$, is formed by series of both clusters $\mathcal{C}^1_1$ and $\mathcal{C}^1_2$ in similar amounts.  

\begin{table}[!ht]
\centering
\begin{tabular}{llcc|}  
     &  & \multicolumn{2}{c}{$\widehat{d}_1$}   \\  \cline{3-4}
     &  &  \multicolumn{1}{|c}{$\mathcal{C}^1_1$} & $\mathcal{C}^1_2$  \\ \cline{2-4}  
& \multicolumn{1}{|l|}{$\mathcal{C}^2_1$}       &    2833  &  45  \\
  $\widehat{d}_2$   & \multicolumn{1}{|l|}{$\mathcal{C}^2_2$}      &    0   & 3121   \\
     & \multicolumn{1}{|l|}{$\mathcal{C}^2_3$}       &    1827  &  1576  \\ \cline{2-4} 
\end{tabular}
\caption{Confusion matrix for the crisp clustering solutions produced by $\widehat{d}_1$ and $\widehat{d}_2$ in the data set of Austrian wage mobility.}
\label{confusionmatrix}
\end{table}

In sum, the analyses carried out throughout Sections~\ref{subsectionapplication1} and \ref{subsectionapplication2} illustrate the usefulness of both metrics $\widehat{d}_1$ and $\widehat{d}_2$ when performing fuzzy clustering of OTS in real data sets. Specifically, they highlight the importance of the fuzzy paradigm when attempting to achieve meaningful conclusions from the resulting partitions.

\section{Conclusions}\label{sectionconclusions}

In this paper, we have proposed two novel distances between OTS which automatically take advantage of the inherent ordering in the series' range. The first metric considers proper estimates of the cumulative probabilities, while the second distance employs some ordinal features describing the behaviour of a given OTS. Both distances are formed by two components. The first one evaluates discrepancies between the marginal distributions of the series, while the second component assesses differences in terms of serial dependence structures. The metrics are used as input to the classical fuzzy $C$-medoids algorithm, which allows for the assignment of gradual memberships of the OTS to the different groups. This is particularly useful when dealing with time series data sets, where different amounts of dissimilarity between the underlying processes or changes on the dynamic behaviours over time are frequent. 

To assess the performance of the proposed clustering algorithms, several simulation experiments were carried out including scenarios formed by OTS pertaining to well-defined clusters and scenarios involving series generated from an outlying stochastic process. Different types of ordinal processes were considered. The methods were compared with several procedures based on alternative dissimilarities. Overall, the proposed clustering techniques showed the best performance. Specifically, they outperformed some techniques specifically designed to deal with real-valued and with nominal time series, which highlights the importance of considering the underlying ordering when performing OTS clustering.  Extensions of both clustering procedures were also constructed by giving different weights to the marginal and serial components of the proposed metrics. The weighting system allows the importance of each component in the computation of the clustering partition to be automatically determined during the minimisation phase. The advantages of the weighted algorithms with respect to the standard ones in terms of clustering accuracy were analysed. The results showed that significant improvements are frequently observed when employing the weighted procedure based on the second of the introduced metrics. The usefulness of the proposed clustering algorithms was illustrated by means of two applications involving economic time series. In both cases, interesting conclusions were reached.

There are at least three interesting ways through which this work could be extended. First, robust versions of the proposed methods could be constructed by considering the so-called metric, noise, and trimmed approaches \cite{lafuente2020robust, d2016garch, lopez2022quantile2}, which adjust the objective function of the clustering algorithm in a suitable manner so that outlier series do not pervert the resulting partition. Second, a spatial penalisation term could be incorporated in the objective function of the procedures in order to deal with OTS data sets containing geographical information \cite{lopez2022spatial, coppi2010fuzzy}, like the one considered in Section \ref{subsectionapplication1}. Third, the clustering methods could be modified in such a way that they can properly handle OTS containing missing data \cite{weiss2021analyzing}. It would be interesting to address these and further topics in future research. 

\section*{Appendix}

In this section, we present the proofs of Propositions~\ref{propd1M} and~\ref{propIterSol}.  

\subsection*{Proof of Proposition~\ref{propd1M}}

\begin{proof}
We shall prove Proposition~\ref{propd1M} by induction in $n$. Denote by $(f_0, f_1, \ldots, f_n)$ and $(g_0, g_1, \ldots, g_n)$ the vectors of cumulative probabilities for processes $\{X_t\}_{t \in \mathbb{Z}}$ and $\{Y_t\}_{t \in \mathbb{Z}}$, respectively. Note that, for $n=1$ (2 states), the distance $d_{1, M}$ can be written as
\begin{equation}
d_{1, M}\big(X_t, Y_t\big)=(f_0-g_0)^2=(p_0-q_0)^2,
\end{equation}
so the assertion of Proposition~\ref{propd1M} is true. Assume now that Proposition~\ref{propd1M} holds for $n=N$, $N \in \mathbb{N}$. For processes $X_t$ and $Y_t$ with range $\{s_0, s_1, \ldots, s_{N+1}\}$, we have
\begin{equation}\label{distanceN+1}
	d_{1, M}\big(X_t, Y_t\big)=\sum_{i=0}^{N-1}(f_i-g_i)^2+(f_N-g_N)^2. 
\end{equation}

Note that the term $\sum_{i=0}^{N-1}(f_i-g_i)^2$ in \eqref{distanceN+1} can be seen as the distance $d_{1, M}$ between two processes $X_t^*$ and $Y_t^*$ with range $\{s_0, s_1, \ldots, s_N\}$, marginal probabilities $(p_0, p_1, \dots, p_{N-1}, 1-f_{N-1})$ and $(q_0, q_1, \dots, q_{N-1}, 1-g_{N-1})$, and cumulative probabilities $(f_0, f_1, \dots, f_{N-1}, 1)$ and $(g_0, g_1, \dots, g_{N-1}, 1)$, respectively. Moreover, by taking into account that $f_i=\sum_{j=0}^{i}p_j$ and $g_i=\sum_{j=0}^{i}q_j$, the term $(f_N-g_N)^2$ in \eqref{distanceN+1} can be expressed as
\begin{equation}
	(f_N-g_N)^2=\bigg(\sum_{i=0}^N(p_i-q_i)\bigg)^2=\sum_{i=0}^{N}(p_i-q_i)^2+2\sum_{j=0}^{N-1}\sum_{k=j+1}^{N}(p_j-q_j)(p_k-q_k).  
\end{equation}

Considering now the induction hypothesis, and plugging-in the previous expression for $(f_N-g_N)^2$ in \eqref{distanceN+1}, the distance $d_{1, M}$ between $X_t$ and $Y_t$ can be written as
\begin{equation}
\begin{split}
	d_{1, M}\big(X_t, Y_t\big)=\sum_{i=0}^{N-1}(N-i)(p_i-q_i)^2+\sum_{i=0}^{N-1}(p_i-q_i)^2+(p_N-q_N)^2+ \\ 2\sum_{j=0}^{N-2}\sum_{k=j+1}^{N-1}(N-k)(p_j-q_j)(p_k-q_k)+2\sum_{j=0}^{N-2}\sum_{k=j+1}^{N-1}(p_j-q_j)(p_k-q_k)+ \\ 2\sum_{j=0}^{N-1}(p_j-q_j)(p_N-q_N)=\sum_{i=0}^{N-1}(N+1-i)(p_i-q_i)^2+(p_N-q_N)^2+  \\ 2\sum_{j=0}^{N-2}\sum_{k=j+1}^{N-1}(N+1-k)(p_j-q_j)(p_k-q_k) + 2\sum_{j=0}^{N-1}(p_j-q_j)(p_N-q_N)= \\ \sum_{i=0}^{N}(N+1-i)(p_i-q_i)^2+2\sum_{j=0}^{N-1}\sum_{k=j+1}^{N}(N+1-k)(p_j-q_j)(p_k-q_k).
\end{split}
\end{equation}

Therefore, the assertion is also true for $n=N+1$, and thus for all $n \in \mathbb{N}$. The proof of Proposition~\ref{propd1M} is completed. 

\end{proof} 

\subsection*{Proof of Proposition~\ref{propIterSol}}

\begin{proof}
The iterative solutions of the constrained minimization problem (15) are obtained via the Lagrangian multipliers method. First, for $j=1,2$, consider the Lagrangian function taking the form
\begin{equation}\label{lagrangian}
	\begin{split}
	L(\bm U, \beta, \bm \lambda) = {} 
	\sum_{i=1}^{s}\sum_{c=1}^{C}u_{ic}^m\Big[\beta^2\widehat{d}_{j, M}(i, c)+(1-\beta)^2\widehat{d}_{j, B}\big(i, c\big)\Big]  - \sum_{i=1}^{s}\lambda_i\big(\sum_{c=1}^{C}u_{ic}-1\big),
	\end{split}
\end{equation}
where $\bm \lambda=\{\lambda_1,\ldots, \lambda_s\}$ is the set of Lagrange multipliers concerning the constraints of the membership degrees. By fixing $\beta \in [0, 1]$ and setting equal to zero the partial derivatives  of $L$ with respect to $u_{ic}$ and $\lambda_{i}$, for arbitrary $i \in \{1,\ldots,s\}$ and $c\in \{1,\ldots,C\}$, we obtain that
\begin{equation*}
	\frac{ \partial L(\bm U, \beta, \bm \lambda)}{\partial u_{ic}}=0 \, \, \, \, \, \text{and} \, \, \, \, \, 
	\frac{ \partial L(\bm U, \beta, \bm \lambda)}{\partial \lambda_{i}}=0
\end{equation*}
is equivalent to
\begin{equation}\label{lagrangianmemberships}
	mu_{ic}^{m-1} \Big[\beta^2\widehat{d}_{j, M}(i, c)+(1-\beta)^2\widehat{d}_{j, B}(i, c)\Big]-\lambda_i=0 \, \, \, \, \, \text{and} \, \, \, \, \, 
	\sum_{c'=1}^{C}u_{ic'}-1=0. 
\end{equation}

From the first equation in \eqref{lagrangianmemberships}, we can express $u_{ic}$ as
\begin{equation}\label{lagrangianmembershipsstep1}
	u_{ic}=\bigg(\frac{\lambda_i}{m}\bigg)^{\frac{1}{m-1}}\Big[\beta^2\widehat{d}_{j, M}(i, c)+(1-\beta)^2\widehat{d}_{j, B}(i, c)\Big]^{\frac{-1}{m-1}}.
\end{equation}

By introducing \eqref{lagrangianmembershipsstep1} in the second equation of \eqref{lagrangianmemberships}, we obtain
\begin{equation}\label{lagrangianmembershipsstep2}
	\bigg(\frac{\lambda_i}{m}\bigg)^{\frac{1}{m-1}}\sum_{c'=1}^{C}\Big[\beta^2\widehat{d}_{j, M}(i, c')+(1-\beta)^2\widehat{d}_{j, B}(i, c')\Big]^{\frac{-1}{m-1}}=1,
\end{equation}
which leads to
\begin{equation}\label{lagrangianmembershipsstep3}
	\bigg(\frac{\lambda_i}{m}\bigg)^{\frac{1}{m-1}}=\bigg[\sum_{c'=1}^{C}\Big[\beta^2\widehat{d}_{j, M}(i, c')+(1-\beta)^2\widehat{d}_{j, B}(i, c')\Big]^{\frac{-1}{m-1}}\bigg]^{-1}.
\end{equation}

Lastly, by replacing \eqref{lagrangianmembershipsstep3} in \eqref{lagrangianmembershipsstep1}, the membership degree $u_{ic}$ can be expressed as
\begin{equation}  \label{itsol1}
	u_{ic}=\Bigg[\sum_{c'=1}^{C}\Bigg(\frac{\beta^2\widehat{d}_{j, M}(i, c)+(1-\beta)^2\widehat{d}_{j, B}(i, c)}{\beta^2\widehat{d}_{j, M}(i, c')+(1-\beta)^2\widehat{d}_{j, B}(i, c')}\Bigg)^{\frac{1}{m-1}}\Bigg]^{-1},
\end{equation}
which gives the iterative solutions for the membership degrees. 

The iterative solution for $\beta$ can be obtained in a similar way. We proceed by fixing $u_{ic}$ and setting equal to zero the partial derivative of $L$ with respect to $\beta$, i.e., $\frac{ \partial L(\bm U, \beta, \bm \lambda)}{\partial \beta}=0$. This is equivalent to 
\begin{equation}
	\sum_{i=1}^{s}\sum_{c=1}^{C}u_{ic}^m\Big[\beta\widehat{d}_{j, M}(i, c)-(1-\beta)\widehat{d}_{j, B}\big(i, c\big)\Big]=0,
\end{equation}
yielding
\begin{equation}\label{lastbeta}
	\sum_{i=1}^{s}\sum_{c=1}^{C}u_{ic}^m\Big[\beta\big(\widehat{d}_{j, M}(i, c)+\widehat{d}_{j, B}(i, c)\big)-\widehat{d}_{j, B}\big(i, c\big)\Big]=0.
\end{equation}

From equation \eqref{lastbeta}, we can conclude that
\begin{equation}\label{itsol2}
	\beta=\frac{\sum_{i=1}^{s}\sum_{c=1}^{C}u_{ic}^m\widehat{d}_{j, B}\big(i, c\big)}{\sum_{i=1}^{s}\sum_{c=1}^{C}u_{ic}^m\big(\widehat{d}_{j, M}(i, c)+\widehat{d}_{j, B}(i, c)\big)},
\end{equation}
which gives the iterative solution for $\beta$. 
\end{proof}
 \section*{Acknowledgments}
The research of Ángel López-Oriona and José. A. Vilar has been supported by the Ministerio de Economía y Competitividad (MINECO) grant MTM2017-87197-C3-1-P,  the Xunta de Galicia through the ERDF (Grupos de Referencia Competitiva ED431C-2016-015), and the Centro de Investigación de Galicia ``CITIC'', funded by Xunta de Galicia and the European Union (European Regional Development Fund-Galicia 2014-2020 Program), by grant ED431G 2019/01. The author Ángel López-Oriona would like to thank Prof.\ Christian H.\ Wei\ss{} for his kindness during the doctoral stay at the Helmut Schmidt University of Hamburg, where this research was carried out.

\bibliography{mybibfile}

\begin{thebibliography}{10}
\expandafter\ifx\csname url\endcsname\relax
  \def\url#1{\texttt{#1}}\fi
\expandafter\ifx\csname urlprefix\endcsname\relax\def\urlprefix{URL }\fi
\expandafter\ifx\csname href\endcsname\relax
  \def\href#1#2{#2} \def\path#1{#1}\fi

\bibitem{liao2005clustering}
T.~W. Liao, Clustering of time series data—a survey, Pattern Recognition
  38~(11) (2005) 1857--1874.

\bibitem{aghabozorgi2015time}
S.~Aghabozorgi, A.~S. Shirkhorshidi, T.~Y. Wah, Time-series clustering--a
  decade review, Information Systems 53 (2015) 16--38.

\bibitem{maharaj2019time}
E.~A. Maharaj, P.~D'Urso, J.~Caiado, Time Series Clustering and Classification,
  Chapman and Hall/CRC, 2019.

\bibitem{izakian2015fuzzy}
H.~Izakian, W.~Pedrycz, I.~Jamal, Fuzzy clustering of time series data using
  dynamic time warping distance, Engineering Applications of Artificial
  Intelligence 39 (2015) 235--244.

\bibitem{luczak2016hierarchical}
M.~{\L}uczak, Hierarchical clustering of time series data with parametric
  derivative dynamic time warping, Expert Systems with Applications 62 (2016)
  116--130.

\bibitem{d2021trimmed}
P.~D’Urso, L.~De~Giovanni, R.~Massari, Trimmed fuzzy clustering of financial
  time series based on dynamic time warping, Annals of Operations Research
  299~(1) (2021) 1379--1395.

\bibitem{frohwirth2008model}
S.~Fr{\"o}hwirth-Schnatter, S.~Kaufmann, Model-based clustering of multiple
  time series, Journal of Business \& Economic Statistics 26~(1) (2008) 78--89.

\bibitem{corduas2008time}
M.~Corduas, D.~Piccolo, Time series clustering and classification by the
  autoregressive metric, Computational Statistics \& Data Analysis 52~(4)
  (2008) 1860--1872.

\bibitem{d2013autoregressive}
P.~D’Urso, D.~Di~Lallo, E.~A. Maharaj, Autoregressive model-based fuzzy
  clustering and its application for detecting information redundancy in air
  pollution monitoring networks, Soft Computing 17~(1) (2013) 83--131.

\bibitem{d2016garch}
P.~D'Urso, L.~De~Giovanni, R.~Massari, {GARCH}-based robust clustering of time
  series, Fuzzy Sets and Systems 305 (2016) 1--28.

\bibitem{d2009autocorrelation}
P.~D’Urso, E.~A. Maharaj, Autocorrelation-based fuzzy clustering of time
  series, Fuzzy Sets and Systems 160~(24) (2009) 3565--3589.

\bibitem{maharaj2011fuzzy}
E.~A. Maharaj, P.~D’Urso, Fuzzy clustering of time series in the frequency
  domain, Information Sciences 181~(7) (2011) 1187--1211.

\bibitem{d2012wavelets}
P.~D'Urso, E.~A. Maharaj, Wavelets-based clustering of multivariate time
  series, Fuzzy Sets and Systems 193 (2012) 33--61.

\bibitem{lafuente2016clustering}
B.~Lafuente-Rego, J.~A. Vilar, Clustering of time series using quantile
  autocovariances, Advances in Data Analysis and Classification 10~(3) (2016)
  391--415.

\bibitem{lopez2021quantile}
{\'A}.~L{\'o}pez-Oriona, J.~A. Vilar, Quantile cross-spectral density: A novel
  and effective tool for clustering multivariate time series, Expert Systems
  with Applications 185 (2021) 115677.

\bibitem{lopez2022quantile1}
{\'A}.~L{\'o}pez-Oriona, J.~A. Vilar, P.~D'Urso, Quantile-based fuzzy
  clustering of multivariate time series in the frequency domain, Fuzzy Sets
  and Systems 443 (2022) 115--154.

\bibitem{lopez2022quantile2}
{\'A}.~L{\'o}pez-Oriona, P.~D'Urso, J.~A. Vilar, B.~Lafuente-Rego,
  Quantile-based fuzzy {C}-means clustering of multivariate time series: Robust
  techniques, International Journal of Approximate Reasoning 150 (2022) 55--82.

\bibitem{singhal2005clustering}
A.~Singhal, D.~E. Seborg, Clustering multivariate time-series data, Journal of
  Chemometrics: A Journal of the Chemometrics Society 19~(8) (2005) 427--438.

\bibitem{egri2017cross}
A.~Egri, I.~Horv{\'a}th, F.~Kov{\'a}cs, R.~Molontay, K.~Varga,
  Cross-correlation based clustering and dimension reduction of multivariate
  time series, in: 2017 IEEE 21st International Conference on Intelligent
  Engineering Systems (INES), IEEE, 2017, pp. 000241--000246.

\bibitem{pealat2021improved}
C.~Pealat, G.~Bouleux, V.~Cheutet, Improved time-series clustering with {UMAP}
  dimension reduction method, in: 2020 25th International Conference on Pattern
  Recognition (ICPR), IEEE, 2021, pp. 5658--5665.

\bibitem{bezdek2013pattern}
J.~C. Bezdek, Pattern Recognition with Fuzzy Objective Function Algorithms,
  Springer Science \& Business Media, 2013.

\bibitem{miyamoto2008algorithms}
S.~Miyamoto, H.~Ichihashi, K.~Honda, H.~Ichihashi, Algorithms for Fuzzy
  Clustering, Vol.~10, Springer, 2008.

\bibitem{etienne2014model}
C.~Etienne, O.~Latifa, Model-based count series clustering for bike sharing
  system usage mining: a case study with the v{\'e}lib' system of {P}aris, ACM
  Transactions on Intelligent Systems and Technology (TIST) 5~(3) (2014) 1--21.

\bibitem{cerqueti2022ingarch}
R.~Cerqueti, P.~D’Urso, L.~De~Giovanni, R.~Mattera, V.~Vitale,
  {INGARCH}-based fuzzy clustering of count time series with a football
  application, Machine Learning with Applications 10 (2022) 100417.

\bibitem{pamminger2010model}
C.~Pamminger, Fr{\"u}hwirth-Schnatter, Model-based clustering of categorical
  time series, Bayesian Analysis 5~(2) (2010) 345--368.

\bibitem{garcia2015framework}
M.~Garc{\'\i}a-Magari{\~n}os, J.~A. Vilar, A framework for dissimilarity-based
  partitioning clustering of categorical time series, Data Mining and Knowledge
  Discovery 29~(2) (2015) 466--502.

\bibitem{jahanshahi2022ntreeclus}
H.~Jahanshahi, M.~G. Baydogan, {nTreeClus}: A tree-based sequence encoder for
  clustering categorical series, Neurocomputing 494 (2022) 224--241.

\bibitem{lopez2023hard}
{\'A}.~L{\'o}pez-Oriona, J.~A. Vilar, P.~D’Urso, Hard and soft clustering of
  categorical time series based on two novel distances with an application to
  biological sequences, Information Sciences 624 (2023) 467--492.

\bibitem{melnykov2016clickclust}
V.~Melnykov, {ClickClust}: An {R} package for model-based clustering of
  categorical sequences, Journal of Statistical Software 74 (2016) 1--34.

\bibitem{melnykov2014package}
V.~Melnykov, R.~Rostamian, M.~V. Melnykov, Package `{ClickClust}'.

\bibitem{weiss2019distance}
C.~H. Wei{\ss}, Distance-based analysis of ordinal data and ordinal time
  series, Journal of the American Statistical Association 115~(531) (2020)
  1189--1200.

\bibitem{weiss2020regime}
C.~H. Wei{\ss}, Regime-switching discrete {ARMA} models for categorical time
  series, Entropy 22~(4) (2020) 458.

\bibitem{koss2022hierarchical}
J.~Koss, S.~Tinaz, H.~D. Tagare, Hierarchical denoising of ordinal time series
  of clinical scores, IEEE Journal of Biomedical and Health Informatics 26~(7)
  (2022) 3507--3516.

\bibitem{weiss2018introduction}
C.~H. Wei{\ss}, An Introduction to Discrete-valued Time Series, John Wiley \&
  Sons, 2018.

\bibitem{hoppner1999fuzzy}
F.~H{\"o}ppner, F.~Klawonn, R.~Kruse, T.~Runkler, Fuzzy Cluster Analysis:
  Methods for Classification, Data Analysis and Image Recognition, John Wiley
  \& Sons, 1999.

\bibitem{kaufman2009finding}
L.~Kaufman, P.~J. Rousseeuw, Finding Groups in Data: an Introduction to Cluster
  Analysis, John Wiley \& Sons, 2009.

\bibitem{dunn1973fuzzy}
J.~C. Dunn, A fuzzy relative of the {ISODATA} process and its use in detecting
  compact well-separated clusters, Journal of Cybernetics 3~(3) (1973) 32--57.

\bibitem{coppi2010fuzzy}
R.~Coppi, P.~D’Urso, P.~Giordani, A fuzzy clustering model for multivariate
  spatial time series, Journal of Classification 27~(1) (2010) 54--88.

\bibitem{lopez2022spatial}
{\'A}.~L{\'o}pez-Oriona, P.~D'Urso, J.~A. Vilar, B.~Lafuente-Rego, Spatial
  weighted robust clustering of multivariate time series based on quantile
  dependence with an application to mobility during {COVID}-19 pandemic, IEEE
  Transactions on Fuzzy Systems 30~(9) (2022) 3990--4004.

\bibitem{alonso2020hierarchical}
A.~M. Alonso, F.~J. Nogales, C.~Ruiz, Hierarchical clustering for smart meter
  electricity loads based on quantile autocovariances, IEEE Transactions on
  Smart Grid 11~(5) (2020) 4522--4530.

\bibitem{weiss2009new}
C.~H. Wei{\ss}, A new class of autoregressive models for time series of
  binomial counts, Communications in Statistics--Theory and Methods 38~(4)
  (2009) 447--460.

\bibitem{ristic2016binomial}
M.~M. Risti{\'c}, C.~H. Wei{\ss}, A.~D. Janji{\'c}, A binomial integer-valued
  {ARCH} model, The International Journal of Biostatistics 12~(2) (2016)
  20150051.

\bibitem{arabie1981overlapping}
P.~Arabie, J.~D. Carroll, W.~DeSarbo, J.~Wind, Overlapping clustering: A new
  method for product positioning, Journal of Marketing Research 18~(3) (1981)
  310--317.

\bibitem{vilar2018quantile}
J.~A. Vilar, B.~Lafuente-Rego, P.~D'Urso, Quantile autocovariances: a powerful
  tool for hard and soft partitional clustering of time series, Fuzzy Sets and
  Systems 340 (2018) 38--72.

\bibitem{campello2007fuzzy}
R.~J. Campello, A fuzzy extension of the rand index and other related indexes
  for clustering and classification assessment, Pattern Recognition Letters
  28~(7) (2007) 833--841.

\bibitem{levinson1949wiener}
N.~Levinson, The {W}iener {RMS} (root mean square) error criterion in filter
  design and prediction, Journal of Mathematics and Physics 25~(1--4) (1949)
  261--278.

\bibitem{durbin1960fitting}
J.~Durbin, The fitting of time-series models, Revue de l'Institut International
  de Statistique 28~(3) (1960) 233--244.

\bibitem{xie1991validity}
X.~L. Xie, G.~Beni, A validity measure for fuzzy clustering, IEEE Transactions
  on Pattern Analysis \& Machine Intelligence 13~(08) (1991) 841--847.

\bibitem{kwon1998cluster}
S.~H. Kwon, Cluster validity index for fuzzy clustering, Electronics Letters
  34~(22) (1998) 2176--2177.

\bibitem{tang2005improved}
Y.~Tang, F.~Sun, Z.~Sun, Improved validation index for fuzzy clustering, in:
  Proceedings of the 2005, American Control Conference, 2005., IEEE, 2005, pp.
  1120--1125.

\bibitem{bensaid1996validity}
A.~M. Bensaid, L.~O. Hall, J.~C. Bezdek, L.~P. Clarke, M.~L. Silbiger, J.~A.
  Arrington, R.~F. Murtagh, Validity-guided (re) clustering with applications
  to image segmentation, IEEE Transactions on Fuzzy Systems 4~(2) (1996)
  112--123.

\bibitem{bayesmcclustpackage}
C.~Pamminger, {bayesMCClust}: Mixtures-of-Experts {M}arkov Chain Clustering and
  {D}irichlet Multinomial Clustering, {R} package version 1.0 (2018).

\bibitem{zweimuller2009austrian}
J.~Zweim{\"u}ller, R.~Winter-Ebmer, R.~Lalive, A.~Kuhn, J.-P. Wuellrich,
  O.~Ruf, S.~B{\"u}chi, Austrian social security database, Available at SSRN
  1399350.

\bibitem{lafuente2020robust}
B.~Lafuente-Rego, P.~D’Urso, J.~A. Vilar, Robust fuzzy clustering based on
  quantile autocovariances, Statistical Papers 61~(6) (2020) 2393--2448.

\bibitem{weiss2021analyzing}
C.~H. Wei{\ss}, Analyzing categorical time series in the presence of missing
  observations, Statistics in Medicine 40~(21) (2021) 4675--4690.

\end{thebibliography}

\end{document}